\documentclass{article}
\usepackage{pgf}
\usepackage{tikz}
\usetikzlibrary{calc,arrows,matrix,graphs,shapes,snakes,automata,backgrounds,petri,positioning,decorations.pathmorphing}
\usepackage{verbatim}
\usepackage{amsmath}
\usepackage{amssymb}
\usepackage{latexsym}
\usepackage{proof}
\usepackage{array}
\usepackage{calc}
\usepackage{mathdots}
\usepackage{stmaryrd}
\usepackage{natbib}
\usepackage[english]{babel}
\usepackage{proof}
\usepackage{extarrows}
\usepackage{linguex}
\usepackage[colorlinks,citecolor=red]{hyperref}

\newtheorem{theorem}{Theorem}
\newtheorem{definition}[theorem]{Definition}

\newtheorem{lemma}[theorem]{Lemma}

\newcommand{\qed}{\hfill $\Box$ \vspace{3mm}}

\newcommand{\editout}[1]{}

\newcommand{\proofspace}{\vphantom{()}}

\newcommand{\ldl}{\mathbin{\backslash}}
\newcommand{\ldr}{\mathbin{/}}
\newcommand{\lpr}{\mathbin{\bullet}}
\newcommand{\bpr}{\mathbin{\varodot}}
\newcommand{\bdl}{\fatbslash}
\newcommand{\bdr}{\!\!\fatslash}
\newcommand{\lstr}{\mathbin{\circ}}
\newcommand{\bstr}{\mathbin{\circledcirc}}
\newcommand{\lidstr}{1}
\newcommand{\lidc}{t}

\newcommand{\apsnode}[2]{\overset{#1}{\underset{#2^{\rule{0pt}{1.2ex}}}{\centerdot}}}

\newcommand{\apsnodei}{\centerdot_{\rule{0pt}{1.5ex}}}
\newcommand{\emptynode}{\phantom{$\apsnodei$}}
\newcommand{\nodeindex}[1]{\textcolor{white}{\textbf{#1}}}

\newcommand{\nllam}{\ensuremath{\text{NL}_{\lambda}}}
\newcommand{\llam}{\ensuremath{\text{L}_{\lambda}}}

\newcommand{\nlambek}{\ensuremath{\text{NL}}}
\newcommand{\betared}{\beta}
\newcommand{\betaexp}{\beta^{-1}}

\tikzset{pas/.style={fill=gray!60}, 
act/.style={fill=gray!30},
main/.style={draw,fill=white},
ctx/.style={rounded rectangle,minimum size=7mm},
val/.style={rectangle,minimum size=7mm},
cmd/.style={chamfered rectangle,draw,fill=white},
tns/.style={circle,minimum size=4.5mm,draw,fill=white},  
par/.style={circle,minimum size=4.5mm,draw,fill=black},  
ttns/.style={circle,minimum size=4.5mm,inner sep=2pt,draw,fill=white,double=white},  
ppar/.style={circle,minimum size=4.5mm,inner sep=2pt,draw=white,fill=black,double=black},  
minipar/.style={circle,minimum size=5mm,draw,fill=black}, 
pn/.style={rounded corners, rectangle,fill=blue!30,draw,minimum size=15mm},
medpn/.style={rounded corners, rectangle,fill=blue!30,draw,minimum size=20mm},
 bigpn/.style={rounded corners, rectangle,fill=blue!30,draw,minimum size=25mm}}

\tikzset{every picture/.style={scale=.9, transform shape, node distance=7mm}}

\title{Proof-theoretic aspects of \nllam{}}
\author{Richard Moot}

\begin{document}
\maketitle

\section{Introduction}

In this paper, we present a proof-theoretic analysis of the logic \nllam{} \citep{bs14cont,barker2019nllam}. We notably introduce a novel calculus of proof nets and prove it is sound and complete with respect to the sequent calculus for the logic. We study decidability and complexity of the logic using this new calculus, proving a new upper bound for complexity of the logic (showing it is in NP) and a new lower bound for the class of formal language generated by the formalism (mildly context-sensitive languages extended with a permutation closure operation). Finally, thanks to this new calculus, we present a novel comparison between \nllam{} and the hybrid type-logical grammars of \citet{kl20tls}. We show there is an unexpected convergence of the natural language analyses proposed in the two formalism.
 In addition to studying the proof-theoretic properties of \nllam{}, we greatly extends its linguistic coverage.


\section{Sequent calculus for \nllam{}}
\label{sec:seq}

The logic \nllam{}, introduced by Chris Barker and Chung-Chieh Shan \citep{bs14cont,barker2019nllam}, is in many ways a standard multimodal logic in the tradition of type-logical grammars \citep{M95}. It uses two families of connectives:
\begin{enumerate}
\item  the connectives $\{ \ldl, \lpr, \ldr \}$ of the non-associative Lambek calculus \nlambek{}  and the corresponding structural connective `$\lstr$', used for the standard composition operations, and 
\item the continuation mode connectives $\{ \bdl, \bpr, \bdr \}$ and corresponding structural connective `$\bstr$', used for scope taking and other discontinuous phenomena.
 \end{enumerate}

  What makes the logic unusual is its main structural rule, stipulated as shown in Equation~\ref{eq:strlambda}.
\begin{align}\label{eq:strlambda}
\Gamma[\Delta] &\Leftrightarrow \Delta\bstr \lambda x.\Gamma[x]
\end{align}
In the equation `$\bstr$' is the structural connective for the scope taking mode. When using Equation~\ref{eq:strlambda} from left to right, $x$ must be a fresh variable, not occurring in the context $\Gamma[]$ or in the structure $\Delta$. When seeing `$\bstr$' as a type of application (with the right daughter the function and the left daughter its argument) then reading the equation from right to left corresponds to a type of beta reduction, and the other direction to a form of beta expansion. We also adopt a principle similar to alpha equivalence in the lambda calculus, by treating proof as equivalent if they differ only in the name of abstracted variables.

\begin{table}
\begin{center}
\begin{tabular}{ccc}
\multicolumn{3}{c}{\textbf{Identity group}}\\[0.8em]
\infer[\textit{Ax}]{A\vdash A} && \infer[\textit{Cut}]{\Gamma[\Delta]\vdash C}{\Delta\vdash A & \Gamma[A]\vdash C} \\[0.8em]
\multicolumn{3}{c}{\textbf{Logical rules for the unit}}\\[0.8em]
\infer[\lidc L]{\Gamma[\lidc]\vdash D}{\Gamma[\lidstr]\vdash D} && \infer[\lidc R]{\lidstr\vdash \lidc}{} \\[0.8em]
\multicolumn{3}{c}{\textbf{Logical rules for \nlambek{}}}\\[0.8em]
\infer[\ldr L]{\Gamma[(C \ldr B \lstr \Delta)] \vdash D}{\Delta \vdash B & \Gamma[C]\vdash D}	&&
\infer[\ldr R]{\Gamma\vdash C\ldr B}{(\Gamma \lstr B)\vdash C} \\[0.8em]
\infer[\ldl L]{\Gamma[(\Delta \lstr A\ldl C)] \vdash D}{\Delta \vdash C & \Gamma[C]\vdash D}	&&
\infer[\ldl R]{\Gamma\vdash A\ldl C}{(A\lstr \Gamma) \vdash C} \\[0.8em]
\infer[\lpr L]{\Gamma[A\lpr B]\vdash D}{\Gamma[(A\lstr B)] \vdash D} &&
\infer[\lpr R]{(\Gamma \lstr \Delta) \vdash A\lpr B}{\Gamma\vdash A &\Delta\vdash B} \\[0.8em]
\multicolumn{3}{c}{\textbf{Logical rules for continuation connectives}}\\[0.8em]
\infer[\bdr L]{\Gamma[(C \bdr B \bstr \Delta)] \vdash D}{\Delta \vdash B & \Gamma[C]\vdash D}	&&
\infer[\bdr R]{\Gamma\vdash C\bdr B}{(\Gamma \bstr B)\vdash C} \\[0.8em]
\infer[\bdl L]{\Gamma[(\Delta \bstr A\bdl C)] \vdash D}{\Delta \vdash C & \Gamma[C]\vdash D}	&&
\infer[\bdl R]{\Gamma\vdash A\bdl C}{(A\bstr \Gamma) \vdash C} \\[0.8em]
\infer[\bpr L]{\Gamma[A\bpr B]\vdash D}{\Gamma[(A\bstr B)] \vdash D} &&
\infer[\bpr R]{(\Gamma \bstr \Delta) \vdash A\bpr B}{\Gamma\vdash A &\Delta\vdash B} 
\end{tabular}
\end{center}
\caption{The sequent calculus rules for \nllam{}: logical rules}
\label{tab:seqnllam}	
\end{table}

\begin{table}
\begin{center}
\begin{tabular}{ccc}
\multicolumn{3}{c}{\textbf{Structural rules for unit}}\\[0.8em]
\infer[\lstr\lidstr]{\Gamma[\Delta]\vdash D}{\Gamma[(\Delta\lstr\lidstr)]\vdash D} && \infer[\lstr\lidstr^{-1}]{\Gamma[(\Delta\lstr\lidstr)]\vdash D}{\Gamma[\Delta]\vdash D} 
\\[0.8em]
\infer[\lidstr\lstr]{\Gamma[\Delta]\vdash D}{\Gamma[(\lidstr\lstr\Delta)]\vdash D} && \infer[\lidstr\lstr^{-1}]{\Gamma[(\lidstr\lstr\Delta)]\vdash D}{\Gamma[\Delta]\vdash D} 
\\[0.8em]
\multicolumn{3}{c}{\textbf{Structural rules for continuations}}\\[0.8em]
\infer[\betared]{\Xi[\Gamma[\Delta]]\vdash D}{\Xi[(\Delta \bstr \lambda x.\Gamma[x])]\vdash D} &&
\infer[\betaexp]{\Xi[(\Delta \bstr \lambda x.\Gamma[x])]\vdash D}{\Xi[\Gamma[\Delta]]\vdash D}
\end{tabular}
\end{center}
\caption{The sequent calculus rules for \nllam{}: structural rules}
\label{tab:seqnllamsr}	
\end{table}

Tables~\ref{tab:seqnllam} and~\ref{tab:seqnllamsr} show the sequent calculus rules for \nllam{}, with the logical rules in Table~\ref{tab:seqnllam} and the structural rules in Table~\ref{tab:seqnllamsr}.

As a simple example, we assign ``John'' the formula $np$ (that is, ``John'' functions as a noun phrase), and ``saw'' the formula $(np\ldl s)\ldr np$ (that is, ``saw'' combines first with a noun phrase $np$ to its right, then with a noun phrase $np$ to its left to produce a sentence $s$).

It it illustrative to first show how to derive ``everyone saw John'' in \nlambek{} (that is, the non-associative Lambek calculus, without any structural rules). When we assign ``everyone'' the NL formula $s\ldr(np\ldl s) $ we obtain the following proof.
\[
 \infer[\ldr L]{s\ldr(np\ldl s) \lstr ( (np\ldl s)\ldr np \lstr np ) \vdash s}{
   \infer[\ldl R]{( (np\ldl s)\ldr np \lstr np)\vdash np\ldl s}{
      \infer[\ldr L]{np \lstr ( (np\ldl s)\ldr np \lstr np ) \vdash s}{
        \infer[\textit{Ax}]{np\vdash np\proofspace}{}
      & \infer[\ldl L]{np \lstr (np\ldl s)\vdash s}{
           \infer[\textit{Ax}]{np\vdash np\proofspace}{}
          &\infer[\textit{Ax}]{s\vdash s\proofspace}{}
        }
      }
    } & \infer[\textit{Ax}]{s\vdash s\proofspace}{}
} 
 \]
 We see that the $\ldr L$ and $\ldl R$ rules combine to replace the quantifier  formula $s\ldr(np\ldl s) $ by the formula $np$, and we complete the proof by having the transitive verb select its two arguments.  
 
The problem with the proof above is that it only works when the quantifier formula $s\ldr(np\ldl s) $ is the leftmost formula in the sentence in which it takes its scope. We would like a quantifier to have the possibility to take scope from anywhere in the sentence in which it occurs. When we assign ``everyone'' the \nllam{} formula $s\bdr(np\bdl s)$, we can derive ``John saw everyone'' as type $s$  as follows. 

\[
 \infer[\betared]{np \lstr ( (np\ldl s)\ldr np \lstr s\bdr(np\bdl s) ) \vdash s}{\infer[\bdr L]{s\bdr(np\bdl s) \bstr \lambda x.\, np \lstr ( (np\ldl s)\ldr np \lstr x)\vdash s}{
      \infer[\bdl R]{\lambda x.\, np \lstr ( (np\ldl s)\ldr np \lstr x)\vdash np\bdl s}{
          \infer[\betaexp]{np \bstr \lambda x.\, np \lstr ( (np\ldl s)\ldr np \lstr x)\vdash s}{\infer*{np \lstr ( (np\ldl s)\ldr np \lstr np)\vdash s}{}
          }
      } 
    & \infer[\textit{Ax}]{s\vdash s\proofspace}{}}} 
 \]
Comparing this proof to the previous one, we see that the $\betared$ rule allows us to move the quantifier to the leftmost position, with the continuation mode `$\bstr$' and with the $\lambda$ binder marking the original position of the quantifier. Then, the $\bdr L$ and $\bdl R$ rules combine to replace the quantifier by an $np$ formula (as the $\ldr L$ and $\ldl R$ rules did in the NL proof), and finally, the $\betaexp$ rule moves this $np$ back to the position of the original quantifier. The proof then continues as before, with the transitive verb selecting its two arguments.

\newcommand{\very}{(n \ldr n) \ldr (n\ldr n)}

\nllam{} has a unit `$\lidc$', which is a 0-ary logical connective corresponding to the identity element `$\lidstr$' for the Lambek mode. The structural rules for the unit show that `$\lidstr$' functions as a two-sided identity element for the Lambek structural connective `$\lstr$'. Lambek calculi are generally defined in a way which excludes empty antecedent derivations. The classic linguistic example against empty antecedent derivations in Lambek calculi is the standard analysis of phrases like ``very interesting book'', where ``book'' is assigned to formula $n$, ``interesting'' the formula $n\ldr n$, and ``very'' the formula $\very$. The problem with allowing empty antecedent derivations is that $\lidstr \vdash n\ldr n$ is derivable, and therefore we not only predict that ``very interesting book'' is grammatically a noun, but also ``very book'', by the following derivation. 
\[
\infer[\lstr\lidstr]{(\very)\lstr n \vdash n}{
   \infer[\ldr E]{((\very) \lstr \lidstr)\lstr n\vdash n}{
       \infer[\ldr I]{\lidstr \vdash n\ldr n}{
         \infer[\lidstr\lstr^{-1}]{\lidstr \lstr n\vdash n\proofspace}{
            \infer[\textit{Ax}]{n\vdash n\proofspace}{}
         }
       }
     & \infer[\ldr E]{n\ldr n \lstr n\vdash n\proofspace}{
         \infer[\textit{Ax}]{n\vdash n\proofspace}{}
       & \infer[\textit{Ax}]{n\vdash n\proofspace}{}
     }
   }
}
\] 
\citet[Section~16.6]{bs14cont} and \citet{barker2019nllam} allow empty antecedent derivations in their logic. However, it is possible to exclude empty antecedent derivations by simply removing the $\lstr\lidstr^{-1}$ and $\lidstr\lstr^{-1}$ structural rules from Table~\ref{tab:seqnllamsr}. This blocks the crucial subproof of $\lidstr \vdash n\ldr n$  in the proof above. In addition to being linguistically desirable, I will argue in Section~\ref{sec:deci} that removing these two structural rules has formal and computational advantages. However, the results for the proof net calculus in the next section and the decidability result of Section~\ref{sec:deci} hold  irrespective of the presence or absence of empty antecedent derivations. 

\nllam{} requires its end-sequents to contain only the \nlambek{} structural connective `$\lstr$'; the continuation mode `$\bstr$', variables, and abstracted antecedent structures can only appear as intermediate structures.
\section{Proof nets}
\label{sec:pn}

Proof nets can be seen either as a parallelised version of the sequent calculus, which abstracts away from inessential rule permutations, or as a multi-conclusion version of natural deduction. The main advantage of proof nets is that different proof nets corresponds to proofs which differ for interesting reasons. In the linguistic context, this means that different proof nets corresponds to different lambda terms, that is, different \emph{readings} of the sentence\footnote{There is a caveat here: while different proof nets will produce different \emph{derivational} semantics, after lexical substitution, some of these readings can become equivalent.}.

Although it is not hard to give an inductive definition of proof nets, we prefer to give the standard presentation using proof structures and a correctness condition. This way of presenting proof nets can easily be turned into an algorithm for proof search.

\begin{enumerate}
\item Unfold the formulas, essentially writing down the tree structure of the formula, until we reach the atomic formulas.
\item Connect atomic occurrences of hypotheses to conclusions, producing a proof structure.
\item Check correctness of the proof structure using a correctness condition. We will use a correctness condition in the form of graph rewriting.
\end{enumerate}

\subsection{Proof structures}

The basic building block of proof structures (and the \emph{abstract} proof structures introduced below) is the link. In a proof structure, a link connects a complex formula to its immediate subformulas.

\begin{definition} A \emph{link} is tuple consisting of a \emph{type} (tensor or par), a \emph{family} (indicating the family of connectives is belongs to), a list of premisses, a list of conclusions, and possibly a main node (either one of the conclusions or one of the premisses).
\end{definition}

A link is essentially a labelled hyperedge connecting a number of vertices in a hypergraph. The premisses of a link are drawn left-to-right above the central node, whereas the conclusions are draw left-to-right below the central node. A par link displays the central node as a filled circle, whereas a tensor uses an open circle. The non-associative Lambek calculus family of connectives ($\ldl$, $\lpr$, $\ldr$) uses a single circle for its links, whereas the continuation family ($\bdl$, $\bpr$, $\bdr$) uses links with double circles. For the binary connectives, this is a normal multimodal setup with two modes, where the modes are visually distinguished by single/double circles in the links rather than mode labels in the centre. The unit $\lidc$ is treated as a 0-ary connective instead of an atomic formula with special properties (this is rather standard in many logics, and in our case it simplifies the proof net calculus, see Appendix~\ref{sec:unitatom} for discussion).  Table~\ref{tab:lamlinks} shows the links for \nllam{} proof structures.

\begin{table}
\begin{center}
\begin{tikzpicture}[scale=0.75]
\node (t) at (8em,22.4em) {$\lidc$};
\node[par] (tpar) at (8em,20.0em) {};
\node (tparl) at (8em,20.0em) {\textcolor{white}{$\lidstr$}};
\draw[>=latex,->] (tpar) edge (t);
\node at (8em,17.6em) {$[\lidc L]$};
\node (t) at (18em,20em) {$\lidc$};
\node[tns] (ttns) at (18em,22.4em) {};
\node (ttnsl) at (18em,22.4em) {$\lidstr$};
\draw (t) -- (ttns);
\node at (18em,17.6em) {$[\lidc R]$};
\node (labl) at (3em,7.5em) {$[\ldr L]$};
\node (ab) at (3em,9.8em) {$C$};
\node (a) at (0,14.6em) {$C\ldr B$};
\node (aa) at (0.7em,14.2em) {};
\node (b) at (6em,14.75em) {$B$};
\node[tns] (c) at (3em,12.668em) {};
\draw (c) -- (ab);
\draw (c) -- (aa);
\draw (c) -- (b);
\node (labl) at (3em,-2.0em) {$[\ldr R]$};
\node (pa) at (0,0) {$C\ldr B$};
\node (pat) at (0.7em,0.44em) {};
\node[par] (pc) at (3em,1.732em) {};
\node (pb) at (6em,0.15em) {$B$};
\node (pd) at (3em,4.8em) {$C$};
\draw (pc) -- (pb);
\draw (pc) -- (pd);
\path[>=latex,->]  (pc) edge (pat);
\node (labl) at (23em,7.5em) {$[\ldl L]$};
\node (ab) at (23em,9.8em) {$C$};
\node (a) at (26em,14.6em) {$A\ldl C$};
\node (aa) at (25.3em,14.2em) {};
\node (b) at (20em,14.75em) {$A$};
\node[tns] (c) at (23em,12.668em) {};
\draw (c) -- (ab);
\draw (c) -- (aa);
\draw (c) -- (b);
\node (labl) at (23em,-2.0em) {$[\ldl R]$};
\node (pa) at (26em,0) {$A\ldl C$};
\node (pat) at (25.3em,0.44em) {};
\node[par] (pc) at (23em,1.732em) {};
\node (pb) at (20em,0.15em) {$A$};
\node (pd) at (23em,4.8em) {$C$};
\draw (pc) -- (pb);
\draw (pc) -- (pd);
\path[>=latex,->]  (pc) edge (pat);
\node (labl) at (13em,7.5em) {$[\lpr L]$};
\node (pa) at (10em,9.8em) {$A$};
\node (pdt) at (13em,14.3em) {};
\node[par] (pc) at (13em,11.532em) {};
\node (pb) at (16em,9.8em) {$B$};
\node (pd) at (13em,14.6em) {$A\lpr B$};
\draw (pc) -- (pb);
\draw (pc) -- (pa);
\path[>=latex,->]  (pc) edge (pdt);
\node (labl) at (13em,-2.0em) {$[\lpr R]$};
\node (ab) at (13em,0em) {$A\lpr B$};
\node (aba) at (13em,0.3em) {};
\node (a) at (16em,4.8em) {$B$};
\node (b) at (10em,4.8em) {$A$};
\node[tns] (c) at (13em,2.868em) {};
\draw (c) -- (aba);
\draw (c) -- (a);
\draw (c) -- (b);
\end{tikzpicture}

\begin{tikzpicture}[scale=0.75]
\node (labl) at (3em,7.5em) {$[\bdr L]$};
\node (ab) at (3em,9.8em) {$C$};
\node (a) at (0,14.6em) {$C\bdr B$};
\node (aa) at (0.7em,14.2em) {};
\node (b) at (6em,14.75em) {$B$};
\node[ttns] (c) at (3em,12.668em) {};
\draw (c) -- (ab);
\draw (c) -- (aa);
\draw (c) -- (b);
\node (labl) at (3em,-2.0em) {$[\bdr R]$};
\node (pa) at (0,0) {$C\bdr B$};
\node (pat) at (0.7em,0.44em) {};
\node[ppar] (pc) at (3em,1.732em) {};
\node (pb) at (6em,0.15em) {$B$};
\node (pd) at (3em,4.8em) {$C$};
\draw (pc) -- (pb);
\draw (pc) -- (pd);
\path[>=latex,->]  (pc) edge (pat);
\node (labl) at (23em,7.5em) {$[\bdl L]$};
\node (ab) at (23em,9.8em) {$C$};
\node (a) at (26em,14.6em) {$A\bdl C$};
\node (aa) at (25.3em,14.2em) {};
\node (b) at (20em,14.75em) {$A$};
\node[ttns] (c) at (23em,12.668em) {};
\draw (c) -- (ab);
\draw (c) -- (aa);
\draw (c) -- (b);
\node (labl) at (23em,-2.0em) {$[\bdl R]$};
\node (pa) at (26em,0) {$A\bdl C$};
\node (pat) at (25.3em,0.44em) {};
\node[ppar] (pc) at (23em,1.732em) {}; 
\node (pb) at (20em,0.15em) {$A$};
\node (pd) at (23em,4.8em) {$C$};
\draw (pc) -- (pb);
\draw (pc) -- (pd);
\path[>=latex,->]  (pc) edge (pat);
\node (labl) at (13em,7.5em) {$[\bpr L]$};
\node (pa) at (10em,9.8em) {$A$};
\node (pdt) at (13em,14.3em) {};
\node[ppar] (pc) at (13em,11.532em) {};
\node (pb) at (16em,9.8em) {$B$};
\node (pd) at (13em,14.6em) {$A\bpr B$};
\draw (pc) -- (pb);
\draw (pc) -- (pa);
\path[>=latex,->]  (pc) edge (pdt);
\node (labl) at (13em,-2.0em) {$[\bpr R]$};
\node (ab) at (13em,0em) {$A\bpr B$};
\node (aba) at (13em,0.3em) {};
\node (a) at (16em,4.8em) {$B$};
\node (b) at (10em,4.8em) {$A$};
\node[ttns] (c) at (13em,2.868em) {};
\draw (c) -- (aba);
\draw (c) -- (a);
\draw (c) -- (b);
\end{tikzpicture}
\end{center}
\caption{Links for \nllam{} proof structures}
\label{tab:lamlinks}
\end{table}

Each connective has two links: one for where it occurs as a hypothesis (its left link) and one for where it occurs as a conclusion (it right link). The left link and the right link for a connective are up-down symmetric and exactly one of them is a par link (with a filled center) and the other is a tensor link (with an empty center). 

From Table~\ref{tab:lamlinks} it is clear that the binary par links have one premiss and two conclusions, whereas the binary tensor links have two premisses and one conclusion (we will see a tensor link with one premiss and two conclusions later). Par links have an arrow pointing to the main formula of the link. Although the main formula of a tensor link can be determined from the formula labels, tensor links do not have a main node.

The basic idea is that tensor links \emph{build} structure, whereas par links \emph{remove} it. This is clear when we compare the links to the logical rules with which they share their label. For example, the $/L$ rule of the sequent calculus, read from top to bottom, creates as new structure with the connective `$\lstr$' whereas the $/R$ rule removes a structural connective `$\lstr$'. The same holds for the other connectives, with the tensor rule adding an occurrence of the structural connective of the corresponding family (`$\lstr$' or `$\bstr$') and the par rule removing one. 

\begin{definition} A proof structure is a tuple $\langle F, L\rangle$, where $F$ is a set of formula occurrences and $L$ is a set of the links shown in Table~\ref{tab:lamlinks} where each local neighbourhood respects the formulas shown in the table and such that: 
\begin{itemize}
\item each formula is at most once the premiss of a link	,
\item each formula is at most once the conclusion of a link.
\end{itemize}

The formulas which are not a conclusion of any link in a proof structure are its \emph{hypotheses}. The formulas which are not a premiss of any link in a proof structure are its \emph{conclusions}. Formulas which are both a premiss and a conclusion of a link are \emph{internal nodes} of the proof structure.	

We say a proof structure with hypotheses $\Gamma$ and conclusions $\Delta$ is a proof structure of $\Gamma \vdash \Delta$, overloading the $\vdash$ symbol.
\end{definition}

\begin{figure}
\begin{center}
\begin{tikzpicture}[scale=0.75]
\node (spar) at (13em,0em) {$s$};
\node (a) at (16em,6em) {$np\bdl s$};
\node (b) at (10em,4.8em) {$s\bdr(np\bdl s)$};
\node[ttns] (c) at (13em,2.668em) {};
\draw (c) -- (spar);
\draw (c) -- (a);
\draw (c) -- (b);
\node (d) at (10em,7em) {$np$};
\node [ppar] (pc) at (13em,8.868em) {};
\node (e) at (13em,11.5em) {$s$};
\path[>=latex,->]  (pc) edge (a);
\draw (pc) -- (d);
\draw (pc) -- (e);
\node (ab) at (3em,4.8em) {$np\ldl s$};
\node (a) at (0,9.4em) {$(np\ldl s)\ldr np$};
\node (aa) at (0.7em,9em) {};
\node (obj) at (6em,8.95em) {$np$};
\node[tns] (c) at (3em,7.486em) {};
\draw (c) -- (ab);
\draw (c) -- (aa);
\draw (c) -- (obj);
\node (subj) at (-3em,4.8em) {$np$};
\node [tns] (cc) at (0em,2.668em) {};
\node (stv) at (0em,0em) {$s$};
\draw (cc) -- (stv);
\draw (cc) -- (subj);
\draw (cc) -- (ab);
\node (np) at (-7em,0em) {$np$};
\node (goal) at (20em,0em) {$s$};
\end{tikzpicture}	
\end{center}
\caption{Formula unfolding for ``John saw everyone''.}
\label{fig:unfolda}
\end{figure}
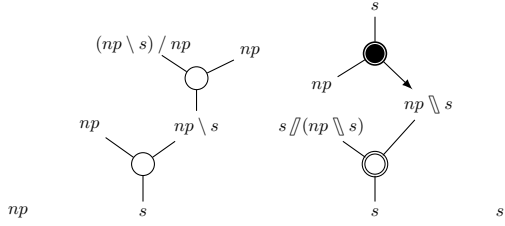

As an example, Figure~\ref{fig:unfolda} shows the formula unfolding of ``John saw everyone'', with ``John'' assigned $np$, ``saw'' $(np\ldl s)\ldr np$, and ``everyone'' $s\bdr(np\bdl s)$ with goal formula $s$. Given that this sentence is grammatical, we want to construct a proof net of $np, (np\ldl s)\ldr np, s\bdr(np\bdl s) \vdash s$, for some structured antecedent $\Gamma$ which has the formulas in the indicated left-to-right order. However, Figure~\ref{fig:unfolda} shows a proof structure with the following hypotheses and conclusions.
\[
 np, np, (np\ldl s)\ldr np, np, s\bdr(np\bdl s), s \vdash np, s, s, np, s 
 \]

\begin{figure}
\begin{center}
\begin{tikzpicture}[scale=0.75]
\node (spar) at (13em,0em) {$s$};
\node (a) at (16em,6em) {$np\bdl s$};
\node (b) at (10em,4.8em) {$s\bdr(np\bdl s)$};
\node[ttns] (c) at (13em,2.668em) {};
\draw (c) -- (spar);
\draw (c) -- (a);
\draw (c) -- (b);
\node (d) at (10em,7em) {$np$};
\node [ppar] (pc) at (13em,8.868em) {};
\node (e) at (13em,11.5em) {$s$};
\path[>=latex,->]  (pc) edge (a);
\draw (pc) -- (d);
\draw (pc) -- (e);
\node (ab) at (3em,4.8em) {$np\ldl s$};
\node (a) at (0,9.4em) {$(np\ldl s)\ldr np$};
\node (aa) at (0.7em,9em) {};
\node (obj) at (6em,8.95em) {$np$};
\node[tns] (c) at (3em,7.486em) {};
\draw (c) -- (ab);
\draw (c) -- (aa);
\draw (c) -- (obj);
\node (subj) at (-3em,4.8em) {$np$};
\node [tns] (cc) at (0em,2.668em) {};
\node (stv) at (0em,0em) {$s$};
\draw (cc) -- (stv);
\draw (cc) -- (subj);
\draw (cc) -- (ab);
\node (np) at (-7em,0em) {$np$};
\node (goal) at (20em,0em) {$s$};
\draw [semithick,gray] (np) to [out=270,in=90] (subj); 
\draw [semithick,gray] (d) to [out=270,in=90] (obj); 
\draw [semithick,gray] (stv) to [out=270,in=90] (e); 
\draw [semithick,gray] (spar) to [out=270,in=90] (goal); 
\end{tikzpicture}
\vspace{-3\baselineskip}	
\end{center}
\caption{Formula unfolding for ``John saw everyone'' with identifications of atomic formulas.}
\label{fig:unfoldb}
\end{figure}
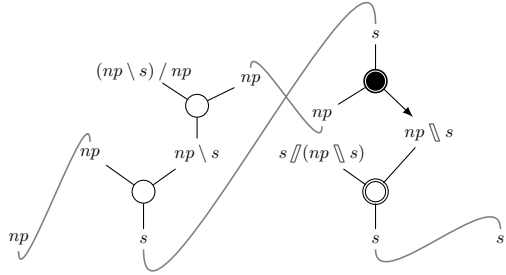

The problem with the proof structure of Figure~\ref{fig:unfolda} is that it has too many atomic formulas, both as hypotheses and as conclusions. For example, the leftmost isolated $np$, corresponding to the formula unfolding of ``John'', is --- as it should be --- a hypothesis of the proof structure. However, it is also a conclusion of the proof structure and our goal is to construct a proof structure with unique conclusion $s$ (the rightmost isolated $s$ node). Identifying the $np$ corresponding to ``John'' with the leftmost $np$ of the subgraph corresponding to ``saw'' will ``cancel out'' one premiss against a conclusion atomic formula (the rightmost $np$ of ``saw'' would also be a possibility, but would produce sentences containing ``saw John'' rather than ``John saw''). Figure~\ref{fig:unfoldb} shows one possibility of connecting the atomic formulas to produce a proof structure of the sequent we are trying to prove.
\[
 np, (np\ldl s)\ldr np, s\bdr(np\bdl s) \vdash s 
 \]

There are four possible proof structures (we have two choices for the $np$ atoms and two for the $s$ atoms). It will turn out that two of these possibilities are proof nets, one of them --- the one following the connections shown in Figure~\ref{fig:unfoldb} --- corresponding to a proof of ``John saw everyone'' and the other to a proof of ``Everyone saw John''.

\begin{figure}
\begin{center}
\begin{tikzpicture}[scale=0.75]
\node (ab) at (13em,0em) {$s$};
\node (a) at (16em,6em) {$np\bdl s$};
\node (b) at (10em,4.8em) {$s\bdr(np\bdl s)$};
\node[ttns] (c) at (13em,2.668em) {};
\draw (c) -- (ab);
\draw (c) -- (a);
\draw (c) -- (b);
%
\node [ppar] (pc) at (13em,8.868em) {};
\node (e) at (13em,11.5em) {$s$};
\path[>=latex,->]  (pc) edge (a);
\draw (pc) -- (e);
\node (ab) at (16em,16.3em) {$np\ldl s$};
\node (a) at (13em,20.9em) {$(np\ldl s)\ldr np$};
\node (aa) at (13.7em,20.5em) {};
\node (b) at (19em,20.45em) {$np$};
\node[tns] (c) at (16em,18.986em) {};
\draw (c) -- (ab);
\draw (c) -- (aa);
\draw (c) -- (b);
\node (subj) at (10em,16.3em) {$np$};
\node [tns] (cc) at (13em,14.168em) {};
\node (s) at (13em,11.5em) {};
\draw (cc) -- (s);
\draw (cc) -- (subj);
\draw (cc) -- (ab);
\draw (pc)..controls(5em,5em)and(5em,30em)..(b);
\node (ab) at (33em,0em) {$\apsnode{}{s}$};
\node (abb) at (33em,0em) {\emptynode};
\node (a) at (36em,6em) {$\apsnodei$};
\node (b) at (30em,4.8em) {$\apsnode{s\bdr(np\bdl s)}{}$};
\node (bx) at (30em,4.8em) {\emptynode};
\node[ttns] (c) at (33em,2.668em) {};
\draw (c) -- (abb);
\draw (c) -- (a);
\draw (c) -- (bx);
%
\node [ppar] (pc) at (33em,8.868em) {};
\node (e) at (33em,11.5em) {$\apsnodei$};
\path[>=latex,->]  (pc) edge (a);
\draw (pc) -- (e);
\node (ab) at (36em,16.3em) {$\apsnodei$};
\node (a) at (33em,20.9em) {$\apsnode{(np\ldl s)\ldr np}{}$};
\node (ax) at (33em,20.9em) {\emptynode};
\node (aa) at (33.7em,20.5em) {};
\node (b) at (39em,20.45em) {$\apsnodei$};
\node[tns] (c) at (36em,18.986em) {};
\draw (c) -- (ab);
\draw (c) -- (ax);
\draw (c) -- (b);
\node (subj) at (30em,16.3em) {$\apsnode{np}{}$};
\node (subjx) at (30em,16.3em) {\emptynode};
\node [tns] (cc) at (33em,14.168em) {};
\node (s) at (33em,11.5em) {};
\draw (cc) -- (s);
\draw (cc) -- (subjx);
\draw (cc) -- (ab);
\draw (pc)..controls(25em,5em)and(25em,30em)..(b);
\end{tikzpicture}	
\end{center}
\caption{Proof structure (left) and abstract proof structure (right) for ``John saw everyone''.}
\label{fig:pstoaps}
\end{figure}
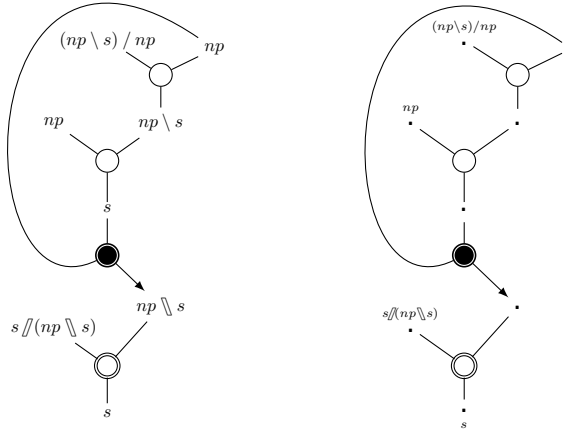

Figure~\ref{fig:pstoaps} shows (on the left hand side) the proof structure after node identifications. The $np$ conclusion of the $\bdl I$ link is drawn with an arc to arrive at the object  $np$ of the transitive verb. Essentially, this is a visual representation of the coindexation used for the introduction rules for the implications in natural deduction. However, we still need a mechanism to ensure the $\bdl R$ rule (corresponding to the par link) has been correctly applied. The premiss of this particular instance of the $\bdl R$ rule requires a proof of $np \bstr \Gamma \vdash s$ (for some $\Gamma$; this is just the $\bdl R$ rule of Table~\ref{tab:seqnllam} for $A=np$ and $C=s$, and where the $np$ in this structure is the one which is the leftmost conclusion of the par link), and we are not in the required structure.

Checking correctness will therefore require us to add two types of operations to our structures:
\begin{enumerate}
	\item  operations which correspond to the removal of a structural connective (tensor link) by a par link, exactly like the par rules of the sequent calculus,
	\item  operations which correspond to changing the structure of a connected group of tensor links, exactly like the structural rules of the sequent calculus change the antecedent.
\end{enumerate}



\subsection{Abstract proof structures}

\editout{
\begin{definition}
A \emph{tensor graph} is a connected proof structure with a unique conclusion (root) node containing only tensor links. The trivial tensor graph is a single node. 

Given a proof structure $P$, the \emph{components} of $P$ are the maximal substructures of $P$ which are tensor graphs. A \emph{tensor tree} is an acyclic tensor graph.
\end{definition}

For standard multimodal proof nets, we define correctness using tensor trees instead of the more general notion used here. Ultimately, we still want to end up with trees, although our intermediate structures may be slightly more general.
}

\begin{table}
\begin{center}
\begin{tikzpicture}[scale=0.75]
\node[tns] (empty) at (8em,12.4em) {};
\node at (8em,12.4em) {$\lidstr$};
\node (ebot) at (8em,10em) {$\apsnodei$};
\draw (ebot) -- (empty);
\node at (8em,8em) {$[\lidstr]$};
\node[par] (tp) at (18em,10em) {};
\node (tpl) at (18em,10em) {\textcolor{white}{$\lidstr$}};
\node (pn) at (18em,12.4em) {$\apsnodei$};
\path[>=latex,->]  (tp) edge (pn);
\node at (18em,8em) {$[tL]$};
\node (labl) at (23em,-2.0em) {$[\lambda]$};
\node (pa) at (26em,0) {$\apsnodei$};
\node[tns] (pc) at (23em,1.732em) {};
\node (pcl) at (23em,1.732em) {$\lambda$};
\node (pb) at (20em,0.15em) {$\apsnodei$};
\node (pd) at (23em,4.8em) {$\apsnodei$};
\draw (pc) -- (pb);
\draw (pc) -- (pd);
\draw (pc) -- (pa);
\node (labl) at (3em,-2.0em) {$[\lstr]$};
\node (ab) at (3em,0em) {$\apsnodei$};
\node (aba) at (3em,0.3em) {};
\node (a) at (6em,4.8em) {$\apsnodei$};
\node (b) at (0em,4.8em) {$\apsnodei$};
\node[tns] (c) at (3em,2.868em) {};
\draw (c) -- (aba);
\draw (c) -- (a);
\draw (c) -- (b);
\node (labl) at (13em,-2.0em) {$[\bstr]$};
\node (ab) at (13em,0em) {$\apsnodei$};
\node (aba) at (13em,0.3em) {};
\node (a) at (16em,4.8em) {$\apsnodei$};
\node (b) at (10em,4.8em) {$\apsnodei$};
\node[ttns] (c) at (13em,2.868em) {};
\draw (c) -- (aba);
\draw (c) -- (a);
\draw (c) -- (b);
\end{tikzpicture}

\begin{tikzpicture}[scale=0.75]
\node (labl) at (3em,-2.0em) {$[\ldr R]$};
\node (pa) at (0,0) {$\apsnodei$};
\node[par] (pc) at (3em,1.732em) {};
\node (pb) at (6em,0.15em) {$\apsnodei$};
\node (pd) at (3em,4.8em) {$\apsnodei$};
\draw (pc) -- (pb);
\draw (pc) -- (pd);
\path[>=latex,->]  (pc) edge (pa);
\node (labl) at (23em,-2.0em) {$[\ldl R]$};
\node (pa) at (26em,0) {$\apsnodei$};
\node[par] (pc) at (23em,1.732em) {}; 
\node (pb) at (20em,0em) {$\apsnodei$};
\node (pd) at (23em,4.8em) {$\apsnodei$};
\draw (pc) -- (pb);
\draw (pc) -- (pd);
\path[>=latex,->]  (pc) edge (pa);
\node (labl) at (13em,-2.0em) {$[\lpr L]$};
\node (pa) at (10em,0em) {$\apsnodei$};
\node[par] (pc) at (13em,1.732em) {};
\node (pb) at (16em,0em) {$\apsnodei$};
\node (pd) at (13em,4.8em) {$\apsnodei$};
\draw (pc) -- (pb);
\draw (pc) -- (pa);
\path[>=latex,->]  (pc) edge (pd);
\end{tikzpicture}

\begin{tikzpicture}[scale=0.75]
\node (labl) at (3em,-2.0em) {$[\bdr R]$};
\node (pa) at (0,0) {$\apsnodei$};
\node[ppar] (pc) at (3em,1.732em) {};
\node (pb) at (6em,0em) {$\apsnodei$};
\node (pd) at (3em,4.8em) {$\apsnodei$};
\draw (pc) -- (pb);
\draw (pc) -- (pd);
\path[>=latex,->]  (pc) edge (pa);
\node (labl) at (23em,-2.0em) {$[\bdl R]$};
\node (pa) at (26em,0) {$\apsnodei$};
\node[ppar] (pc) at (23em,1.732em) {}; 
\node (pb) at (20em,0.15em) {$\apsnodei$};
\node (pd) at (23em,4.8em) {$\apsnodei$};
\draw (pc) -- (pb);
\draw (pc) -- (pd);
\path[>=latex,->]  (pc) edge (pa);
\node (labl) at (13em,-2.0em) {$[\bpr L]$};
\node (pa) at (10em,0em) {$\apsnodei$};
\node[ppar] (pc) at (13em,1.732em) {};
\node (pb) at (16em,0em) {$\apsnodei$};
\node (pd) at (13em,4.8em) {$\apsnodei$};
\draw (pc) -- (pb);
\draw (pc) -- (pa);
\path[>=latex,->]  (pc) edge (pd);
\end{tikzpicture}

\end{center}
\caption{Links for \nllam{} abstract proof structures}
\label{tab:apslinks}
\end{table}

Given a proof structure, we obtain an abstract proof structure by erasing the formulas for all internal nodes of the proof structure: only the hypotheses and the conclusions of an abstract proof structure are assigned a formula. Given a proof  structure $\Pi$, we denote the underlying abstract proof structure by $\mathcal{A}(\Pi)$.

\begin{definition}
An abstract proof structure $A$ is a tuple $\langle V, L, h, c\rangle$ where $V$ is a set of vertices, $L$ is a set of the links shown in Table~\ref{tab:apslinks} connecting the vertices of $V$, $h$ is a function from the hypothesis vertices of $A$ to formulas, and $c$ is a function from the conclusion vertices of $A$ to formulas (a hypothesis vertex is a vertex which is not the conclusion of any link in $L$,  and a conclusion vertex is a vertex which is not the premiss of any link in $L$). 	
\end{definition}

The links for abstract proof structures are shown in Table~\ref{tab:apslinks}. The the top row shows the links for the 0-ary connective $t$, the second row shows the binary tensor links, the third row shows the par links for the Lambek connectives, and the bottom row shows the par links for the continuation connectives.


The $\lambda$ tensor link is the only non-standard link. Even though it has the same shape as the link for the Grishin  connectives of \citet{mm12pnlg}, it is used in a rather different way. The $\lambda$ tensor link does not correspond to a logical connective in \nllam{}. Rather, it corresponds to the $\lambda$ constructor in \nllam{} antecedent terms. As in multimodal proof nets, where tensor trees correspond to sequents, here tensor graphs correspond to sequents. 


Figure~\ref{fig:pstoaps} shows how the proof structure on the left hand side is transformed into an abstract proof structure on the right hand side. The procedure consists simply of removing all formulas on the internal nodes. 
Vertices which are hypotheses of the abstract proof structure have the corresponding formula written above the vertex, whereas vertices which are conclusions have the corresponding formula written below them. 

\begin{definition}
We say a proof structure, an abstract proof structure or one of their substructures, is a \emph{tensor tree} iff it is a tree containing only tensor links.

We say a substructure of a proof structure or of an abstract proof structure is a \emph{component} iff it is a maximal, connected substructure containing only tensor links. 

An abstract proof	structure or one of its substructures is, is a \emph{tensor graph} iff it is a connected graph of tensor links such that:
\begin{enumerate}
	\item\label{it:1} removing the connection between all $\lambda$ links and their leftmost conclusions produces a tree (we call this the \emph{underlying tree} of a tensor graph),
	\item for each $\lambda$ link in the graph, the leftmost conclusion $l$ of this $\lambda$ link is an ancestor of the premiss of the link in this graph.  
\end{enumerate}

We call a connected graph of tensor links \emph{cyclic} (resp.\ \emph{disconnected}), if removing the between the lambda link and their leftmost conclusion according to item~\ref{it:1} produces a cyclic (resp.\ disconnected) structure.
\end{definition}

Tensor graphs are graphical representations of the antecedent structure in \nllam{}. The conditions on tensor graphs ensure that each $\lambda$ binder in the abstract proof structure binds exactly one hypothesis in its scope. 

\subsection{Contractions and structural coversions}

To check correctness of a proof structure, we have two types of graph rewrite operations on its underlying abstract proof structure. The first are the contractions, shown in Table~\ref{tab:contr}. Each contraction has the condition that the vertex labeled $h$ (which is possibly a hypothesis of the abstract proof structure) and the vertex labeled $c$ (which is possibly a conclusion of the abstract proof structure) are distinct. The contractions replace the displayed pair of links by a single vertex (deleting the two links and any internal nodes, while identifying the nodes $h$ and $c$). 

The contractions are essentially a way of verifying the antecedent is in the right configuration for the application of the corresponding sequent rule. In this way the par link for $\bdl R$ and the contraction for $\bdl R$ work together to emulate the $\bdl R$ rule of the sequent calculus.

\begin{table}
\begin{center}
\begin{tikzpicture}[scale=0.75]
\node (labm) at (33em,-2.5em) {$[\lidc L ]$};
\node (tab) at (33em,0.0em) {$\apsnode{}{c}$};
\node (tabb) at (33em,0.0em) {\emptynode};
\node[tns] (tc) at (33em,2.868em) {};
\node (tcl) at (33em,2.868em) {$\lidstr$};
\draw (tc) -- (tabb);
\node (tabx) at (33em,9.6em) {$\apsnode{h}{}$};
\node (tabbx) at (33em,9.6em) {\emptynode};
\node[par] (tcx) at (33em,6.532em) {};
\node (tcxl) at (33em,6.532em) {\textcolor{white}{$\lidstr$}};
\path[>=latex,->]  (tcx) edge (tabbx);
\node (labl) at (3em,-2.5em) {$[\ldr R]$};
\node (ab) at (3em,4.8em) {$\apsnodei$};
\node (a) at (0,9.6em) {$\apsnode{h}{}$};
\node (b) at (6em,9.6em) {$\apsnodei$};
\node[tns] (c) at (3em,7.668em) {};
\draw (c) -- (ab);
\draw (c) -- (a);
\draw (c) -- (b);
\node (pa) at (0,0) {$\apsnode{}{c}$};
\node[par] (pc) at (3em,1.732em) {};
\draw (pc) -- (ab);
\path[>=latex,->]  (pc) edge (pa);
\draw (b) to [out=50,in=330] (pc);
\node (labm) at (13em,-2.5em) {$[\lpr L ]$};
\node (tab) at (13em,0.0em) {$\apsnode{}{c}$};
\node (tabb) at (13em,0.0em) {\emptynode};
\node (ta) at (10em,4.8em) {$\apsnodei$};
\node (tb) at (16em,4.8em) {$\apsnodei$};
\node[tns] (tc) at (13em,2.868em) {};
\draw (tc) -- (tabb);
\draw (tc) -- (ta);
\draw (tc) -- (tb);
\node (tabx) at (13em,9.6em) {$\apsnode{h}{}$};
\node (tabbx) at (13em,9.6em) {\emptynode};
\node[par] (tcx) at (13em,6.532em) {};
\path[>=latex,->]  (tcx) edge (tabbx);
\draw (tcx) -- (ta);
\draw (tcx) -- (tb);
\node (labr) at (23em,-2.5em) {$[\ldl R ]$};
\node (ab) at (23em,4.8em) {$\apsnodei$};
\node (a) at (26em,9.6em) {$\apsnode{h}{}$};
\node (b) at (20em,9.6em) {$\apsnodei$};
\node[tns] (c) at (23em,7.668em) {};
\draw (c) -- (ab);
\draw (c) -- (a);
\draw (c) -- (b);
\node (pa) at (26em,0) {$\apsnode{}{c}$};
\node[par] (pc) at (23em,1.732em) {};
\draw (pc) -- (ab);
\path[>=latex,->]  (pc) edge (pa);
\draw (b) to [out=130,in=210] (pc);
\end{tikzpicture}
\begin{tikzpicture}[scale=0.75]
\node (labl) at (3em,-2.5em) {$[\bdr R]$};
\node (ab) at (3em,4.8em) {$\apsnodei$};
\node (a) at (0,9.6em) {$\apsnode{h}{}$};
\node (b) at (6em,9.6em) {$\apsnodei$};
\node[ttns] (c) at (3em,7.668em) {};
\draw (c) -- (ab);
\draw (c) -- (a);
\draw (c) -- (b);
\node (pa) at (0,0) {$\apsnode{}{c}$};
\node[ppar] (pc) at (3em,1.732em) {};
\draw (pc) -- (ab);
\path[>=latex,->]  (pc) edge (pa);
\draw (b) to [out=50,in=330] (pc);
\node (labm) at (13em,-2.5em) {$[\bpr L ]$};
\node (tab) at (13em,0.0em) {$\apsnode{}{c}$};
\node (tabb) at (13em,0.0em) {\emptynode};
\node (ta) at (10em,4.8em) {$\apsnodei$};
\node (tb) at (16em,4.8em) {$\apsnodei$};
\node[ttns] (tc) at (13em,2.868em) {};
\draw (tc) -- (tabb);
\draw (tc) -- (ta);
\draw (tc) -- (tb);
\node (tabx) at (13em,9.6em) {$\apsnode{h}{}$};
\node (tabbx) at (13em,9.6em) {\emptynode};
\node[ppar] (tcx) at (13em,6.532em) {};
\path[>=latex,->]  (tcx) edge (tabbx);
\draw (tcx) -- (ta);
\draw (tcx) -- (tb);
\node (labr) at (23em,-2.5em) {$[\bdl R ]$};
\node (ab) at (23em,4.8em) {$\apsnodei$};
\node (a) at (26em,9.6em) {$\apsnode{h}{}$};
\node (b) at (20em,9.6em) {$\apsnodei$};
\node[ttns] (c) at (23em,7.668em) {};
\draw (c) -- (ab);
\draw (c) -- (a);
\draw (c) -- (b);
\node (pa) at (26em,0) {$\apsnode{}{c}$};
\node[ppar] (pc) at (23em,1.732em) {};
\draw (pc) -- (ab);
\path[>=latex,->]  (pc) edge (pa);
\draw (b) to [out=130,in=210] (pc);
\end{tikzpicture}
\end{center}
\caption{Contractions for \nllam.}
\label{tab:contr}
\end{table}

Graphically, we can see that the configurations which allow us to perform a contraction all connect a par link to a tensor link (respecting the left/right distinction) by the tentacles of the par link which do not have the arrow. Each contraction removes the two links (and the two internal nodes) while performing a vertex contraction on the two external vertices.

The contraction for $[\lidc L]$ looks a bit strange, but it is essentially the $[\lpr L]$ contraction with the two branches removed ($\lidc$ is a 0-ary connective, whereas $\lpr$ is a binary one). It therefore creates a new connection while identifying $h$ and $c$. 



\begin{figure}
\begin{center}
\begin{tikzpicture}[scale=0.75]
\node[pn] at (33em,-11em) {$\Gamma$}; 
\node at (33em,-8em) {$x$};
\node at (33em,-14em) {$z$}; 
\node at (25em,-11.5em) {$\longleftarrow$};
\node at (25em,-10.5em) {$\longrightarrow$};
\node at (25em,-12.5em) {$\betaexp$};
\node at (25em,-9.5em) {$\betared$};
%
\node (lt) at (13em,-13.5em) {$x$}; 
\node[ttns] (tc) at (16em,-15.232em) {}; 
\node (rt) at (19em,-13.5em) {$\apsnodei$}; 
\node (top) at (16em,-18.3em) {$z$}; 
\draw (lt) -- (tc);
\draw (rt) -- (tc);
\draw (tc) -- (top);
\node[tns] (lam) at (16em,-10.232em) {}; 
\node (lamlab) at (16em,-10.232em) {$\lambda$}; 
\node (bl) at (16em,-7.5em) {$\centerdot$}; 
\draw (bl) -- (lam);
\draw (lam) -- (rt);
\node[pn] at (16.5em,-5em) {$\Gamma$}; 
\node (bot) at (16em,-2.2em) {$\centerdot$}; 
\draw (bot) to [out=130,in=230] (lam); 
\end{tikzpicture}
\end{center}
\caption{Structural rule for \nllam, `sugared' version.}
\label{fig:srlamsugared}
\end{figure}
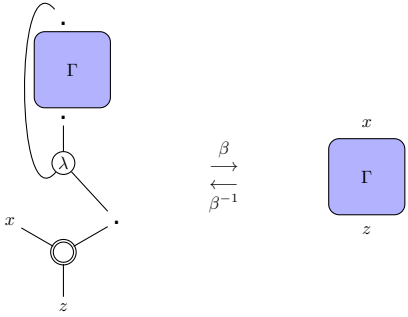

\begin{table}
\begin{center}
\begin{tikzpicture}[scale=0.75]
\node at (32em,-9.5em) {$\apsnode{h_1}{c_1}$};
\node at (32em,-15.5em) {$\apsnode{h_2}{c_2}$}; 
\node at (25em,-14em) {$\betaexp$};
\node at (25em,-11em) {$\betared$};
\node at (25.0em,-13.0em) {$\longleftarrow$}; 
\node at (25.0em,-12.0em) {$\longrightarrow$}; 
\node (lt) at (13em,-13.5em) {$\apsnode{h_1}{}$}; 
\node[ttns] (tc) at (16em,-15.232em) {}; 
\node (rt) at (19em,-12.5em) {$\apsnode{}{}$}; 
\node (top) at (16em,-18.3em) {$\apsnode{}{c_2}$}; 
\node (topp) at (16em,-18.3em) {\emptynode};
\draw (lt) -- (tc);
\draw (rt) -- (tc);
\draw (tc) -- (topp);
\node[tns] (lam) at (16em,-9.232em) {}; 
\node (lamlab) at (16em,-9.232em) {$\lambda$}; 
\node (bl) at (16em,-6.7em) {$\apsnode{h_2}{}$};
\node (bll) at (16em,-6.7em) {\phantom{M}}; 
\draw (bll) -- (lam);
\draw (lam) -- (rt);
\node (bot) at (13em,-10.5em) {$\apsnode{}{c_1}$};
\draw (lam) -- (bot); 
\node (ab) at (16em,-2.0em) {$\apsnode{}{c}$};
\node (abb) at (16em,-2.0em) {\emptynode};
\node (a) at (13em,2.8em) {$\apsnodei$};
\node (b) at (19em,2.8em) {$\apsnode{h}{}$};
\node[tns] (c) at (16em,0.868em) {};
\draw (c) -- (abb);
\draw (c) -- (a);
\draw (c) -- (b);
\node[tns] (e) at (13em,5.2em) {};
\node at (13em,5.2em) {\lidstr};
\draw (e) -- (a);
\node at (32em,1em) {$\apsnode{h}{c}$};
\node at (25em,-0.5em) {$\lidstr\lstr^{-1}$};
\node at (25em,2.5em) {$\lidstr\lstr$};
\node at (25.0em,0.5em) {$\longleftarrow$}; 
\node at (25.0em,1.5em) {$\longrightarrow$}; 
\node (ab) at (16em,10.0em) {$\apsnode{}{c}$};
\node (abb) at (16em,10.0em) {\emptynode};
\node (a) at (19em,14.8em) {$\apsnodei$};
\node (b) at (13em,14.8em) {$\apsnode{h}{}$};
\node[tns] (c) at (16em,12.868em) {};
\draw (c) -- (abb);
\draw (c) -- (a);
\draw (c) -- (b);
\node[tns] (e) at (19em,17.2em) {};
\node at (19em,17.2em) {\lidstr};
\draw (e) -- (a);
\node at (32em,13em) {$\apsnode{h}{c}$};
\node at (25em,11.5em) {$\lstr\lidstr^{-1}$};
\node at (25em,14.5em) {$\lstr\lidstr$};
\node at (25.0em,12.5em) {$\longleftarrow$}; 
\node at (25.0em,13.5em) {$\longrightarrow$}; 
\end{tikzpicture}
\end{center}
\caption{Structural rules for \nllam.}
\label{tab:srlam}
\end{table}

Figure~\ref{fig:srlamsugared} shows a `sugared' version of the conversion for the $\betared$ and $\betaexp$ rules for abstract proof structures. The left hand side is a graphical representation of $x \bstr \lambda v. \Gamma[v]$ and the right hand side of $\Gamma[x]$, where $x$ can represent any complex structure $\Delta$. Like the corresponding structural rules,  $\betaexp$ moves the constituent $x$ out and marks its position using a lambda binder, whereas $\betared$ moves $x$ back to the placed marked by the abstracted variable.
 The two rewrites have the side condition that the path through $\Gamma$ (from $x$ to $z$ on the left hand side and from the anonymous nodes at the top and bottom of $\Gamma$ at the right hand side) does not pas through any par links. 
Table~\ref{tab:srlam} shows the full set of structural rules. The side conditions on the $\betared$ and $\betaexp$ rules are that the node labeled $c_1$ is a descendant of the node labeled $h_2$ through a path not passing through any par links --- the side condition can be seen as a way of avoiding `accidental capture' of variables by the lambda binder, and it guarantees that all rewrites occur in a single component (essential for correctness). 


\begin{definition}\label{def:size} The \emph{size} of an \nllam{} proof structure or abstract proof structure with $p$ par links and $t$ tensor links is equal to $2p + t$.
\end{definition}

\begin{definition}\label{def:exp}\label{def:contr}
We say a graph rewrite is \emph{expanding} whenever the size of right hand side	of the rule is larger than the size of the left hand side, and \emph{reducing} whenever the size of the right hand side of the rule is smaller than the size of the left hand side.
\end{definition}

The contactions of Table~\ref{tab:contr} are all reducing (they all reduce the size of the structure by 3, removing one par link and one tensor link). The structural rewrites of Table~\ref{tab:srlam} are reducing in their left to right application but expanding in the right to left application. The reducing structural rules are also the rewrites which shorten some of the paths in the abstract proof structure.

\subsection{Proof nets}

\begin{definition}\label{def:pn} A proof structure is a \emph{proof net} iff its abstract proof structure contracts to a tensor graph using the contractions of Table~\ref{tab:contr} and the structural conversions of Table~\ref{tab:srlam}.
\end{definition}

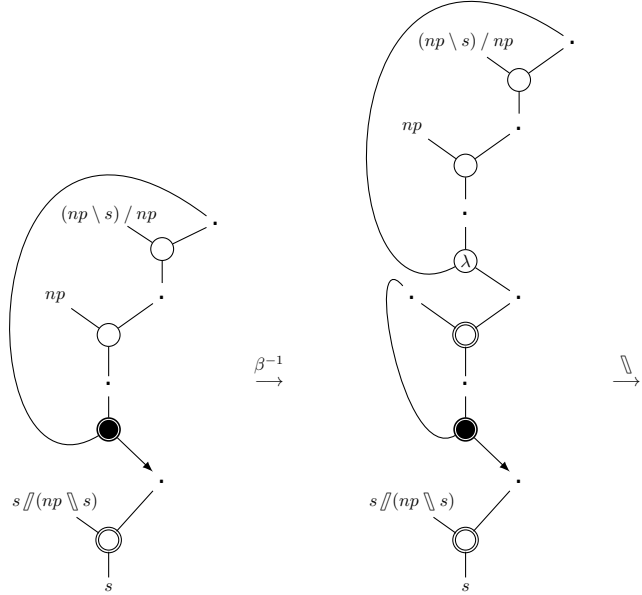
\begin{figure}
\begin{center}
\begin{tikzpicture}[scale=0.75]
\node (ab) at (3em,0em) {$s$};
\node (a) at (6em,6em) {$\apsnodei$};
\node (b) at (0em,4.8em) {$s\bdr(np\bdl s)$};
\node[ttns] (c) at (3em,2.668em) {};
\draw (c) -- (ab);
\draw (c) -- (a);
\draw (c) -- (b);
%
\node [ppar] (pc) at (3em,8.868em) {};
\node (e) at (3em,11.5em) {$\apsnodei$};
\path[>=latex,->]  (pc) edge (a);
\draw (pc) -- (e);
\node (ab) at (6em,16.3em) {$\apsnodei$};
\node (a) at (3em,20.9em) {$(np\ldl s)\ldr np$};
\node (aa) at (3.7em,20.5em) {};
\node (b) at (9em,20.45em) {$\apsnodei$};
\node[tns] (c) at (6em,18.986em) {};
\draw (c) -- (ab);
\draw (c) -- (aa);
\draw (c) -- (b);
\node (subj) at (0em,16.3em) {$np$};
\node [tns] (cc) at (3em,14.168em) {};
\node (s) at (3em,11.5em) {};
\draw (cc) -- (s);
\draw (cc) -- (subj);
\draw (cc) -- (ab);
\draw (pc)..controls(-5em,4em)and(-5em,31em)..(b);
\node (ab) at (23em,0em) {$s$};
\node (a) at (26em,6em) {$\apsnodei$};
\node (b) at (20em,4.8em) {$s\bdr(np\bdl s)$};
\node[ttns] (c) at (23em,2.668em) {};
\draw (c) -- (ab);
\draw (c) -- (a);
\draw (c) -- (b);
\node [ppar] (pc) at (23em,8.868em) {};
\node (e) at (23em,11.5em) {$\apsnodei$};
\path[>=latex,->]  (pc) edge (a);
\draw (pc) -- (e);
\node (ab) at (26em,16.3em) {$\apsnodei$};
\node (subj) at (20em,16.3em) {$\apsnodei$};
\node [ttns] (cc) at (23em,14.168em) {};
\node (s) at (23em,11.5em) {};
\draw (cc) -- (s);
\draw (cc) -- (subj);
\draw (cc) -- (ab);
\draw (subj) to [out=130,in=210] (pc);
\node [tns] (lam) at (23em,18.312em) {};
\node (llam) at (23em,18.312em) {$\lambda$};
\node (ltop) at (23em,21em) {$\apsnodei$};
\draw (lam) -- (ab);
\draw (lam) -- (ltop);
\node[tns] (tt) at (23em,23.668em) {};
\draw (ltop) -- (tt);
\node (sub) at (20em,25.8em) {$np$};
\draw (sub) -- (tt);
\node (vp) at (26em,25.8em) {$\apsnodei$};
\draw (vp) -- (tt);
\node [tns] (ttt) at (26em,28.468em) {};
\draw (vp) -- (ttt);
\node (tv) at (23em,30.6em) {$(np\ldl s)\ldr np$};
\node (gap) at (29em,30.6em) {$\apsnodei$};
\draw (ttt) -- (gap);
\draw (ttt) -- (tv);
\draw (lam)..controls(15em,14em)and(15em,40em)..(gap);
\node (a1) at (12em,11.5em) {$\longrightarrow$};
\node (a1) at (12em,12.5em) {$\betaexp$};
\node (a2) at (32em,11.5em) {$\longrightarrow$};
\node (a2) at (32em,12.5em) {$\bdl$};
\end{tikzpicture}	
\end{center}
\caption{Abstract proof structure of Figure~\ref{fig:pstoaps} before and after the $\betaexp$ structural conversion.}
\label{fig:excont}
\end{figure}

\begin{figure}
\begin{center}
\begin{tikzpicture}[scale=0.75]
\node (ab) at (6em,16.3em) {$\apsnodei$};
\node (subj) at (0em,16.3em) {$s\bdr(np\bdl s)$};
\node [ttns] (cc) at (3em,14.168em) {};
\node (s) at (3em,11.5em) {$s$};
\draw (cc) -- (s);
\draw (cc) -- (subj);
\draw (cc) -- (ab);
\node [tns] (lam) at (3em,18.312em) {};
\node (llam) at (3em,18.312em) {$\lambda$};
\node (ltop) at (3em,21em) {$\apsnodei$};
\draw (lam) -- (ab);
\draw (lam) -- (ltop);
\node[tns] (tt) at (3em,23.668em) {};
\draw (ltop) -- (tt);
\node (sub) at (0em,25.8em) {$np$};
\draw (sub) -- (tt);
\node (vp) at (6em,25.8em) {$\apsnodei$};
\draw (vp) -- (tt);
\node [tns] (ttt) at (6em,28.468em) {};
\draw (vp) -- (ttt);
\node (tv) at (3em,30.6em) {$(np\ldl s)\ldr np$};
\node (gap) at (9em,30.6em) {$\apsnodei$};
\draw (ttt) -- (gap);
\draw (ttt) -- (tv);
\draw (lam)..controls(-5em,14em)and(-5em,40em)..(gap);
\node (ab) at (21em,16.3em) {$\apsnodei$};
\node (subj) at (15em,16.3em) {$np$};
\node [tns] (cc) at (18em,14.168em) {};
\node (s) at (18em,11.5em) {$s$};
\draw (cc) -- (s);
\draw (cc) -- (subj);
\draw (cc) -- (ab);
\node [tns] (c) at (21em,18.968em) {};
\node (tv) at (18em,21.1em) {$(np\ldl s)\ldr np$};
\node (q) at (24em,21.1em) {$s\bdr(np\bdl s)$};
\draw (c) -- (ab);
\draw (c) -- (tv);
\draw (c) -- (q);
\node (a1) at (11em,18.5em) {$\longrightarrow$};
\node (a1) at (11em,19.5em) {$\betared$};
\end{tikzpicture}	
\end{center}
\caption{Abstract proof structure of Figure~\ref{fig:excont} after the $\bdl$ contraction.}
\label{fig:continue}
\end{figure}
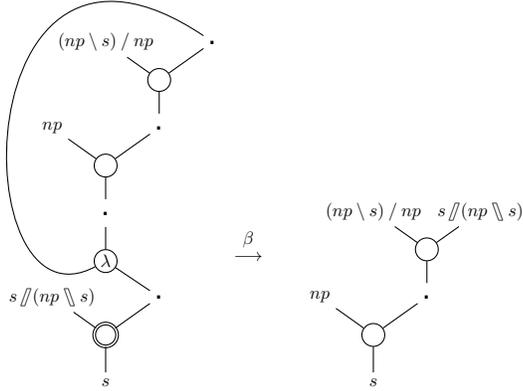

As an illustration, Figure~\ref{fig:excont} shows how we can take the abstract proof structure of Figure~\ref{fig:pstoaps}  and apply the $\betaexp$ rule to produce the abstract proof structure shown in Figure~\ref{fig:excont} on the right. This produces a redex for the $\bdl$ contraction. Figure~\ref{fig:continue} shows the result of applying the contraction and how applying the $\betared$ rule then produces the required tree structure. We have therefore shown that our example proof structure is a proof net, as it should be.

The end-sequents in \nllam{} are required to contain only the Lambek structural connective `$\lstr$'. 
 However, our intermediate structures can contain the `$\bstr$' connective and the $\lambda$ binder, and we therefore need to consider them for our inductive proofs.

\begin{theorem} $\Gamma \vdash C$ is derivable using the sequent calculus rules of Tables~\ref{tab:seqnllam} and~\ref{tab:seqnllamsr} if and only if there is a proof net contracting to the tensor graph $\Gamma\vdash C$.
\end{theorem}

\paragraph{Proof} This is a trivial adaptation of the proofs of \citet{mp} and \citet{msg19htlg}. 

\subparagraph{$\Longrightarrow$} From sequent proofs to proof nets, we proceed by induction on the depth of the proof. Only the axiom rule and the $\lidc R$ rule have depth 1.

 The axiom $A\vdash A$ corresponds to the single vertex proof structure $A$, shown below on the left, which produces the abstract proof structure of $A\vdash A$, shown below on the right.
\begin{center}
\begin{tikzpicture}[scale=0.75]
\node at (4em,4em) {$A$};
\node at (8em,3.8em) {$\longrightarrow$};
\node at (8em,4.8em) {$\mathcal{A}$};
\node at (12em,4em) {$\apsnode{A}{A}$};
\end{tikzpicture}	
\end{center}

For the $\lidc R$ rule, we unfold the conclusion formula $\lidc$ to obtain the proof structure shown below on the left,  and it corresponds immediately to the abstract proof structure of $1\vdash t$, as required. 
\begin{center}
\begin{tikzpicture}[scale=0.75]
\node at (4em,4em) {$t$};
\node (a) at (4em,4em) {\emptynode};
\node (b) [tns] at (4em,6.4em) {};
\node at (4em,6.4em) {$\lidstr$};
\draw (a) -- (b);	
\node at (8em,4.5em) {$\longrightarrow$};
\node at (8em,5.5em) {$\mathcal{A}$};
\node at (12em,4em) {$\apsnode{}{t}$};
\node (a) at (12em,4em) {\emptynode};
\node (b) [tns] at (12em,6.4em) {};
\node at (12em,6.4em) {$\lidstr$};
\draw (a) -- (b);	
\end{tikzpicture}	
\end{center}
For the inductive cases, we use the induction hypothesis to obtain proof nets (and rewrite sequences) for the immediate subproofs, add the appropriate link (for the logical rules of the connectives), extend the rewrite sequence with the appropriate conversion (a contraction whenever we add a par link, and a structural conversion for the structural rules). 

We show only the $\lidc L$, $\bdr R$ and $\betaexp$ cases, the other cases are similar.

\begin{description}
\item [$\lidc L$] For the $\lidc L$ rule, induction hypothesis gives us a proof net of $\Gamma[\lidstr]\vdash D$. That means we have a way to construct a proof structure $\Pi$ with the formulas of $\Gamma$ as its hypotheses, $D$ as its conclusion, and a rewrite sequence $\rho$ converting the abstract proof structure corresponding to $\Pi$ into $\Gamma[1] \vdash D$. Graphically, we are therefore in the following situation (where we tacitly convert proof structure $\Pi$ to its corresponding abstract proof structure $\mathcal{A}(\Pi)$ before applying the conversions in $\rho$).
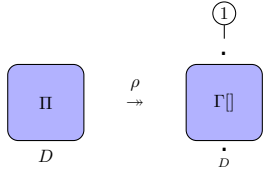
\begin{center}
\begin{tikzpicture}[scale=0.75]
\node[pn] at (10em,10em) {};
\node (p) at (10em,10em) {$\Pi$};	
\node (d) at (10em,7em) {$D$};
\node[pn] at (20em,10em) {};
\node (a) at (20em,10em) {$\Gamma[]$};
\node (dg) at (20em,12.8em) {$\apsnodei$};	
\node [tns] (t) at (20em,15em) {};
\node at (20em,15em) {$\lidstr$};
\draw (t) -- (dg);
\node (d) at (20em,7.2em) {$\apsnode{}{D}$};
\node at (15em,10em) {$\twoheadrightarrow$};
\node at (15em,11em) {$\rho$};
\end{tikzpicture}	
\end{center}

We need to produce a proof net of $\Gamma[\lidc]\vdash D$. The proof net therefore needs to have an additional $\lidc$ hypothesis, and in the final structure this hypothesis needs to be at the position where $\lidstr$ was in the proof net obtained by induction hypothesis.
We add a $\lidc$ as the hypothesis of a $\lidc L$ link. We then apply the rewrites in $\rho$ as before and end with the $\lidc L$ contraction to produce a proof net of $\Gamma[\lidc]\vdash D$ as follows.

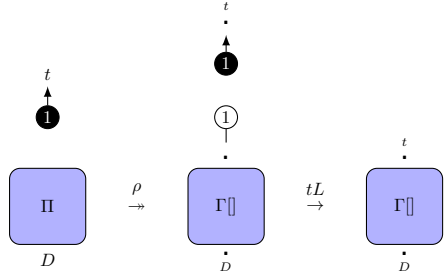
\begin{center}
\begin{tikzpicture}[scale=0.75]
\node[pn] at (10em,10em) {};
\node (p) at (10em,10em) {$\Pi$};	
\node (d) at (10em,7em) {$D$};
\node [par] (par) at (10em,15em) {};
\node at (10em,15em) {\textcolor{white}{$\lidstr$}};
\node (topt) at (10em,17.4em) {$\lidc$};
\path[>=latex,->]  (par) edge (topt);
\node[pn] at (20em,10em) {};
\node (a) at (20em,10em) {$\Gamma[]$};
\node (dg) at (20em,12.8em) {$\apsnodei$};	
\node [tns] (t) at (20em,15em) {};
\node at (20em,15em) {$\lidstr$};
\draw (t) -- (dg);
\node (d) at (20em,7.2em) {$\apsnode{}{D}$};
\node [par] (par) at (20em,18em) {};
\node at (20em,18em) {\textcolor{white}{$\lidstr$}};
\node (topt) at (20em,20.4em) {$\apsnode{\lidc}{}$};
\node (toptt) at (20em,20.4em) {\emptynode};
\path[>=latex,->]  (par) edge (toptt);
\node at (15em,10em) {$\twoheadrightarrow$};
\node at (15em,11em) {$\rho$};
\node[pn] at (30em,10em) {};
\node (a) at (30em,10em) {$\Gamma[]$};
\node (dg) at (30em,12.8em) {$\apsnode{t}{}$};	
\node (d) at (30em,7.2em) {$\apsnode{}{D}$};
\node at (25em,10em) {$\rightarrow$};
\node at (25em,11em) {$\lidc L$};
\end{tikzpicture}	
\end{center}


\item[$\bdr R$] The case for $\bdr R$ gives us a proof net of $\Gamma \bstr B \vdash C$ by induction hypothesis. The means we have the proof structure $\Pi$ shown below on the left and a conversion sequence $\rho$ producing the antecedent structure $\Gamma\bstr B$.
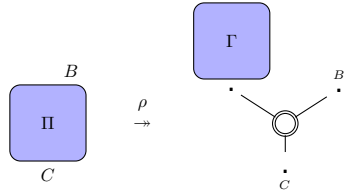
\begin{center}
\begin{tikzpicture}[scale=0.75]
\node[pn] at (-10.3em,7.7em) {$\Pi$}; 
\node (ab) at (-10.3em,4.8em) {$C$};
\node (b) at (-9em,10.6em) {$B$};
\node (ab) at (3em,4.8em) {$\apsnode{}{C}$};
\node (a) at (0,9.6em) {$\apsnodei$};
\node (b) at (6em,9.6em) {$\apsnode{B}{}$};
\node[ttns] (c) at (3em,7.668em) {};
\draw (c) -- (ab);
\draw (c) -- (a);
\draw (c) -- (b);
\node[pn] at (0.0em,12.3em) {$\Gamma$}; 
\node (labl) at (-5em,8.7em) {$\rho$}; 
\node  at (-5em,7.7em) {$\twoheadrightarrow$}; 
\end{tikzpicture}
\end{center}

To produce a proof net of $\Gamma \vdash C\bdr B$, we attach the par link for $\bdr R$ to the proof net given by induction hypothesis, and extend the conversion sequence $\rho$ with the $\bdr R$ contraction. This produces a proof net of $\Gamma \vdash C\bdr B$ as shown below.

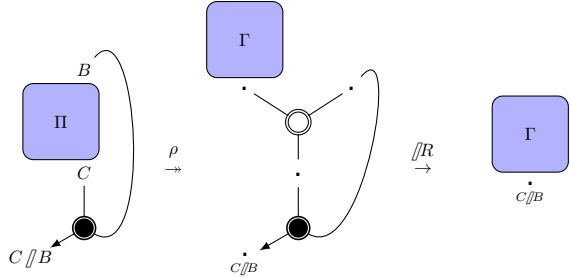
\begin{center}	
\begin{tikzpicture}[scale=0.75]
\node[pn] at (-10.3em,7.7em) {$\Pi$}; 
\node (ab) at (-9em,4.8em) {$C$};
\node (b) at (-9em,10.6em) {$B$};
\node (pa) at (-12em,0em) {$C\bdr B$};
\node (pad) at (-12em,0em) {\phantom{Mx}};
\node[ppar] (pc) at (-9em,1.732em) {};
\draw (pc) -- (ab);
\path[>=latex,->]  (pc) edge (pad);
\draw (b) to [out=50,in=330] (pc);
\node (ab) at (3em,4.8em) {$\apsnodei$};
\node (a) at (0,9.6em) {$\apsnodei$};
\node (b) at (6em,9.6em) {$\apsnodei$};
\node[ttns] (c) at (3em,7.668em) {};
\draw (c) -- (ab);
\draw (c) -- (a);
\draw (c) -- (b);
\node (pa) at (0,0) {$\apsnode{}{C\bdr B}$};
\node (pad) at (0em,0em) {\phantom{M}};
\node[ppar] (pc) at (3em,1.732em) {};
\draw (pc) -- (ab);
\path[>=latex,->]  (pc) edge (pad);
\draw (b) to [out=50,in=330] (pc);
\node[pn] at (0.0em,12.3em) {$\Gamma$}; 
\node[pn] at (16.0em,7.0em) {$\Gamma$}; 
\node at (16em,4.0em) {$\apsnode{}{C\bdr B}$}; 
\node (labl) at (-4em,6.0em) {$\rho$}; 
\node  at (-4em,5.0em) {$\twoheadrightarrow$}; 
\node (labl) at (10em,6.0em) {$\bdr R$}; 
\node  at (10em,5.0em) {$\rightarrow$}; 
%
\end{tikzpicture}
\end{center}
\item[$\betaexp$] When the last rule is the $\betaexp$ rule, induction hypothesis gives us a proof net of $\Xi[\Gamma[\Delta]] \vdash D$; in other words, a proof structure $\Pi$ and a rewrite sequence $\rho$ converting the corresponding abstract proof structure $\mathcal{A}(\Pi)$ into  $\Xi[\Gamma[\Delta]] \vdash D$. We can extend this conversion sequence with the $\betaexp$ conversion to produce a proof net of $\Xi[\Delta\bstr \lambda x. \Gamma[x]]\vdash D$. 
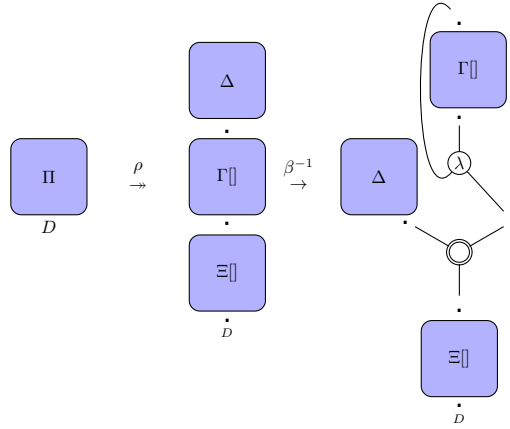
\begin{center}
\begin{tikzpicture}[scale=0.75]
\node[pn] at (-7em,-11em) {$\Pi$};
\node at (-7em,-13.8em) {$D$};
\node at (-2em,-11.5em) {$\twoheadrightarrow$};
\node at (-2em,-10.5em) {$\rho$};
\node[pn] at (3em,-5.6em) {$\Delta$}; 
\node[pn] at (3em,-11em) {$\Gamma[]$}; 
\node[pn] at (3em,-16.4em) {$\Xi[]$}; 
\node at (3em,-19.2em) {$\apsnode{}{D}$};
\node at (3em,-8.4em) {$\apsnodei$};
\node at (3em,-13.6em) {$\apsnodei$}; 
\node at (7em,-11.5em) {$\rightarrow$};
\node at (7em,-10.5em) {$\betaexp$};
%
\node[pn] at (11.5em,-11em) {$\Delta$};
\node (lt) at (13em,-13.5em) {$\apsnodei$}; 
\node[ttns] (tc) at (16em,-15.232em) {}; 
\node (rt) at (19em,-13.5em) {$\apsnodei$}; 
\node (top) at (16em,-18.3em) {$\centerdot$};
\node[pn] at (16em,-21.2em) {$\Xi[]$}; 
\node (d) at (16em,-24.0em) {$\apsnode{}{D}$};
\draw (lt) -- (tc);
\draw (rt) -- (tc);
\draw (tc) -- (top);
\node[tns] (lam) at (16em,-10.232em) {}; 
\node (lamlab) at (16em,-10.232em) {$\lambda$}; 
\node (bl) at (16em,-7.5em) {$\centerdot$}; 
\draw (bl) -- (lam);
\draw (lam) -- (rt);
\node[pn] at (16.5em,-5em) {$\Gamma[]$}; 
\node (bot) at (16em,-2.2em) {$\centerdot$}; 
\draw (bot) to [out=130,in=230] (lam); 
\end{tikzpicture}
\end{center}
\end{description}

\subparagraph{$\Longleftarrow$} The sequentialisation part of the proof, showing that for each proof net we can construct a corresponding sequent proof, is generally the difficult part of the proof. Given that we have a proof net, we know that we have a proof structure and a conversion sequence producing the abstract proof structure representing $\Gamma \vdash D$. 

We proceed first by induction on the length $l$ of the conversion sequence. If $l=0$, there are no conversions, and as a consequence there can be no par links and no lambda links (par links require a contraction to produce a tensor graph, whereas lambda links can only be produced by a structural conversion), our tensor graph must therefore be a tensor tree. 

We proceed by induction on the number of tensor links in the abstract proof structure. In the base case, $t=0$ there are no tensor links and the proof structure and abstract proof structure both consists of a single node, which is an $A$ hypothesis and an $A$ conclusion, corresponding to the axiom $A\vdash A$. 
For the inductive case, $t > 0$, take any tensor link in the tensor tree. We look at only one binary case (they are all similar) and at the 0-ary case.

In the case for the binary $[\bdl L]$, we have the following proof net, which is a tensor tree.
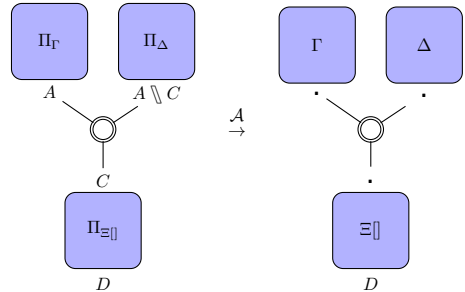
\begin{center}
\begin{tikzpicture}[scale=0.75]
\node (ab) at (23em,9.8em) {$\apsnodei$};
\node (a) at (26em,14.6em) {$\apsnodei$};
\node (aa) at (25.3em,14.2em) {};
\node (b) at (20em,14.75em) {$\apsnodei$};
\node[ttns] (c) at (23em,12.668em) {};
\draw (c) -- (ab);
\draw (c) -- (aa);
\draw (c) -- (b);
\node[pn] at (23em,7em) {$\Xi[]$};
\node[pn] at (20em,17.4em) {$\Gamma$};
\node[pn] at (26em,17.4em) {$\Delta$};
\node (d) at (23em,4em) {$D$};
\node at (15.5em,12.5em) {$\rightarrow$};
\node at (15.5em,13.5em) {$\mathcal{A}$};
\node (ab) at (8em,9.8em) {$C$};
\node (a) at (11em,14.6em) {$A\bdl C$};
\node (aa) at (10.3em,14.2em) {};
\node (b) at (5em,14.75em) {$A$};
\node[ttns] (c) at (8em,12.668em) {};
\draw (c) -- (ab);
\draw (c) -- (aa);
\draw (c) -- (b);
\node[pn] at (8em,7em) {$\Pi_{\Xi[]}$};
\node[pn] at (5em,17.6em) {$\Pi_{\Gamma}$};
\node[pn] at (11em,17.6em) {$\Pi_{\Delta}$};
\node (d) at (8em,4em) {$D$};
\end{tikzpicture}
\end{center}
Removing the tensor link produces three smaller trees, and the induction hypothesis therefore gives us three proofs, $\delta_1$ of $\Gamma\vdash A$, $\delta_2$ of $\Delta\vdash A\bdl C$, and $\delta_3$ of $\Xi[C]\vdash D$. We combine these three proofs as follows to produce a proof of $\Xi[\Gamma \bstr \Delta]\vdash D$ as required.
\[
\infer[\textit{Cut}]{\Xi[\Gamma\bstr\Delta]\vdash D}{
   \infer[\textit{Cut}]{\Gamma\bstr \Delta\vdash C\proofspace}{
       \infer*[\delta_2]{\Delta \vdash A\bdl C}{}
      & \infer[\bdl L]{\Gamma\bstr A\bdl C\vdash C}{
          \infer*[\delta_1]{\Gamma\vdash A\proofspace}{}
        & \infer[\textit{Ax}]{C\vdash C\proofspace}{}
      }
   }
 & \infer*[\delta_3]{\Xi[C]\vdash D}{}
}
\]
In the case for $[\lidc R]$, we have the following proof structure and abstract proof structure.
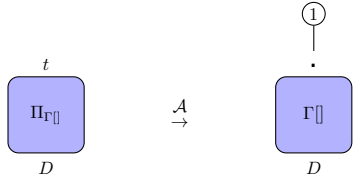
\begin{center}
\begin{tikzpicture}[scale=0.75]
\node (ab) at (23em,9.8em) {$\apsnodei$};
\node[tns] (c) at (23em,12.668em) {};
\node (cl) at (23em,12.668em) {$1$};
\draw (c) -- (ab);
\node[pn] at (23em,7em) {$\Gamma[]$};
\node (d) at (23em,4em) {$D$};
\node at (15.5em,6.5em) {$\rightarrow$};
\node at (15.5em,7.5em) {$\mathcal{A}$};
\node (ab) at (8em,9.8em) {$\lidc$};
\node[pn] at (8em,7em) {$\Pi_{\Gamma[]}$};
\node (d) at (8em,4em) {$D$};
\end{tikzpicture}
\end{center}
We remove the $[\lidc R]$ tensor link from the abstract proof structure and apply the induction hypothesis to obtain a proof $\delta$ of $\Gamma[\lidc]\vdash D$ (removing the $[\lidc R]$ link leaves a $t$ hypothesis). We can then produce a proof of $\Gamma[\lidstr]\vdash D$ as follows.
\[
\infer[\textit{Cut}]{\Gamma[\lidstr]\vdash D}{
 \infer[\lidc R]{\lidstr\vdash \lidc\proofspace}{}
 & \infer*[\delta]{\Gamma[\lidc]\vdash D}{}
}
\] 

This concludes the proof of the base case, where there sequence of conversions on the abstract proof structure is empty. We now look at the inductive case, where there are $l > 0$ conversions. We look at the last conversion in the sequence. 


The key case is when the last conversion is the contraction of a par link. We only look at the cases for $\lidc L$ and  $\bdr R$. The structural rules correspond rather directly to the structural conversion of the same name, and we only show the $\betaexp$ case.

\begin{description}

\item[$\lidc L$] When the last contraction is a $\lidc L$ contraction, we are in the following situation. 

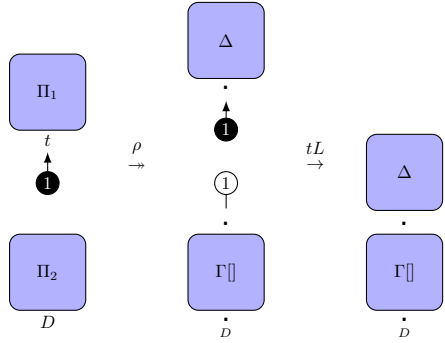
\begin{center}
\begin{tikzpicture}[scale=0.75]
\node[pn] at (10em,20.1em) {};
\node at (10em,20.1em) {$\Pi_1$};
\node[pn] at (10em,10em) {};
\node (p) at (10em,10em) {$\Pi_2$};	
\node (d) at (10em,7.2em) {$D$};
\node [par] (par) at (10em,15em) {};
\node at (10em,15em) {\textcolor{white}{$\lidstr$}};
\node (topt) at (10em,17.4em) {$\lidc$};
\path[>=latex,->]  (par) edge (topt);
\node[pn] at (20em,10em) {};
\node (a) at (20em,10em) {$\Gamma[]$};
\node (dg) at (20em,12.8em) {$\apsnodei$};	
\node [tns] (t) at (20em,15em) {};
\node at (20em,15em) {$\lidstr$};
\draw (t) -- (dg);
\node (d) at (20em,7.2em) {$\apsnode{}{D}$};
\node [par] (par) at (20em,18em) {};
\node at (20em,18em) {\textcolor{white}{$\lidstr$}};
\node (topt) at (20em,20.4em) {$\apsnodei$};
\path[>=latex,->]  (par) edge (topt);
\node[pn] at (20em,23em) {};
\node at (20em,23em) {$\Delta$};
\node at (15em,17em) {$\rho$};
\node at (15em,16em) {$\twoheadrightarrow$};
%
\node[pn] at (30em,10em) {};
\node (a) at (30em,10em) {$\Gamma[]$};
\node[pn] at (30em,15.6em) {};
\node at (30em,15.6em) {$\Delta$};
\node (dg) at (30em,12.8em) {$\apsnodei$};	
\node (d) at (30em,7.2em) {$\apsnode{}{D}$};
\node at (25em,16.0em) {$\rightarrow$};
\node at (25em,17.0em) {$\lidc L$};

\end{tikzpicture}	
\end{center}
The general idea of the proof is that the par link corresponding to the last contraction forms a `barrier' forcing all other conversions before it to be on one or the other side of it. This allows us to remove the par link and end up with two proof nets.

Suppose a conversion operates on both sides of the par link simultaneously, this means that, after the conversions in $\rho$,  $\Gamma[\lidstr]$ and $\Delta$ must be the same component (since they share at least a vertex), and the $\lidc L$ contraction connects two vertices which were already connected in this component. But then $\Gamma[\Delta]$ cannot be a tensor graph, since it contains a cycle. 



We can therefore remove the par link, and divide the conversions in $\rho$ into two subsequences, $\rho_1$ transforming the substructure $\Pi_1$ into $\Delta\vdash \lidc$ and $\rho_2$ transforming the substructure $\Pi_2$ into $\Gamma[\lidstr] \vdash D$ as follows.

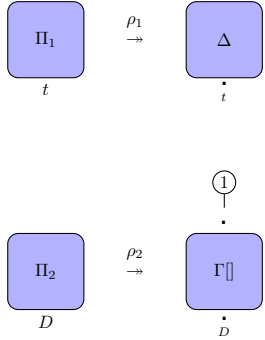
\begin{center}
\begin{tikzpicture}[scale=0.75]
\node[pn] at (10em,23em) {};
\node at (10em,23em) {$\Pi_1$};
\node[pn] at (10em,10em) {};
\node (p) at (10em,10em) {$\Pi_2$};	
\node (d) at (10em,7.2em) {$D$};
\node (topt) at (10em,20.2em) {$\lidc$};
\node[pn] at (20em,10em) {};
\node (a) at (20em,10em) {$\Gamma[]$};
\node (dg) at (20em,12.8em) {$\apsnodei$};	
\node [tns] (t) at (20em,15em) {};
\node at (20em,15em) {$\lidstr$};
\draw (t) -- (dg);
\node (d) at (20em,7.2em) {$\apsnode{}{D}$};
\node (topt) at (20em,20.4em) {$\apsnode{}{t}$};
\node[pn] at (20em,23em) {};
\node at (20em,23em) {$\Delta$};
\node at (15em,11em) {$\rho_2$};
\node at (15em,10em) {$\twoheadrightarrow$};
\node at (15em,24em) {$\rho_1$};
\node at (15em,23em) {$\twoheadrightarrow$};
\end{tikzpicture}	
\end{center}
These two substructures are therefore proof nets, and since the total number of conversions in the two sequences sums to $l-1$ (we started with $l$ conversions and removed the final  $\lidc L$ contraction), we can apply the induction hypothesis to obtain proofs $\delta_1$ of $\Delta\vdash \lidc$ and $\delta_2$ of $\Gamma[\lidstr]\vdash D$. We can combine these to produce the required proof of $\Gamma[\Delta]\vdash D$ as follows.
\[
\infer[\textit{Cut}]{\Gamma[\Delta]\vdash D}{
  \infer*[\delta_1]{\Delta\vdash \lidc}{}
 &  \infer[\lidc L]{\Gamma[\lidc]\vdash D}{
   \infer*[\delta_2]{\Gamma[\lidstr]\vdash D}{}
   }
  }
\]

\item[$\bdr R$] When the last contraction is a $\bdr R$ contraction, we are in the following situation.	
	
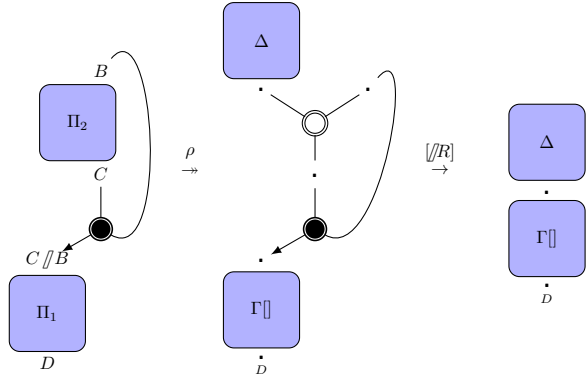
\begin{center}	
\begin{tikzpicture}[scale=0.75]
\node[pn] at (-10.3em,7.7em) {$\Pi_2$}; 
\node (ab) at (-9em,4.8em) {$C$};
\node (b) at (-9em,10.6em) {$B$};
\node (pa) at (-12em,0em) {$C\bdr B$};
\node[pn] at (-12em,-3.0em) {$\Pi_1$};
\node at (-12em,-5.8em) {$D$};
\node (pad) at (-12em,0em) {\phantom{M}};
\node[ppar] (pc) at (-9em,1.732em) {};
\draw (pc) -- (ab);
\path[>=latex,->]  (pc) edge (pad);
\draw (b) to [out=50,in=330] (pc);
\node (ab) at (3em,4.8em) {$\apsnodei$};
\node (a) at (0,9.6em) {$\apsnodei$};
\node (b) at (6em,9.6em) {$\apsnodei$};
\node[ttns] (c) at (3em,7.668em) {};
\draw (c) -- (ab);
\draw (c) -- (a);
\draw (c) -- (b);
\node[pn] at (0em,-2.8em) {$\Gamma[]$};
\node at (0em,-5.6em) {$\apsnode{}{D}$};
\node (pa) at (0,0) {$\apsnodei$};
\node (pad) at (0em,0em) {\emptynode};
\node[ppar] (pc) at (3em,1.732em) {};
\draw (pc) -- (ab);
\path[>=latex,->]  (pc) edge (pad);
\draw (b) to [out=50,in=330] (pc);
\node[pn] at (0.0em,12.3em) {$\Delta$}; 
\node[pn] at (16.0em,6.6em) {$\Delta$};
\node[pn] at (16em,1.2em) {$\Gamma[]$}; 
\node at (16em,3.8em) {$\apsnodei$}; 
\node at (16em,-1.6em) {$\apsnode{}{D}$};
\node (labl) at (-4em,6.0em) {$\rho$}; 
\node  at (-4em,5.0em) {$\twoheadrightarrow$}; 
\node (labl) at (10em,6.0em) {$[\bdr R]$}; 
\node  at (10em,5.0em) {$\rightarrow$}; 
%
\end{tikzpicture}
\end{center}

As in the previous case, we remove the par link and divide the conversions in $\rho$ depending on the `side' of the par link  where they applied. This gives the following two proof nets, each with a shorter conversion sequence (since the final contraction has been removed).

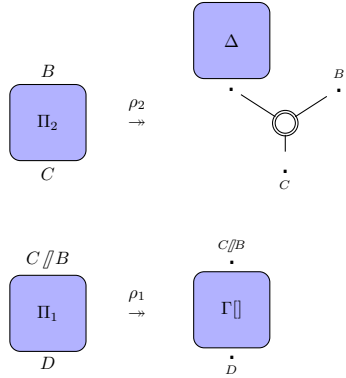
\begin{center}	
\begin{tikzpicture}[scale=0.75]
\node[pn] at (-10.3em,7.7em) {$\Pi_2$}; 
\node (ab) at (-10.3em,4.8em) {$C$};
\node (b) at (-10.3em,10.6em) {$B$};
\node (pa) at (-10.3em,0em) {$C\bdr B$};
\node[pn] at (-10.3em,-3.0em) {$\Pi_1$};
\node at (-10.3em,-5.8em) {$D$};
\node (ab) at (3em,4.8em) {$\apsnode{}{C}$};
\node (a) at (0,9.6em) {$\apsnodei$};
\node (b) at (6em,9.6em) {$\apsnode{B}{}$};
\node[ttns] (c) at (3em,7.668em) {};
\draw (c) -- (ab);
\draw (c) -- (a);
\draw (c) -- (b);
\node[pn] at (0em,-2.8em) {$\Gamma[]$};
\node at (0em,-5.6em) {$\apsnode{}{D}$};
\node (pa) at (0,0) {$\apsnode{C\bdr B}{}$};
\node[pn] at (0.0em,12.3em) {$\Delta$}; 
\node (labl) at (-5.3em,8.7em) {$\rho_2$}; 
\node  at (-5.3em,7.7em) {$\twoheadrightarrow$}; 
\node (labl) at (-5.3em,-2em) {$\rho_1$}; 
\node  at (-5.3em,-3em) {$\twoheadrightarrow$}; 
\end{tikzpicture}
\end{center}

Induction hypothesis gives us a proof $\delta_1$ of $\Delta\bstr B\vdash C$ and a proof $\delta_2$ of $\Gamma[C\bdr B]\vdash D$. We combine these into the required proof of $\Gamma[\Delta]\vdash D$ as follows.
\[
\infer[\textit{Cut}]{\Gamma[\Delta]\vdash D}{
  \infer[\bdr R]{\Delta\vdash C\bdr B}{\infer*[\delta_1]{\Delta\bstr B \vdash C}{}
  }
 & \infer*[\delta_2]{\Gamma[C\bdr B]\vdash D}{}
}
\]	
The other contractions for the binary connectives are treated similarly.
\item[$\betaexp$] If the last conversion is a $\betaexp$ conversion, we are schematically in the following case.

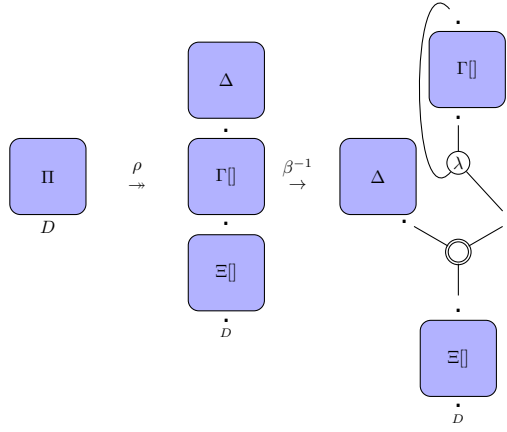
\begin{center}
\begin{tikzpicture}[scale=0.75]
\node[pn] at (-7em,-11em) {$\Pi$};
\node at (-7em,-13.8em) {$D$};
\node at (-2em,-11.5em) {$\twoheadrightarrow$};
\node at (-2em,-10.5em) {$\rho$};
\node[pn] at (3em,-5.6em) {$\Delta$}; 
\node[pn] at (3em,-11em) {$\Gamma[]$}; 
\node[pn] at (3em,-16.4em) {$\Xi[]$}; 
\node at (3em,-19.2em) {$\apsnode{}{D}$};
\node at (3em,-8.4em) {$\apsnodei$};
\node at (3em,-13.6em) {$\apsnodei$}; 
\node at (7em,-11.5em) {$\rightarrow$};
\node at (7em,-10.5em) {$\betaexp$};
%
\node[pn] at (11.5em,-11em) {$\Delta$};
\node (lt) at (13em,-13.5em) {$\apsnodei$}; 
\node[ttns] (tc) at (16em,-15.232em) {}; 
\node (rt) at (19em,-13.5em) {$\apsnodei$}; 
\node (top) at (16em,-18.3em) {$\centerdot$};
\node[pn] at (16em,-21.2em) {$\Xi[]$}; 
\node (d) at (16em,-24.0em) {$\apsnode{}{D}$};
\draw (lt) -- (tc);
\draw (rt) -- (tc);
\draw (tc) -- (top);
\node[tns] (lam) at (16em,-10.232em) {}; 
\node (lamlab) at (16em,-10.232em) {$\lambda$}; 
\node (bl) at (16em,-7.5em) {$\centerdot$}; 
\draw (bl) -- (lam);
\draw (lam) -- (rt);
\node[pn] at (16.5em,-5em) {$\Gamma[]$}; 
\node (bot) at (16em,-2.2em) {$\centerdot$}; 
\draw (bot) to [out=130,in=230] (lam); 
\end{tikzpicture}
\end{center}
Removing the final $\betaexp$ conversion produces a shorter conversion sequence and we can therefore apply the induction hypothesis to obtain a proof $\delta$ of $\Xi[\Gamma[\Delta]] \vdash D$. We can extend this proof as follows to produce the required proof.
\[
\infer[\betaexp]{\Xi[(\Delta \bstr \lambda x.\Gamma[x])]\vdash D}{\infer*[\delta]{\Xi[\Gamma[\Delta]] \vdash D}{}}
\]
The variables $x$ in the conclusion of the proof is chosen to not appear elsewhere in $\Gamma$ or $\Delta$. The other structural rules are similar.
\end{description}

\qed

\subsection{Extending \nllam{}}

One advantage of the proof net calculus presented in this section is that it is easy to adapt when extending the logic. For example, we can add an associative mode to \nllam{} with the corresponding structural rules. This would give this extended logic an easy treatment of phenomena like right-node raising which are a known problem for non-associative logics. 

In general, we can import the entire multimodal setup of \citep{mp} --- modes, unary connectives, structural rules --- into \nllam{} with the following proviso: all structural rules must be confluent with respect to the $\beta$ redex (see also Section~\ref{sec:confl}). Well-behavedness with respect to beta reduction is a standard restriction from the literature on adding rewrite rules or equations to the lambda calculus \citep[Chapter~4]{lambda}.

The simplest way to enforce this is to ensure a $\beta$ redex cannot be part of a critical pair with another structural rewrite, by prohibiting structural rules other than $\betaexp$ and $\betared$ (plus, eventually, an $\eta$ reduction as discussed in Section~\ref{sec:confl}) to modify `$\bstr$' and `$\lambda$' links.

\section{Decidability and Complexity}
\label{sec:deci}

For decidability, we essentially follow the argument of \citet{barker2019nllam}, with the minor modification that decidability for proof nets corresponds to forward chaining proof search, as opposed to the backward chaining proof search of the sequent calculus used by Barker. The advantage of a forward chaining proof search strategy is that we compute the antecedent structure as output, rather than requiring it as part of the input.

Looking at the rewrite rules for abstract proof structures, the only rules which increase the size of the structure are the right-to-left versions of the structural conversions on Table~\ref{tab:srlam}, the expanding conversions $\betaexp$, $\lidstr\lstr^{-1}$ and $\lstr\lidstr^{-1}$. 
This is a potential problem for decidability, since these rules can be used to add links to the abstract proof structure (at least in principle) without limit.
However, for each of these conversions, we can restrict their application in such a way that the size of the structure no longer increases. This is easiest for the $\lidstr$ rules. We can assume, without loss of generality, that we always remove the occurrences of $\lidstr$ as much as possible (given that the end-sequent is required to only contain the \nlambek{} structural connective `$\lstr$', with `$\lidstr$' its identity element, removal of $\lidstr$ is only impossible when the antecedent is itself identical to $\lidstr$). Now from inspection of the rewrite rules, the only rule which could require one of the $\lidstr^{-1}$ rules are the following contractions.
\begin{enumerate}
\item $\lidc L$ can require either the $\lidstr\lstr^{-1}$ or the $\lstr\lidstr^{-1}$ rewrite,
\item  $\ldl R$ can require the $\lstr\lidstr^{-1}$ rewrite,
\item  $\ldr R$ can require the  $\lidstr\lstr^{-1}$ rewrite.
\end{enumerate}
 

 \begin{figure}
\begin{center}
\begin{tikzpicture}[scale=0.75]
\node (la) at (1em,0.0em) {$\apsnode{h_2}{c}$};
\node (lb) at (-3em,0em) {$\apsnode{h_1}{}$}; 
\node (lbb) at (-3em,0em) {\emptynode};
\node [par] (pt) at (-3em,-2.4em) {};
\node at (-3em,-2.4em) {\textcolor{white}{$1$}};
\path[>=latex,->]  (pt) edge (lbb);
\node (lb) at (10em,0em) {$\apsnode{h_1}{}$}; 
\node (lbb) at (10em,0em) {\emptynode};
\node [par] (pt) at (10em,-2.4em) {};
\node at (10em,-2.4em) {\textcolor{white}{$1$}};
\path[>=latex,->]  (pt) edge (lbb);
\node (ab) at (16em,-2.0em) {$\apsnode{}{c}$};
\node (abb) at (16em,-2.0em) {\emptynode};
\node (a) at (13em,2.8em) {$\apsnodei$};
\node (b) at (19em,2.8em) {$\apsnode{h_2}{}$};
\node[tns] (c) at (16em,0.868em) {};
\draw (c) -- (abb);
\draw (c) -- (a);
\draw (c) -- (b);
\node[tns] (e) at (13em,5.2em) {};
\node at (13em,5.2em) {\lidstr};
\draw (e) -- (a);
\node at (23em,1.0em) {$\lidc L$};
\node at (23.0em,0.0em) {$\longrightarrow$}; 
\node (ab) at (31em,-2.0em) {$\apsnode{}{c}$};
\node (abb) at (31em,-2.0em) {\emptynode};
\node (a) at (28em,2.8em) {$\apsnode{h_1}{}$};
\node (b) at (34em,2.8em) {$\apsnode{h_2}{}$};
\node[tns] (c) at (31em,0.868em) {};
\draw (c) -- (abb);
\draw (c) -- (a);
\draw (c) -- (b);
\node at (5em,1.0em) {$\lidstr\lstr^{-1}$};
\node at (5.0em,0.0em) {$\longrightarrow$}; 
\end{tikzpicture}
\end{center}
\caption{Combination of the $\lidstr\lstr^{-1}$ and $\lidc L$ graph rewrites.}
\label{fig:telim}
\end{figure}

We can combine the rewrites of $\lidstr\lstr^{-1}$ and $\lidc L$ as shown in Figure~\ref{fig:telim}. Doing this rewrite in a single step, reduces the size of the structure (given that Definition~\ref{def:size} counts par links as 2 but tensor links as 1, the size of the structure is reduced by 1). The combination of  $\lstr\lidstr^{-1}$ with $\lidc L$ is left-right symmetric with Figure~\ref{fig:telim}.

 We can combine the rewrites of $\lstr\lidstr^{-1}$ and $\ldl R$ as shown in Figure~\ref{fig:idelim}. Adding a new rule transforming the left hand side of the figure directly to the right hand side ensures that the size of the structure decreases. A similar strategy can be used for the combination of the $\lidstr\lstr^{-1}$ rule with the $\ldr R$ rule (it is left-right symmetric with Figure~\ref{fig:idelim}). 

\begin{figure}
\begin{center}
\begin{tikzpicture}[scale=0.75]
\node (ab) at (23em,4.8em) {$\apsnodei$};
\node (a) at (26em,9.6em) {$\apsnodei$};
\node (b) at (20em,9.6em) {$\apsnodei$};
\node[tns] (c) at (23em,7.668em) {};
\draw (c) -- (ab);
\draw (c) -- (a);
\draw (c) -- (b);
\node (pa) at (26em,0) {$\apsnode{}{c}$};
\node[par] (pc) at (23em,1.732em) {};
\draw (pc) -- (ab);
\path[>=latex,->]  (pc) edge (pa);
\draw (b) to [out=130,in=210] (pc);
\node[tns] (tid) at (26em,12em) {};
\node at (26em,12em) {$\lidstr$};
\draw (tid) -- (a);
\node (ab) at (10em,4.8em) {$\apsnodei$};
\node (pa) at (13em,0em) {$\apsnode{}{c}$};
\node[par] (pc) at (10em,1.732em) {};
\draw (pc) -- (ab);
\path[>=latex,->]  (pc) edge (pa);
\draw (ab) to [out=130,in=210] (pc);
\node (pa) at (36em,0em) {$\apsnode{}{c}$}; 
\node (paa) at (36em,0em) {\emptynode};
\node[tns] (tid) at (36em,2.4em) {};
\node at (36em,2.4em) {$\lidstr$};
\draw (tid) -- (paa);
\node at (16em,2.0em) {$\longrightarrow$};
\node at (16em,3.0em) {$\lstr\lidstr^{-1}$};
\node at (30em,2.0em) {$\longrightarrow$};
\node at (30em,3.0em) {$\ldl$};
\end{tikzpicture}
\end{center}
\caption{Combination of the $\lidstr$ and $\ldl$ graph rewrites.}
\label{fig:idelim}
\end{figure}

For the $\betaexp$ case, we again use essentially the same argument as \citet{barker2019nllam} to show decidability. 
In a proof net and its reduction sequence, the only ways to introduce a $\bstr$ tensor link are the $\bdr L$, $\bdl L$ and $\bpr R$ links, and the $\betaexp$ structural rule. The only ways to remove a $\bstr$ tensor link are the $\bdr$, $\bdl$ and $\bpr$ contractions and the $\betared$ structural rule.

We can remove cases where a $\bstr$ link is introduced by a $\betaexp$ structural rule and removed by a $\betared$ one, since in that case both structural rules can be removed.

\begin{figure}
\begin{center}
\begin{tikzpicture}[scale=0.75]
\node (h1) at (0em,-8.5em) {$\apsnode{}{c_1}$};
\node at (0em,-14.5em) {$\apsnode{h_2}{}$}; 
\node (h2) at (0em,-14.7em) {};
\node at (7em,-11.0em) {$\betaexp$};
\node at (7.0em,-12.0em) {$\longrightarrow$}; 
\node at (23em,-11.0em) {$\bdl$};
\node at (23.0em,-12.0em) {$\longrightarrow$}; 
\node (lt) at (13em,-13.5em) {$\apsnodei$}; 
\node[ttns] (tc) at (16em,-15.232em) {}; 
\node (rt) at (19em,-12.5em) {$\apsnode{}{}$}; 
\node (top) at (16em,-18.3em) {$\apsnodei$}; 
\draw (lt) -- (tc);
\draw (rt) -- (tc);
\draw (tc) -- (top);
\node[tns] (lam) at (16em,-9.232em) {}; 
\node (lamlab) at (16em,-9.232em) {$\lambda$}; 
\node (bl) at (16em,-6.7em) {$\apsnode{h_2}{}$};
\node (bll) at (16em,-6.7em) {\phantom{M}}; 
\draw (bll) -- (lam);
\draw (lam) -- (rt);
\node (bot) at (16em,-1.5em) {$\apsnode{}{c_1}$};
\draw (bot) to [out=130,in=230] (lam); 
\node[ppar] (pc) at (0em,-16.732em) {}; 
\node (c) at (2.5em,-18.9em) {$\apsnode{}{c_2}$};
\node (cc) at (2.6em,-18.7em) {};
\draw (h1) to [out=130,in=230] (pc);
\draw (h2) -- (pc);
\path[>=latex,->] (pc) edge (cc);
\node (cr) at (18.5em,-23.0em) {$\apsnode{}{c_2}$};
\node (crb) at (18.5em,-23.0em) {};
\node[ppar] (pcb) at (16em,-21em) {};
\draw (pcb) -- (top);
\path[>=latex,->] (pcb) edge (crb);
\draw (lt) to [out=130,in=230] (pcb);
\node (h1) at (30em,-8.5em) {$\apsnode{}{c_1}$};
\node at (30em,-14.5em) {$\apsnode{h_2}{}$}; 
\node (h2) at (30em,-14.7em) {};
\node[tns] (pc) at (30em,-16.732em) {}; 
\node at (30em,-16.732em) {$\lambda$};
\node (c) at (32.5em,-18.9em) {$\apsnode{}{c_2}$};
\node (cc) at (32.6em,-18.7em) {};
\draw (h1) to [out=130,in=230] (pc);
\draw (h2) -- (pc);
\draw (pc) -- (cc);
\end{tikzpicture}
\end{center}
\caption{Combination of the $\betaexp$ rule and the $\bdl R$ contraction.}
\label{fig:combbetabdl}
\end{figure}

This leaves us with the cases where the $\bstr$ link is introduced by a $\betaexp$ but removed by a contraction. However, looking at the shape of the contractions, this can only be a $\bdl R$ contraction, because the $\betaexp$ rule introduces a $\bstr$ link connected by its rightmost premiss to a $\lambda$ link, leaving only the conclusion and leftmost premiss free for connection to a par link and thereby excluding the $\bdr R$ and $\bpr L$ contractions (which both require the rightmost premiss to be connected to their par link instead of to the $\lambda$ link; although the reducing structural conversions can shorten a path, a $\lambda$ link can only be removed together with its paired $\bstr$ link). 

Figure~\ref{fig:combbetabdl} shows what happens when the $\betared$ structural rule is combined with the $\bdl R$ contraction. When we combine the two contractions, the size of the structure no longer increases.

To conclude, we can provide a decidable calculus by removing the expanding rules (the right to left versions of the rules in Table~\ref{tab:srlam}, that is, the $\betaexp$, $\lidstr\lstr^{-1}$ and $\lstr\lidstr^{-1}$ rewrite) and replacing them by the rewrite rules of Table~\ref{tab:deriv}. Each of the derived rules reduces the number of par links and is therefore bounded in its number of applications.
The $\betaexp\bdl$ rewrite rule has the side condition that the there is a path from node $c_1$ to node $h$ which passes only through tensor links (this is the standard condition on the $\betaexp$ rule).

\begin{table}
\begin{center}
\begin{tikzpicture}[scale=0.75]
\node (phantom) at (7em,0em) {\emptynode};
\node (phantomb) at (27em,0em) {\emptynode};
\node (ab) at (10em,4.8em) {$\apsnodei$};
\node (pa) at (7em,0em) {$\apsnode{}{c}$};
\node[par] (pc) at (10em,1.732em) {};
\draw (pc) -- (ab);
\path[>=latex,->]  (pc) edge (pa);
\draw (ab) to [out=50,in=330] (pc);
\node (pa) at (24em,0) {$\apsnode{}{c}$}; 
\node[tns] (tid) at (24em,2.4em) {};
\node at (24em,2.4em) {$\lidstr$};
\draw (tid) -- (pa);
\node at (17em,2.0em) {$\longrightarrow$};
\node at (17em,3.0em) {$\lidstr\lstr^{-1}\ldr$};
\node at (40em,2.4em) {$\apsnode{h_1}{c}$};
\node [par] (par) at (36em,0em) {};
\node at (36em,0em) {\textcolor{white}{$\lidstr$}};
\node (topt) at (36em,2.4em) {$\apsnode{h_2}{}$};
\node (toptt) at (36em,2.4em) {\emptynode};
\path[>=latex,->]  (par) edge (toptt);
\node (bc) at (48em,0em) {\emptynode};
\node (lc) at (45em,4.8em) {\emptynode};
\node (rc) at (51em,4.8em) {\emptynode};
\node [tns] (tns) at (48em,2.868em) {};
\draw (tns) -- (bc);
\draw (tns) -- (lc);
\draw (tns) -- (rc);
\node (b) at (48em,0em) {$\apsnode{}{c}$};
\node (l) at (45em,4.8em) {$\apsnode{h_1}{}$};
\node (l) at (51em,4.8em) {$\apsnode{h_2}{}$};
\node at (43em,2.0em) {$\longrightarrow$};
\node at (43em,3.0em) {$\lidstr\lstr\lidc$};
\end{tikzpicture}

\bigskip

\begin{tikzpicture}[scale=0.75]
\node (phantom) at (7em,0em) {\emptynode};
\node (phantom) at (27em,0em) {\emptynode};
\node (ab) at (10em,4.8em) {$\apsnodei$};
\node (pa) at (13em,0em) {$\apsnode{}{c}$};
\node[par] (pc) at (10em,1.732em) {};
\draw (pc) -- (ab);
\path[>=latex,->]  (pc) edge (pa);
\draw (ab) to [out=130,in=210] (pc);
\node (pa) at (24em,0) {$\apsnode{}{c}$}; 
\node[tns] (tid) at (24em,2.4em) {};
\node at (24em,2.4em) {$\lidstr$};
\draw (tid) -- (pa);
\node at (17em,2.0em) {$\longrightarrow$};
\node at (17em,3.0em) {$\lstr\lidstr^{-1}\ldl$};
\node at (40em,2.4em) {$\apsnode{h_2}{c}$};
\node [par] (par) at (36em,0em) {};
\node at (36em,0em) {\textcolor{white}{$\lidstr$}};
\node (topt) at (36em,2.4em) {$\apsnode{h_1}{}$};
\node (toptt) at (36em,2.4em) {\emptynode};
\path[>=latex,->]  (par) edge (toptt);
\node (bc) at (48em,0em) {\emptynode};
\node (lc) at (45em,4.8em) {\emptynode};
\node (rc) at (51em,4.8em) {\emptynode};
\node [tns] (tns) at (48em,2.868em) {};
\draw (tns) -- (bc);
\draw (tns) -- (lc);
\draw (tns) -- (rc);
\node (b) at (48em,0em) {$\apsnode{}{c}$};
\node (l) at (45em,4.8em) {$\apsnode{h_1}{}$};
\node (l) at (51em,4.8em) {$\apsnode{h_2}{}$};
\node at (43em,2.0em) {$\longrightarrow$};
\node at (43em,3.0em) {$\lstr\lidstr\lidc$};
\end{tikzpicture}

\bigskip

\begin{tikzpicture}[scale=0.75]
\node (pa) at (12em,0) {$\apsnode{}{c_2}$};
\node[ppar] (pc) at (9em,1.732em) {}; 
\node (pb) at (6em,0.15em) {$\apsnode{}{c_1}$};
\node (pdd) at (9em,4.8em) {\emptynode};
\node (pd) at (9em,4.8em) {$\apsnode{h}{}$};
\draw (pc) -- (pb);
\draw (pc) -- (pdd);
\path[>=latex,->]  (pc) edge (pa);
\node (pa) at (26em,0) {$\apsnode{}{c_2}$};
\node[tns] (pc) at (23em,1.732em) {};
\node (pcl) at (23em,1.732em) {$\lambda$};
\node (pb) at (20em,0.15em) {$\apsnode{}{c_1}$};
\node (pd) at (23em,4.8em) {$\apsnode{h}{}$};
\node (pdd) at (23em,4.8em) {\emptynode};
\draw (pc) -- (pb);
\draw (pc) -- (pdd);
\draw (pc) -- (pa);
\node at (16em,2em) {$\longrightarrow$};
\node at (16em,3em) {$\betaexp\bdl$};
\end{tikzpicture}
\end{center}
\caption{Derived rules for \nllam.}
\label{tab:deriv}
\end{table}

Given that each of the rules reduces the size of the abstract proof structure (according to Definition~\ref{def:size}, replacing a par link by a tensor links amounts to a size reduction) showing decidability is easy. However, we can do a bit better than just showing decidability. We first define the size of words in the lexicon.

\begin{definition}\label{def:wordsize} The \emph{size} of of word $w$ from the lexicon is the sum of the sizes (according to Definition~\ref{def:size}) of all words assigned to it in the lexicon. The \emph{size} of a set of goal formulas $G$ is the sum of the sizes of all formulas $g\in G$
\end{definition}


\begin{lemma}\label{lem:dec} \nllam{} is decidable in non-deterministic linear space.
\end{lemma}

\paragraph{Proof} We only present a proof sketch, abstracting away from the actual coding of graphs as strings on the tape of a Turing maching. 

Given a sentence and an \nllam{} grammar, specified by a lexicon and a set of goal formulas, the input space is the sum of the sizes for all words, plus the sum of the sizes for all goal formulas. Non-deterministically do lexical lookup (erasing all but one of the possible formulas for each word), non-deterministically enumerate all proof structures, then non-deterministically enumerate the rewrite sequences (each rewrite reducing the size). Finally, check whether the result is a tree. 

We also note that the proof of Theorem~1 of \citet{barker2019nllam} already implicitly establishes a linear space bound on proof search.\qed

\subsection{Confluence}
\label{sec:confl}

While replacing the expanding structural rules by the derived rules of Table~\ref{tab:deriv} makes the rewrite system decidable, it is unfortunately not confluent. This is not a defect of the derived rules, but of \nllam{} itself: for example, there is a divergence between the $\betaexp$ and $\bdl R$ conversions, and between multiple $\lidc L$ redexes (one $\lidc L$ link can reduces with different 0-ary tensor links, and this is aggravated by the fact that the expanding rewrites for $\lidstr$ can introduce these tensor links anywhere in the structure). 

Confluence is not a standard property for proof nets in the style of \cite{mp}, at least not in the presence of structural rules. Generally, a proof structure represents a potential reading of a given sentence, and absence of confluence corresponds to the possibility that different structures and different word orders can correspond to the same meaning, something which the method of proof nets by graph rewriting is designed to accommodate. However, from the point of view of rewriting, confluence is extremely desirable and in this section we will look at some conditions on the rewrite rule which entail confluence for the rewriting component of \nllam{} proof nets.

\paragraph{The unit} The $\lidstr\lstr\lidc$ and $\lstr\lidstr\lidc$ rewrites of Table~\ref{tab:deriv} are not confluent: we can choose any vertex for the $h_1/c$ or $h_2/c$ node, with different vertices producing different structures not convertible to a common structure. A similar argument can be used when rewriting the same vertices using a $\lidstr\lstr\lidc$ and a $\lstr\lidstr\lidc$ rewrite, which produces structures with the left and right subtree swapped, and not reducible to a common structure (given a non-commutative logic).

 The $\lidc L$ link introduces quite a bit of proof combinatorics: given that $\lidc$ is the formula equivalent of the empty string, a $\lidc$ formula can in principle be inserted anywhere in the antecedent (that is, to the left or to the right of any other node). It seems that --- at least for the examples used in natural language grammars --- we can restrict the $\lidc L$ contraction to its eta expansion case, where it only combines with a $\lidc R$ link present in the proof structure (and not introduced by a structural rule in the abstract proof structure). This reduces a lot of the combinatorics and keeps the meaning assignment by means of Curry-Howard terms simple (see Appendix~\ref{sec:unitsem} for discussion about the term assignment for the unit: it is not clear that the term constructor for $\lidc E$ is ever useful outside of occurrences which can be directly reduced). When we require that a $\lidc L$ link must be matched with a $\lidc R$ link from the lexicon (that is, both links correspond to the logical connective $\lidc$; the $\lidc R$ link is not introduced by a structural rule), we can eliminate the  $\lidstr\lstr\lidc$ and $\lstr\lidstr\lidc$ conversions entirely.
 
 At the same time, the derived contractions $\lstr\lidstr^{-1}\ldl$ and $\lidstr\lstr^{-1}\ldr$ correspond to empty antecedent derivations for the Lambek calculus connectives. Although logically unproblematic, these are generally disallowed for linguistic reasons, as discussed at the end of Section~\ref{sec:seq} --- even though \citet[Section~16.6]{bs14cont} admit them. So  
a  more drastic solution would be to restrict the application of the structural rules for the identity element to only those which reduce the structure, by removing the $\lidstr\lstr^{-1}$ and $\lstr\lidstr^{-1}$ structural rewrites (and the corresponding structural rules from the sequent calculus). This would make the derived rules of Table~\ref{tab:deriv} mentioning $\lidstr$ superfluous and only keep the $\betaexp\bdl$ rewrite. 

Even though from an algebraic point of view it is preferable to have `$\lidstr$' function as a true identity element for `$\lstr$', it makes sense to apply the structural rules for the unit only in one direction, both from a term/graph rewrite perspective and from a linguistic perspective. In the term- and graph rewrite literature, it is fairly standard to give equivalences an orientation towards the simpler terms/graphs, and, linguistically,  orienting the equations for the identity element gives an easy way to solve the problems with empty antecedent derivations.

\paragraph{The $\betaexp\bdl$ rule} The $\betaexp\bdl$ rewrite rule is not confluent either. 
Figure~\ref{fig:confl} shows the simplest example of this critical pair.
It is a proof net of $A\bdl B\vdash A\bdl B$, and following the topmost path and applying the $\bdl R$ contraction produces a proof net, but applying the $\betaexp\bdl$ rewrite instead produces another structure to which no further rewrites apply (assuming only the derived rules and simplification structural rules apply).
 
\begin{figure}
\begin{center}
\begin{tikzpicture}[scale=0.75]
\node (ab) at (8em,4.8em) {$B$};
\node (a) at (11em,9.6em) {$A\bdl B$};
\node (aa) at (11em,9.6em) {\phantom{Mx}}; 
\node (b) at (5em,9.6em) {$A$};
\node[ttns] (c) at (8em,7.668em) {};
\draw (c) -- (ab);
\draw (c) -- (a);
\draw (c) -- (b);
\node (pa) at (11em,0) {$A\bdl B$};
\node (paa) at (11em,0em) {\phantom{Mx}};
\node[ppar] (pc) at (8em,1.732em) {};
\draw (pc) -- (ab);
\path[>=latex,->]  (pc) edge (paa);
\draw (b) to [out=130,in=210] (pc);
\node (ab) at (23em,4.8em) {$\apsnodei$};
\node (a) at (26em,9.6em) {$\apsnode{A\bdl B}{}$};
\node (aa) at (26em,9.6em) {\emptynode};
\node (b) at (20em,9.6em) {$\apsnodei$};
\node[ttns] (c) at (23em,7.668em) {};
\draw (c) -- (ab);
\draw (c) -- (aa);
\draw (c) -- (b);
\node (pa) at (26em,0) {$\apsnode{}{A\bdl B}$};
\node (paa) at (26em,0em) {\emptynode};
\node[ppar] (pc) at (23em,1.732em) {};
\draw (pc) -- (ab);
\path[>=latex,->]  (pc) edge (paa);
\draw (b) to [out=130,in=210] (pc);
\node (ab) at (38em,2.8em) {$\apsnodei$};
\node (a) at (41em,7.6em) {$\apsnode{A\bdl B}{}$};
\node (aa) at (41em,7.6em) {\emptynode};
\node (b) at (35em,7.6em) {$\apsnodei$};
\node[ttns] (c) at (38em,5.668em) {};
\draw (c) -- (ab);
\draw (c) -- (aa);
\draw (c) -- (b);
\node (pa) at (41em,-2em) {$\apsnode{}{A\bdl B}$};
\node (paa) at (41em,-2em) {\emptynode};
\node[tns] (pc) at (38em,-0.268em) {};
\node (pcl) at (38em,-0.268em) {$\lambda$};
\draw (pc) -- (ab);
\path[>=latex,->]  (pc) edge (paa);
\draw (b) to [out=130,in=210] (pc);
\node at (38em,13em) {$\apsnode{A\bdl B}{A\bdl B}$};
\node at (15.5em,5.8em) {$\mathcal{A}$};
\path [->] (14.5em,4.8em) edge (16.5em,4.8em);
\path [->] (30em,4.8em) edge (32em,3.8em);
\path [->] (30em,9.8em) edge (32em,10.8em);
\node at (31em,5.8em) {$\betaexp\bdl$};
\node at (31em,12em) {$\bdl R$};
\path [densely dotted,->] (38em,8.8em) edge (38em,10.5em);
\node at (37.2em,9.6em) {$\eta$};
\end{tikzpicture}
\end{center}
\caption{Failure of confluence: a conflict between the $\betaexp\bdl$ and the $\bdl R$ contraction.}
\label{fig:confl}
\end{figure}

From the perspective of the lambda calculus, we have produced an $\eta$ redex. One solution would therefore be to add the equivalent of the $\eta$ rule to \nllam{}, which corresponds to Equation~\ref{eq:eta} below.
\begin{align}\label{eq:eta}
\Gamma &\Leftrightarrow \lambda x. (x \bstr \Gamma)	
\end{align}

The two proof nets in Figure~\ref{fig:confl} correspond to the following two sequent proofs, and under the $\eta$ rule, these two proofs are equivalent.
\begin{align*}
\infer[\bdl R]{A\bdl B\vdash A\bdl B}{
   \infer[\bdl L]{A \bstr A\bdl B \vdash B}{
       \infer[\textit{Ax}]{A\vdash A}{}
      &\infer[\textit{Ax}]{B\vdash B}{}
   }
}
& \equiv_{\eta}
\infer[\bdl R]{\lambda x. x\bstr A\bdl B \vdash A\bdl B}{
   \infer[\betaexp]{A \bstr \lambda x. x\bstr A\bdl B}{
   \infer[\bdl L]{A \bstr A\bdl B \vdash B}{
       \infer[\textit{Ax}]{A\vdash A}{}
      &\infer[\textit{Ax}]{B\vdash B}{}
   }}
}	
\end{align*}

At the level of proof nets, adding the $\eta$ reduction rule means adding the conversion labeled $\eta$ in Figure~\ref{fig:confl}.

 
 
If adding an $\eta$ equivalence rule to \nllam{} is undesirable --- the logics of \citet{bs14cont} and \citet{barker2019nllam} do not have such a rule after all --- then we can instead apply a greedy conversion strategy, as we discuss in the next paragraph. 

\paragraph{A confluent fragment} We now present a confluent fragment of the proof net calculus for \nllam{}. It deviates from the calculus \nllam{} of \citet{barker2019nllam} only in requiring each $\lidc L$ to be matched with with a $\lidc R$ rule, which amounts to removing the $\lidstr\lstr\lidc$ and $\lstr\lidstr\lidc$ rewrites. Depending on whether or not we want to allow empty antecedent derivations, the $\lstr\lidstr^{-1}\ldl$ and $\lidstr\lstr^{-1}\ldr$ rewrites can be either present or absent; this does not affect confluence.

\begin{definition}\label{def:eal}
An \emph{extended axiom linking} for a proof	 structure is:
\begin{enumerate}
\item a 1-1 matching between atomic hypothesis formulas and atomic conclusion formulas,
\item\label{it:eal2} a 1-1 matching between $\lidc L$ and $\lidc R$ links.
\end{enumerate}
\end{definition}

In the absence of the $\lidstr\lstr\lidc$ and $\lstr\lidstr\lidc$ rewrites, we can treat the logical rules for $\lidc$ as a sort of axiom links (see Appendix~\ref{sec:unitatom} for discussion). The notion of extended axiom link makes this explicit.

\begin{lemma}\label{lem:confl} Given a proof structure $\Pi$, an extended axiom linking $e$, the contractions of Table~\ref{tab:contr}, the reducing structural rules of Table~\ref{tab:srlam} and the derived $\lstr\lidstr^{-1}\ldl$, $\lidstr\lstr^{-1}\ldr$, and  $\betaexp\bdl$ rules of Table~\ref{tab:deriv}, reductions of the corresponding abstract proof structure are confluent modulo eager reductions. 
\end{lemma}

\paragraph{Proof} We have essentially used brute force to remove all critical pairs. 
\begin{itemize}
\item the critical pairs between a $\lidc L$ link and two $\lidc R$ links has been removed by item~\ref{it:eal2} of the definition of extended axiom links (Definition~\ref{def:eal}),
\item the critical pairs between $\betaexp\bdl$ and $\bdl R$, between $\lstr\lidstr^{-1}$ and $\ldr R$, and between $\lidstr\lstr^{-1}$ and $\ldl R$ have been removed by the rule ordering, removing the potential deadlocks.
\end{itemize}

Only the conflict between $\betaexp\bdl$ and $\bdl R$ requires a bit of reflection, since the other critical pairs are confluent. Suppose we are in a situation where the $\bdl R$ contraction (and not the $\betaexp\bdl$ conversion) must apply at some point. Look at the path between the $\bdl R$ par link and the $\bstr$ tensor link with which it contracts. Given that the $\bdl R$ contraction is necessary by assumption, the only conversions which can apply between the two links are any of the logical contractions (reducing the size by 3) and any of the $\lstr\lidstr$, $\lidstr\lstr$, and $\beta$ conversions (reducing the size by 2). Since the $\betaexp\bdl$ reduces the size by only 1, greedy reduction will be guaranteed to produce the required $\bdl R$ redex and contract it before considering the $\betaexp\bdl$ converion. \qed

\subsection{Time complexity}

The decidability result of Lemma~\ref{lem:dec} gave only an upper bound for the space complexity of parsing and proof search in \nllam{}.
Given the confluence result of Lemma~\ref{lem:confl}, we can now give a time complexity as well.

\begin{lemma}\label{lem:nllamnp} Parsing for \nllam{}, when excluding $\lidstr\lstr\lidc$ and $\lstr\lidstr\lidc$ conversions, is in NP.
\end{lemma}

\paragraph{Proof} To show the problem is in NP, it suffices to show that we can verify whether a proof candidate is a proof in polynomial time. Given a sentence and a set of goal formulas, non-deterministically select one of the goal formulas and, for each of the words in the sentence, one of the formulas the lexicon assigns to it. Non-deterministically compute the extended axiom linking to the given proof structure to produce an abstract proof structure. Reduce the resulting abstract proof structure using the eager rewrite system. Each rewrite involves scanning the abstract proof structure for redexes, noting the size reduction of the conversion, then performing one of the reductions which maximally decreases the size; each rewrite can be performed in time linear to the size of the abstract proof structure. Since each rewrite reduces the size, only a number of conversions linear in the size of the abstract proof structure have to be performed. Finally, check whether the result is a Lambek tree with the correct yield. \qed

It may be possible to sharpen Lemma~\ref{lem:nllamnp} to remove the restriction on empty antecedent derivations, or otherwise to directly refine the decidability proof of \citet[Theorem~1]{barker2019nllam} (or the decidability proof of Lemma~\ref{lem:dec}) for this purpose. We will leave this for further research.

Given that \nllam{} uses a non-associative base, we cannot apply the NP-completeness result for the associative Lambek calculus \citep{pentus06np} to show NP completeness of \nllam{}. However, I conjecture that it is possible to use the \nllam{} mechanisms for scope and extraction to prove NP completeness for parsing.  

\section{\nllam{} and hybrid type-logical grammars}
\label{sec:htlg}
\newcommand{\himpl}{\multimap}

Hybrid type-logical grammars \citep{kl20tls} are a logic combining the Lambek calculus implications with the lexical lambda term assignments of lambda grammars \citep{oehrle}. At first sight, the similarities between hybrid type-logical grammars (HTLG) and \nllam{} seem superficial: both have the Lambek calculus slashes\footnote{The standard definition of \nllam{}, as presented here, is as a non-associative logic, whereas hybrid type-logical grammars are generally presented as an associative logic. However, we can add or remove associativity (at least for the Lambek connectives) in either system. In what follows, we will compare \nllam{} to non-associative hybrid type-logical grammars.}, both have a $\lambda$ operator to build structures, but the logical foundations appear rather different: HTLG uses lambda term assignments for its lexical entries, which \nllam{} doesn't, and \nllam{} uses a second mode which is a standard residuated connective, unlike the linear implication of HTLG.
However, there is a surprising amount of overlap between the links and conversions of \nllam{} in this paper and the links and conversions used for proof nets for HTLG \citep{msg19htlg}. 

For reasons of space, we will not give a complete introduction to proof nets for hybrid type-logical grammars in this section. Instead, we will introduce the logical calculus by emphasising the similarity with the proof net calculus for \nllam{}. The reader interested in the full details of proof nets for HTLG is invited to read the paper by \citet{msg19htlg}.


Hybrid type-logical grammars allow lexical entries to provide lexical lambda terms which specify how the strings should be formed. 

\newcommand{\str}{s}

If we return to our previous example ``John saw everyone'', a possible HTLG lexicon for this same sentence would be the following.

\medskip
\begin{tabular}{llll}
\emph{Word} & \emph{Syntactic type} & \emph{Prosodic type} & \emph{Prosodic term} \\[0.6ex]
John & $np$ & $\str$ & $\textit{John}^{\str}$ \\
saw & $np \himpl (np \himpl s)$ & $\str \rightarrow (\str\rightarrow\str)$	 & $\lambda y^{\str} \lambda x^{\str}.\  x+\textit{saw}+y$ \\
everyone & $(np\himpl s)\himpl s$ &  $(\str \rightarrow \str)\rightarrow\str$	 & $\lambda P. (P\, \textit{everyone})$ 
\end{tabular}
\medskip

``John'' is simply assigned the formula $np$ and the prosodic term $\textit{John}$, a string. The lexical entry for ``saw'' is a function from two noun phrases to a sentence at the syntactic level and a function from two strings to a string at the prosodic level\footnote{HTLG also allows us to assign ``saw'' the lexical entry $(np\ldl s)\ldr np$, which is provably equivalent to the given entry.}. The prosodic term uses concatenation `+' to concatenate the string $x$ (corresponding to the subject) to the string $\textit{saw}$ followed by the string $y$ (corresponding to the object noun phrase). 

The key entry is the  lexical entry for ``everyone''. It is assigned the linear logic formula $(np\himpl s)\himpl s$. Syntactically, is takes a sentence missing a noun phrase as an argument to produce a sentence. Prosodically it is a function of type $(\str \rightarrow \str)\rightarrow\str$, which we assign the term $\lambda P. (P\, \textit{everyone})$. This means it takes a function from strings to strings as an argument (in this particular case the string of a sentence missing a noun phrase string) and fills this position with the word $\textit{everyone}$.

\subsection{Translations of links and rewrites}

We begin by a simple illustration to make the similarities immediate.

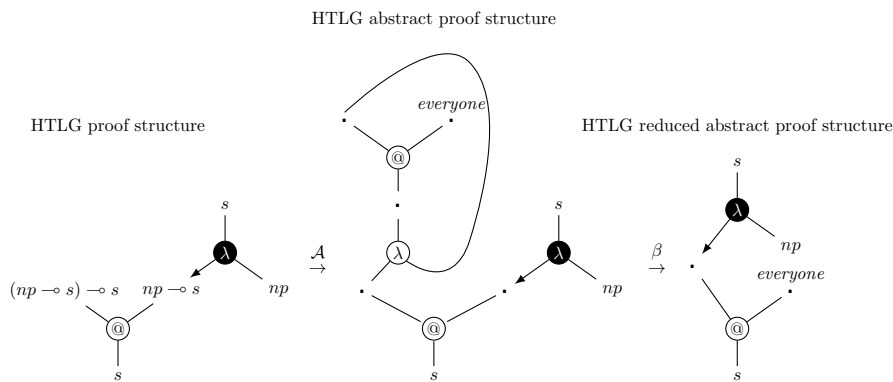
\begin{figure}
\begin{center}
\begin{tikzpicture}[scale=0.75]
\node at (3em,14em) {HTLG proof structure};
\node (spar) at (3em,0em) {$s$};
\node (a) at (6em,4.8em) {$np\himpl s$};
\node (b) at (0em,4.8em) {$(np\himpl s)\himpl s$};
\node[tns] (c) at (3em,2.668em) {}; 
\node (cl) at (3em,2.668em) {$@$}; 
\draw (c) -- (spar);
\draw (c) -- (a);
\draw (c) -- (b);
\node (d) at (12em,4.8em) {$np$};
\node [par] (pc) at (9em,6.932em) {};
\node (l) at  (9em,6.932em) {\textcolor{white}{$\lambda$}};
\node (e) at (9em,9.6em) {$s$};
\path[>=latex,->]  (pc) edge (a);
\draw (pc) -- (d);
\draw (pc) -- (e);
\end{tikzpicture}
\begin{tikzpicture}[scale=0.75]
\node at (2em,20em) {HTLG abstract proof structure};
\node (spar) at (2em,0em) {$s$};
\node (a) at (6em,4.8em) {$\apsnodei$};
\node (ev) at (-2em,4.8em) {$\apsnodei$};
\node[tns] (c) at (2em,2.668em) {}; 
\node (c) at (2em,2.668em) {$@$}; 
\draw (c) -- (spar);
\draw (c) -- (a);
\draw (c) -- (ev);
%
\node [par] (pc) at (9em,6.932em) {};
\node (l) at  (9em,6.932em) {\nodeindex{$\lambda$}};
\node (e) at (9em,9.6em) {$s$};
\node (f) at (12em,4.8em) {$np$}; 
\path[>=latex,->]  (pc) edge (a);
\draw (pc) -- (f);
\draw (pc) -- (e);
%
%
\node [tns] (pl) at (0em,6.932em) {};
\draw  (pl) -- (ev);
\node (l) at  (0em,6.932em) {$\lambda$};
\node (e) at (0em,9.6em) {$\apsnodei$}; 
\draw (pl) -- (e);
%
\node (a) at (3em,14.4em) {$\apsnodei$};
\node (b) at (-3em,14.4em) {$\apsnodei$};
\node (p) at (-3em,14.8em) {};
\node[tns] (c) at (0em,12.268em) {}; 
\node (cl) at (0em,12.268em) {$@$}; 
\draw (c) -- (e);
\draw (c) -- (a);
\draw (c) -- (b);
%
\node at (19em,14em) {HTLG reduced abstract proof structure};
\node (everyone) at (3em,15.1em) {\textit{everyone}};
\draw plot [smooth, tension=1] coordinates { (p) (4em,17.6em) (4em,7.6em) (pl.south east)};
\node (b) at (16.5em,6.2em) {$\apsnodei$};
\node [par] (pc) at (19em,9.332em) {};
\node (l) at  (19em,9.332em) {\nodeindex{$\lambda$}};
\node (e) at (19em,12.0em) {$s$};
\node (f) at (22em,7.2em) {$np$}; 
\path[>=latex,->]  (pc) edge (b);
\draw (pc) -- (f);
\draw (pc) -- (e);
\node (e) at (19em,0.0em) {$s$};
\node (a) at (22em,4.8em) {$\apsnodei$};
\node (ev) at (22em,5.6em) {\textit{everyone}};
\node[tns] (c) at (19em,2.668em) {}; 
\node (cl) at (19em,2.668em) {$@$}; 
\draw (c) -- (e);
\draw (c) -- (a);
\draw (c) -- (b);
\node (labl) at (14.5em,7.0em) {$\beta$}; 
\node  at (14.5em,6.0em) {$\rightarrow$}; 
\node (labl) at (-4.5em,7.0em) {$\mathcal{A}$}; 
\node  at (-4.5em,6.0em) {$\rightarrow$}; 
\end{tikzpicture}
\end{center}
\caption{Proof structure, abstract proof structure and $\beta$-reduced abstract proof structure for a quantifier in HTLG.}
\label{fig:qhtlg}
\end{figure}

Unfolding the HTLG lexical entry for \textit{everyone} with syntactic type $(np\himpl s)\himpl s$ and prosodic type $\lambda P. (P\, \textit{everyone})$ produces the proof structure shown on the left of Figure~\ref{fig:qhtlg}. As usual, tensor nodes are drawn with an open central circle whereas par nodes are drawn with a filled central circle; the mode of the the link is indicated by an index (for binary links the indices are `$+$' for Lambek calculus links, and `$@$' and `$\lambda$' for lambda grammar links, respectively corresponding to application and abstraction at the term level). 

\begin{figure}
\begin{center}
\begin{tikzpicture}[scale=0.75]
\node at (3em,14em) {$\nllam$ proof structure};
\node (spar) at (3em,0em) {$s$};
\node (a) at (6em,6em) {$np\bdl s$};
\node (b) at (0em,4.8em) {$s\bdr(np\bdl s)$};
\node[ttns] (c) at (3em,2.668em) {};
\draw (c) -- (spar);
\draw (c) -- (a);
\draw (c) -- (b);
\node (d) at (0em,7em) {$np$};
\node [ppar] (pc) at (3em,8.868em) {};
\node (e) at (3em,11.5em) {$s$};
\path[>=latex,->]  (pc) edge (a);
\draw (pc) -- (d);
\draw (pc) -- (e);
\node at (18em,14em) {$\nllam$ abstract proof structure};
\node (spar) at (18em,0em) {$s$};
\node (a) at (21em,6em) {$\apsnodei$};
\node (b) at (15em,4.8em) {\textit{everyone}};
\node[ttns] (c) at (18em,2.668em) {};
\draw (c) -- (spar);
\draw (c) -- (a);
\draw (c) -- (b);
\node (d) at (15em,7em) {$np$};
\node [ppar] (pc) at (18em,8.868em) {};
\node (e) at (18em,11.5em) {$s$};
\path[>=latex,->]  (pc) edge (a);
\draw (pc) -- (d);
\draw (pc) -- (e);
\node (labl) at (10.5em,7.0em) {$\mathcal{A}$}; 
\node  at (10.5em,6.0em) {$\rightarrow$}; 
\end{tikzpicture}
\end{center}
\caption{Proof structure and abstract proof structure for a quantifier in \nllam{}.}
\label{fig:qnllam}
\end{figure}
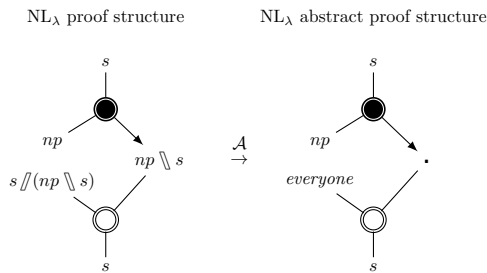

The corresponding abstract proof structure is shown in the middle of Figure~\ref{fig:qhtlg}. As usual, the $\mathcal{A}$ arrow denotes the conversion of a proof structure to an abstract proof structure. It removes formula information from internal nodes and replaces the lexical leaf $(np\himpl s)\himpl s$ by a graphical representation of the lambda term $\lambda P. (P\, \textit{everyone})$, where the binding of $P$ is indicated by the curved edge from the lambda tensor node to the left premiss of the $@$ tensor node --- just like the lambda abstractor link for \nllam{} proof nets indicates its bound variable, although the order of the conclusions differs between HTLG and \nllam{} proof nets. Using a partial evaluation strategy --- introduced in the context of Lambek calculus semantics by \citet{m99geo} and by \citet{sempn} --- we can then reduce the beta redex in this abstract proof structure to produce the structure on the right, as indicated by the $\beta$ arrow. 

For the quantifier in $\nllam$, which is assigned the formula $s\bdr(np\bdl s)$, we produce the proof structure shown on the left of Figure~\ref{fig:qnllam} and the abstract proof structure shown on the right of the figure. 

The two abstract proof structures are left-right symmetric. This is a consequence the different notational choices: the `functor' of an `application' is on the right for \citet{barker2019nllam} and the left for \citet{kl20tls}. This left-right symmetry applies only for the continuation connectives, the Lambek calculus connective of both calculi are identical.

The translation between the quantifier lexical entries is no accident. There is a simple equivalence between many of the links and rewrites in both logics.
Table~\ref{tab:transl} shows the translations between HTLG and \nllam{}. Elements of one logic which have no direct translation in the other are marked as `???' in the logic where this element is missing.

\begin{table}
\begin{tabular}{lcl}
HTLG && $\nllam$ \\ \hline
$+$ link & $\leftrightarrow$ &$\lstr$ link \\
$@$ with premisses $p_1-p_2$ & $\leftrightarrow$&
$\bstr$ with premisses $p_2-p_1$ \\
$\lambda$ tensor (lexicon) && ??? \\
$\lambda$ par with conclusions $c_1-c_2$ & $\leftrightarrow$ &
$\bdl$ par with conclusions $c_2-c_1$ \\
??? && $\lidc$, $\bdr$, $\bpr$ par links \\ \hline
contractions for $\ldr$, $\ldl$ &$\leftrightarrow$&contractions for $\ldr$, $\ldl$\\
??? && contraction for $\lpr$ \\
$\lambda$ par rewrite &$\leftrightarrow$& $\betaexp\bdl$ rewrite \\
$\beta$ rewrite &$\leftrightarrow$& $\beta$ rewrite \\
$\eta$ rewrite &$\leftrightarrow$& contraction for $\bdl$ \\
??? &&  contractions for $\lidc$, $\bdr$, $\bpr$ \\
\end{tabular}
\caption{Translations between HTLG and \nllam{}.}
\label{tab:transl}
\end{table}

The $\lambda$ tensor link (used to represent the prosodic lambda terms from the lexicon, as shown, for example, in the abstract proof structure in the middle of Figure~\ref{fig:qhtlg}) does not have a direct translation into $\nllam$ formulas: in $\nllam$, the $\lambda$ tensor link appears only in antecedent terms (however, it appears that we can emulate most and maybe even all of this functionality judiciously using atomic formulas and the $\bdl R$ rule to simulate extraction; see Section~\ref{sec:dutch} for examples and discussion). Inversely, the par links for $\bdr$ and $\bpr$ in $\nllam$ (and their logical contractions) have no direct translation into HTLG.

However, there are many similarities. Besides the shared links and contractions for the Lambek calculus implications, the key element of similarity is the presence of the $\beta$ rewrite in both logics and the fact that the $\lambda$ par rewrite in HTLG is equivalent to the $\betaexp\bdl$ rewrite of Table~\ref{tab:deriv}, with identical constraints on the rule application. 

\begin{lemma}\label{lem:nllamhtlg} When \nllam{} and (non-associative) HTLG lexical entries reduce to isomorphic (with respect to left-right symmetry of the continuation/linear links) abstract proof structures and require only the $\betared$, $\betaexp\bdl$/$\multimap I$, $\lidstr\lstr$, $\lstr\lidstr$, $\bdl R$/$\eta$, $\ldl R$, $\ldl R$ rewrites, then these lexical entries are logically equivalent.

The same equivalence holds between \nllam{} with added associativity (\llam{}) and standard, associative HTLG, when adding the associativity rewrites to both proof net calculi. 
\end{lemma}

\paragraph{Proof} Given that we produce isomorphic structures by assumption and that all the rewrite rules which can apply are equivalent as well, this is trivial.
\qed


As a consequence of Lemma~\ref{lem:nllamhtlg}, 
 many of the signature linguistic analyses proposed in the respective formalisms can be translated between the formalisms with ease. For each of these cases, we not only have equivalence at the level of the abstract proof structures, but also at level of the graph rewrites which apply to them. So in spite of the difference in logical foundations, the analyses proposed for these two logics converge in many interesting cases.

\subsection{Gapping} 

Gapping is a phenomenon which has received a lot of attention in the categorial grammar literature \citep{cgellipsis,mvf11displacement,kl12gap}. The basic idea is that sentences like the following can be analysed as a type of coordination.

\ex. John studies logic, and Charles phonetics.	

In the sentence above, the intended meaning is equivalent to the meaning of ``John studies logic and Charles \emph{studies} phonetics'', with the word ``studies'' missing syntactically from the second conjunct.

\begin{figure}
\vspace{-10\baselineskip}
\begin{center}
\begin{tikzpicture}[scale=0.75]
\path (34em,26.8em) coordinate (sametop);
\path (sametop) ++(-2.4em,0em) coordinate (pcoord);
\path (sametop) ++(2.4em,0em) coordinate (vcoord);
\path (sametop) ++(0em,-4.8em) coordinate (samebot);
\path (samebot) ++(0em, +2.668em) coordinate (sameat);
\node (p) at (pcoord) {$\apsnodei$};
\node (v) at (vcoord) {$\apsnodei$};
\node [tns] (tat) at (sameat) {};
\node at (sameat) {$@$};
\node (bot) at (samebot) {$\apsnodei$};
\draw (p) -- (tat);
\draw (bot) -- (tat);
\draw (v) -- (tat);
\path (samebot) ++(3.6em,0em) coordinate (xattop);
\path (xattop) ++(3.6em,0em) coordinate (xatr);
\path (xattop) ++(0em,-4.8em) coordinate (xatbot);
\path (xatbot) ++(0em,+2.668em) coordinate (xat);
\node [tns] (tx) at (xat) {};
\node at (xat) {$+$}; 
\node (xr) at (xatr) {$\apsnodei$};
\node (xb) at (xatbot) {$\apsnodei$};
\draw (bot) -- (tx);
\draw (tx) -- (xr);
\draw (tx) -- (xb);
\path (xatr) coordinate (andbot);
\path (andbot) ++(0em,2.668em) coordinate (andc);
\path (andbot) ++(0em,4.8em) coordinate (andtop);
\path (andtop) ++(-2.4em,0em) coordinate (andl);
\path (andtop) ++(2.4em,0em) coordinate (andr);
\node [tns] (andplus) at (andc) {};
\node at (andc) {$+$};
\node (and) at (andl) {\textit{and}};
\node (ar) at (andr) {$\apsnodei$};
\draw (xr) -- (andplus);
\draw (andplus) -- (and);
\draw (andplus) -- (ar);
\path (andr) coordinate (qebot);
\path (qebot) ++(0em,2.668em) coordinate (qec);
\path (qebot) ++(0em,4.8em) coordinate (qetop);
\path (qetop) ++(-2.4em,0em) coordinate (qel);
\path (qetop) ++(2.4em,0em) coordinate (qer);
\node [tns] (qeat) at (qec) {};
\node at (qec) {$@$};
\node (q) at (qel) {$\apsnodei$};
\node (eps) at (qer) {$\apsnodei$};
\draw (ar) -- (qeat);
\draw (qeat) -- (q);
\draw (qeat) -- (eps);
\path (qer) ++(0em,2.0em) coordinate (epsnp);
\node [tns] (epsc) at (epsnp) {};
\node at (epsnp) {$\epsilon$};
\draw (eps) -- (epsc);
\path (xatbot) coordinate (lamxtop);
\path (lamxtop) ++(0em,-2.668em) coordinate (lamx);
\path (lamxtop) ++(0em,-4.8em) coordinate (lamxbot);
\path (lamxbot) ++(-2.4em,0em) coordinate (lamxl); 
\node [tns] (lamc) at (lamx) {};
\node at (lamx) {$\lambda$};
\node (ll) at (lamxl) {$\apsnodei$};
\draw (xb) -- (lamc);
\draw (lamc) -- (ll);
\draw (lamc)..controls(55em,2em) and(55em,56em)..(v);
\path (lamxl) coordinate (patbot);
\path (patbot) ++(0em,-2.668em) coordinate (lamp);
\path (patbot) ++(0em,-4.8em) coordinate (lampbot);
\path (lampbot) ++(-2.4em,0em) coordinate (lampl);
\node [tns] (lamc) at (lamp) {};
\node at (lamp) {$\lambda$};
\node (llp) at (lampl) {$\apsnodei$};
\draw (llp) -- (lamc);
\draw (lamc) -- (ll);
\draw (lamc)..controls(58em,2em) and(58em,58em)..(p);
\path (lampl) coordinate (patbot);
\path (patbot) ++(0em,-2.668em) coordinate (lamp);
\path (patbot) ++(0em,-4.8em) coordinate (lampbot);
\path (lampbot) ++(-2.4em,0em) coordinate (lampl);
\node [tns] (lamc) at (lamp) {};
\node at (lamp) {$\lambda$};
\node (llpb) at (lampl) {$\apsnodei$};
\draw (llpb) -- (lamc);
\draw (llp) -- (lamc);
\draw (lamc)..controls(52em,-2em) and(52em,48em)..(q);
\path (lampbot) ++(4.8em,0em) coordinate (cattop);
\path (cattop) ++(4.8em,0em) coordinate (catr);
\path (cattop) ++(0em,-4.8em) coordinate (catbot);
\path (catbot) ++(0em,+2.668em) coordinate (cat);
\node [tns] (tx) at (cat) {};
\node at (cat) {$@$}; 
\node (cr) at (catr) {$\apsnodei$};
\node (cb) at (catbot) {$\apsnodei$};
\draw (llpb) -- (tx);
\draw (tx) -- (cr);
\draw (tx) -- (cb);
\path (catbot) ++(4.8em,0em) coordinate (bmid);
\path (bmid) ++(0em,-4.8em) coordinate (bot);
\path (bmid) ++(4.8em,0em) coordinate (br);
\path (bot) ++(0em,+2.668em) coordinate (bc);
\node [tns] (tb) at (bc) {};
\node at (bc) {$@$};
\node (nph) at (br) {$\apsnodei$};
\node (sc) at (bot) {$\apsnodei$}; 
\draw (sc) -- (tb);
\draw (nph) -- (tb);
\draw (tb) -- (cb);
\path (sc) ++(3.6em,0em) coordinate (bmid);
\path (bmid) ++(0em,-4.8em) coordinate (bot);
\path (bmid) ++(3.6em,0em) coordinate (brr);
\path (bot) ++(0em,+2.668em) coordinate (bc);
\node [tns] (tb) at (bc) {};
\node at (bc) {$@$};
\node (tvh) at (brr) {$tv$};
\node (ssc) at (bot) {$s$};
\draw (ssc) -- (tb);
\draw (tvh) -- (tb);
\draw (sc) -- (tb);
\path (catr) coordinate (bparl);
\path (bparl) ++(2.4em,0em) coordinate (bparm);
\path (bparm) ++(2.4em,0em) coordinate (bparr);
\path (bparm) ++(0em,4.8em) coordinate (bpart);
\path (bpart) ++(0em,-2.668em) coordinate (bparc);
\node [par] (bp) at (bparc) {};
\node at (bparc) {\textcolor{white}{\textbf{$\lambda$}}};
\node (nn) at (bparr) {$tv$};
\node (cpr) at (bpart) {$s$};
\draw (bp) -- (nn);
\draw (bp) -- (cpr);
\path[>=latex,->]  (bp) edge (cr);
\path (br) coordinate (parl);
\path (parl) ++(2.4em,0em) coordinate (parm);
\path (parm) ++(2.4em,0em) coordinate (parr);
\path (parm) ++(0em,4.8em) coordinate (part);
\path (part) ++(0em,-2.668em) coordinate (parc);
\node [par] (bp) at (parc) {};
\node at (parc) {\textcolor{white}{\textbf{$\lambda$}}};
\node (npc) at (parr) {$tv$};
\node (sh) at (part) {$s$};
\draw (bp) -- (npc);
\draw (bp) -- (sh);
\path[>=latex,->]  (bp) edge (nph);
\end{tikzpicture}
\kern-2em
\begin{tikzpicture}[scale=0.75]
\path (34em,26.8em) coordinate (sametop);
\path (sametop) ++(-2.4em,1em) coordinate (pcoord);
\path (sametop) ++(2.4em,0em) coordinate (vcoord);
\path (sametop) ++(0em,-4.8em) coordinate (samebot);
\path (samebot) ++(0em, +2.668em) coordinate (sameat);
\node (p) at (pcoord) {$\apsnodei$};
\node (v) at (vcoord) {$tv$};
\node [tns] (tat) at (sameat) {};
\node at (sameat) {$@$};
\node (bot) at (samebot) {$\apsnodei$};
\draw (p) -- (tat);
\draw (bot) -- (tat);
\draw (v) -- (tat);
\path (samebot) ++(3.6em,0em) coordinate (xattop);
\path (xattop) ++(3.6em,0em) coordinate (xatr);
\path (xattop) ++(0em,-4.8em) coordinate (xatbot);
\path (xatbot) ++(0em,+2.668em) coordinate (xat);
\node [tns] (tx) at (xat) {};
\node at (xat) {$+$}; 
\node (xr) at (xatr) {$\apsnodei$};
\node (xb) at (xatbot) {$\apsnodei$};
\draw (bot) -- (tx);
\draw (tx) -- (xr);
\draw (tx) -- (xb);
\path (xatr) coordinate (andbot);
\path (andbot) ++(0em,2.668em) coordinate (andc);
\path (andbot) ++(0em,4.8em) coordinate (andtop);
\path (andtop) ++(-2.4em,0em) coordinate (andl);
\path (andtop) ++(2.4em,0em) coordinate (andr);
\node [tns] (andplus) at (andc) {};
\node at (andc) {$+$};
\node (and) at (andl) {\textit{and}};
\node (ar) at (andr) {$\apsnodei$};
\draw (xr) -- (andplus);
\draw (andplus) -- (and);
\draw (andplus) -- (ar);
\path (andr) coordinate (qebot);
\path (qebot) ++(0em,2.668em) coordinate (qec);
\path (qebot) ++(0em,4.8em) coordinate (qetop);
\path (qetop) ++(-2.4em,1.5em) coordinate (qel);
\path (qetop) ++(2.4em,0em) coordinate (qer);
\node [tns] (qeat) at (qec) {};
\node at (qec) {$@$};
\node (q) at (qel) {$\apsnodei$};
\node (eps) at (qer) {$\apsnodei$};
\draw (ar) -- (qeat);
\draw (qeat) -- (q);
\draw (qeat) -- (eps);
\path (qer) ++(0em,2.0em) coordinate (epsnp);
\node [tns] (epsc) at (epsnp) {};
\node at (epsnp) {$\epsilon$};
\draw (eps) -- (epsc);
\path (pcoord) coordinate (bparl);
\path (bparl) ++(2.4em,1em) coordinate (bparm);
\path (bparm) ++(2.4em,0em) coordinate (bparr);
\path (bparm) ++(0em,4.8em) coordinate (bpart);
\path (bpart) ++(0em,-2.668em) coordinate (bparc);
\node [par] (bp) at (bparc) {};
\node at (bparc) {\textcolor{white}{\textbf{$\lambda$}}};
\node (nn) at (bparr) {$tv$};
\node (cpr) at (bpart) {$s$};
\draw (bp) -- (nn);
\draw (bp) -- (cpr);
\path[>=latex,->]  (bp) edge (p);
\path (qel) coordinate (parl);
\path (parl) ++(2.4em,3em) coordinate (parm);
\path (parm) ++(2.4em,0em) coordinate (parr);
\path (parm) ++(0em,4.8em) coordinate (part);
\path (part) ++(0em,-2.668em) coordinate (parc);
\node [par] (bp) at (parc) {};
\node at (parc) {\textcolor{white}{\textbf{$\lambda$}}};
\node (npc) at (parr) {$tv$};
\node (sh) at (part) {$s$};
\draw (bp) -- (npc);
\draw (bp) -- (sh);
\path[>=latex,->]  (bp) edge (q);
\path (pcoord) ++(-3em,0em) coordinate (arrowc);
\path (arrowc) ++(0em,1em) coordinate (alc);
\node at (arrowc) {$\rightarrow$};
\node at (alc) {$\beta \times 3$};
\end{tikzpicture}
\end{center}
\vspace{-\baselineskip}
\caption{Abstract proof structure of the lexical entry for gapping from \citet{kl20tls}, before and after $\beta$ reductions.}
\label{fig:gap} 
\end{figure}
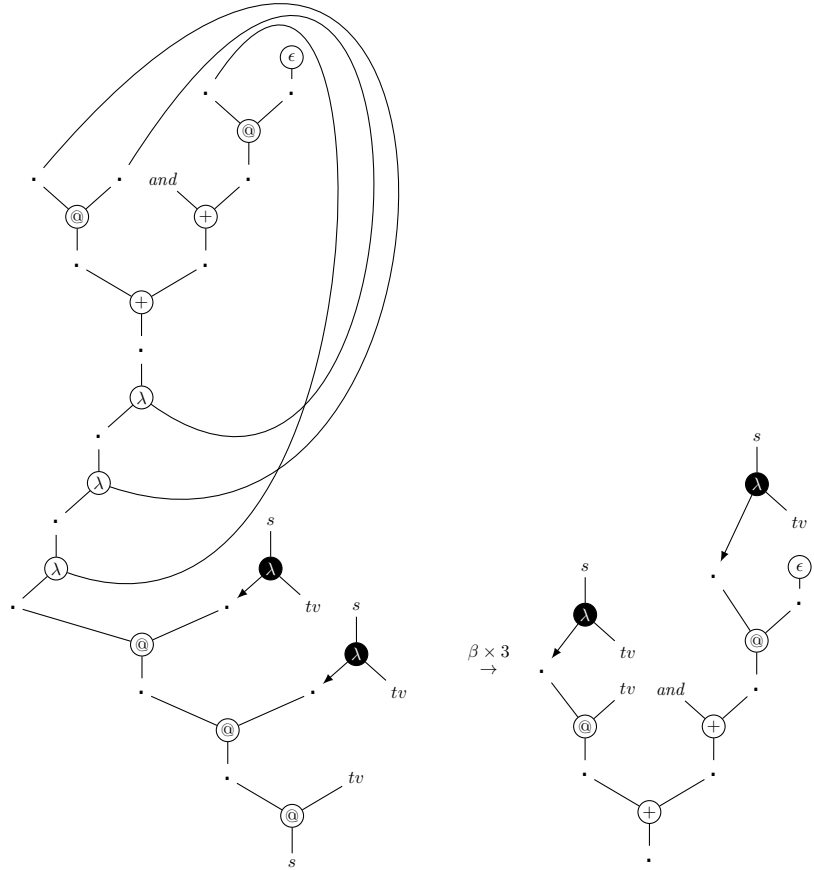

\citet{kl12gap,kl20tls} propose the formula $(tv \himpl s) \himpl (tv \himpl s) \himpl tv \himpl s$  with prosodic term $\lambda Q. \lambda P. \lambda v. (P\, v)+\textit{and}+(Q\, \epsilon)$ for the coordinator ``and'' in gapping constructions (where $tv$ is short for $(np\backslash s)/np$). The idea of this lexical entry is that it selects two sentences, each missing a transitive verb $tv$, then selects a transitive verb and inserts it in the leftmost sentence, whereas in the rightmost sentence the missing transitive verb is assigned the empty string at the term level. The advantage of such an analysis is that it is now easy to get the desired semantics of the sentence.

Looking at this lexical entry in terms of proof nets, 
the abstract proof structure corresponding to this formula and its assigned term is shown in Figure~\ref{fig:gap} (the occurrences of $tv$ have not been unfolded). The three $\lambda$ links correspond to the abstractions over $Q$ (corresponding to the rightmost sentence missing a transitive verb), $P$ (corresponding to the leftmost one) and $v$ (corresponding to the transitive verb). This is again just a graphical way to represent the lambda term assigned to the lexical entry. The unfolding of the lexical formula and its prosodic term has again produced an abstract proof structure which can be further reduced by beta reduction.
After performing the three beta reductions, we obtain the abstract proof structure shown below on the right.

We can obtain the $\nllam${} formula corresponding to this abstract proof structure by first mirroring the premisses of the $@$ links and the conclusions of the $\lambda$ par links, then taking as the main formula of each link the vertex closest to the lexical leaf ``and''. This entails that the two $@$ links become continuation product formulas in $\nllam$. 
Completing the computations produces the formula $((tv\bpr (tv\bdl s)) \ldl s)/(t\bpr (tv\bdl s))$. 
Although it is rather similar to the analysis of \citet[Section~3.4]{morrill}, this formula doesn't look like a typical coordination formula. Compared to the HTLG formula and the formula of \citet[Section~3.2.6]{mvf11displacement}, it uses a form of de-Currying on the last two arguments ($(tv \bdl s)$ and $tv$)  (although this is not a \emph{derivable} form of de-Currying since it mixes the Lambek and continuation modes).

\subsection{Parasitic scope: ``same'' and ``different''}
\label{sec:parasitic}

Words like ``same'' and ``different'' allow what \citet{bs14cont} call `pa\-ra\-sitic scope'. 
Take the following sentences, for example.

\ex.\label{ex:same1} Everyone read the same book.

\ex.\label{ex:diff} Everyone read different books.

\ex.\label{ex:noone} No one read the same book.

\ex.\label{ex:waiter} The same waiter served everyone.	

The reading of Sentence~\ref{ex:same1} is that everyone read some books, and that there is one specific book read by everyone. Sentence~\ref{ex:diff}, on the other hand, has the meaning that everyone read some set of books, but that these sets of books are all disjoint. Sentences~\ref{ex:diff} and~\ref{ex:noone} have essentially the same meaning. Sentence~\ref{ex:waiter} shows that the same type on phenomenon is possible with ``same'' occurring in the subject and when ``everyone'' is the object.

\citet{bs14cont} propose the formula $(np \bdl s) \bdr ((n\ldl n)\bdl (np \bdl s))$ for the word ``same'' to get the required semantic readings. It is an adjective $n\ldl n$ lifted with respect to the formula $np \bdl s$ using the continuation mode. This allows it to function locally as an adjective, while taking scope over the same $np \bdl s$ formula as selected by a quantifier. A fully worked out example with ``same'' can be found in Appendix~\ref{sec:example}.

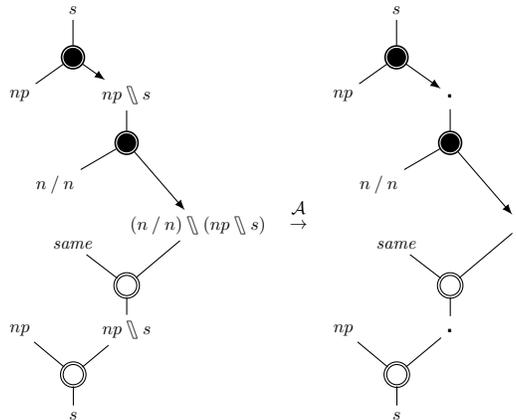
\begin{figure}
\begin{center}
\begin{tikzpicture}[scale=0.75]
\path (34em,26.8em) coordinate (sametop);
\path (sametop) ++(0em,-2.668em) coordinate (sparb);
\path (sametop) ++(-3em,-4.8em) coordinate (parbl);
\path (sametop) ++(3em,-4.8em) coordinate (parbr);
\node (sbot) at (sametop) {$s$};
\node [ppar] (parb) at (sparb) {};
\draw (sbot) -- (parb);
\node (npa) at (parbl) {$np$};
\node (vpb) at (parbr) {$np\bdl s$};
\draw (parb) -- (npa);
\path[>=latex,->]  (parb) edge (vpb);
\path (parbr) ++(0em,0em) coordinate (parcb);
\path (parcb) ++(0em,-2.668em) coordinate (parcc);
\path (parcb) ++(-4em,-5em) coordinate (parcl);
\path (parcb) ++(4em,-7.3em) coordinate (parcr);
\node (nn) at (parcl) {$n\ldr n$};
\node (nnnps) at (parcr) {$(n\ldr n)\bdl(np\bdl s)$};
\node [ppar] (prc) at (parcc) {};
\draw (vpb) -- (prc);
\draw (prc) -- (nn); 
\path[>=latex,->]  (prc) edge (nnnps);
\path (nn) ++(0em,0em) coordinate (tnnl);
\path (tnnl) ++(4em,0em) coordinate (tnnr);
\path (tnnl) ++(2em,-2.3em) coordinate (tnnc);
\path (tnnl) ++(2em,-4.8em) coordinate (tnnb);
\path (nnnps) ++(0em,0em) coordinate (nnnpsa);
\path (parcc) ++(0em,-8em) coordinate (sametp);
\path (sametp) ++(0em,-2.5em) coordinate (samebp);
\path (sametp) ++(-3em,+2.3em) coordinate (samep);
\node (same) at (samep) {\textit{same}};
\node [ttns] (st) at (sametp) {};
\node (vpr) at (samebp) {$np\bdl s$};  
\draw (st) -- (same);
\draw (st) -- (vpr);
\draw (st) -- (nnnps);
\path (samebp) ++(0em,0em) coordinate (vprp);
\path (vprp) ++(-6em,0em) coordinate (vplp);
\path (vprp) ++(-3em,-2.5em) coordinate (vpcp);
\path (vpcp) ++(0em,-2.3em) coordinate (vpbp);
\node (subj) at (vplp) {$np$};
\node [ttns] (tc) at (vpcp) {};
\node (bot) at (vpbp) {$s$};
\draw (tc) -- (bot);
\draw (tc) -- (subj);
\draw (tc) -- (vpr);
\end{tikzpicture}
\begin{tikzpicture}[scale=0.75]
\path (34em,26.8em) coordinate (sametop);
\path (sametop) ++(0em,-2.668em) coordinate (sparb);
\path (sametop) ++(-3em,-4.8em) coordinate (parbl);
\path (sametop) ++(3em,-4.8em) coordinate (parbr);
\node (sbot) at (sametop) {$s$};
\node [ppar] (parb) at (sparb) {};
\draw (sbot) -- (parb);
\node (npa) at (parbl) {$np$};
\node (vpb) at (parbr) {$\apsnodei$};
\draw (parb) -- (npa);
\path[>=latex,->]  (parb) edge (vpb);
\path (parbr) ++(0em,0em) coordinate (parcb);
\path (parcb) ++(0em,-2.668em) coordinate (parcc);
\path (parcb) ++(-4em,-5em) coordinate (parcl);
\path (parcb) ++(4em,-7.3em) coordinate (parcr);
\node (nn) at (parcl) {$n\ldr n$};
\node (nnnps) at (parcr) {$\apsnodei$};
\node [ppar] (prc) at (parcc) {};
\draw (vpb) -- (prc);
\draw (prc) -- (nn); 
\path[>=latex,->]  (prc) edge (nnnps);
\path (nn) ++(0em,0em) coordinate (tnnl);
\path (tnnl) ++(4em,0em) coordinate (tnnr);
\path (tnnl) ++(2em,-2.3em) coordinate (tnnc);
\path (tnnl) ++(2em,-4.8em) coordinate (tnnb);
\path (nnnps) ++(0em,0em) coordinate (nnnpsa);
\path (parcc) ++(0em,-8em) coordinate (sametp);
\path (sametp) ++(0em,-2.5em) coordinate (samebp);
\path (sametp) ++(-3em,+2.3em) coordinate (samep);
\node (same) at (samep) {\textit{same}};
\node [ttns] (st) at (sametp) {};
\node (vpr) at (samebp) {$\apsnodei$};  
\draw (st) -- (same);
\draw (st) -- (vpr);
\draw (st) -- (nnnps);
\path (samebp) ++(0em,0em) coordinate (vprp);
\path (vprp) ++(-6em,0em) coordinate (vplp);
\path (vprp) ++(-3em,-2.5em) coordinate (vpcp);
\path (vpcp) ++(0em,-2.3em) coordinate (vpbp);
\node (subj) at (vplp) {$np$};
\node [ttns] (tc) at (vpcp) {};
\node (bot) at (vpbp) {$s$};
\draw (tc) -- (bot);
\draw (tc) -- (subj);
\draw (tc) -- (vpr);
\path (parcr) ++(-12.5em,0em) coordinate (arrowc);
\path (arrowc) ++(0em,1em) coordinate (alc);
\path (arrowc) ++(-0.5em,0em) coordinate (spcc);
\node at (arrowc) {$\rightarrow$};
\node at (alc) {$\mathcal{A}$};
\node at (spcc) {$\ $};
\end{tikzpicture}
\end{center}
\caption{Proof structure and abstract proof structure of the word ``same'' according to the analysis of \citet{bs14cont}}
\label{fig:same}
\end{figure}

 Unfolding this formula produces the proof structure and abstract proof structure shown respectively on the left and right of Figure~\ref{fig:same}. To reduce complexity,  the subformula $n\backslash n$ has been not unfolded.

Translating the formula to HTLG produces the following formula. 
\[
((n\backslash n) \himpl np \himpl s) \himpl np \himpl s
\] 
By its structure, we know the prosodic type should be $(s\rightarrow s \rightarrow s) \rightarrow s \rightarrow s$. Given a constant `same' of type $s$, there are two linear lambda terms of this type (slightly more when we add constants `$+$' and `$\epsilon$'). In general, there can be many choices for the linear lambda term and choosing the correct one (to obtain a given abstract proof structure) can be tricky. This is because it is easier to evaluate a program (i.e.\ reduce a lambda term) than to construct one given its output.

In this case, it is easy to see the prosodic term should be $\lambda P. \lambda x. ((P\, \textit{same})\, x)$. The abstract proof structure for this lexical entry is shown on the left of Figure~\ref{fig:samehtlg}. Performing the two beta reductions produces the abstract proof structure shown on the right of the figure. 

\begin{figure}
\begin{center}
\begin{tikzpicture}[scale=0.75]
\path (34em,26.8em) coordinate (sametop);
\path (sametop) ++(-2.4em,0em) coordinate (pcoord);
\path (sametop) ++(2.4em,0em) coordinate (samecoord);
\path (sametop) ++(0em,-4.8em) coordinate (samebot);
\path (samebot) ++(0em, +2.668em) coordinate (sameat);
\node (p) at (pcoord) {$\apsnodei$};
\node (sc) at (samecoord) {\textit{same}};
\node [tns] (tat) at (sameat) {};
\node at (sameat) {$@$};
\node (bot) at (samebot) {$\apsnodei$};
\draw (p) -- (tat);
\draw (bot) -- (tat);
\draw (sc) -- (tat);
\path (samebot) ++(2.4em,0em) coordinate (xattop);
\path (xattop) ++(2.4em,0em) coordinate (xatr);
\path (xattop) ++(0em,-4.8em) coordinate (xatbot);
\path (xatbot) ++(0em,+2.668em) coordinate (xat);
\node [tns] (tx) at (xat) {};
\node at (xat) {$@$}; 
\node (xr) at (xatr) {$\apsnodei$};
\node (xb) at (xatbot) {$\apsnodei$};
\draw (bot) -- (tx);
\draw (tx) -- (xr);
\draw (tx) -- (xb);
\path (xatbot) coordinate (lamxtop);
\path (lamxtop) ++(0em,-2.668em) coordinate (lamx);
\path (lamxtop) ++(0em,-4.8em) coordinate (lamxbot);
\path (lamxbot) ++(-2.4em,0em) coordinate (lamxl); 
\node [tns] (lamc) at (lamx) {};
\node at (lamx) {$\lambda$};
\node (ll) at (lamxl) {$\apsnodei$};
\draw (xb) -- (lamc);
\draw (lamc) -- (ll);
\draw (lamc)..controls(41.5em,11em) and(41.5em,26em)..(xr);
\path (lamxl) coordinate (patbot);
\path (patbot) ++(0em,-2.668em) coordinate (lamp);
\path (patbot) ++(0em,-4.8em) coordinate (lampbot);
\path (lampbot) ++(-2.4em,0em) coordinate (lampl);
\node [tns] (lamc) at (lamp) {};
\node at (lamp) {$\lambda$};
\node (llp) at (lampl) {$\apsnodei$};
\draw (llp) -- (lamc);
\draw (lamc) -- (ll);
\draw (lamc)..controls(45em,2em) and(45em,36em)..(p);
\path (llp) ++(4.8em,0em) coordinate (cattop);
\path (cattop) ++(4.8em,0em) coordinate (catr);
\path (cattop) ++(0em,-4.8em) coordinate (catbot);
\path (catbot) ++(0em,+2.668em) coordinate (cat);
\node [tns] (tx) at (cat) {};
\node at (cat) {$@$}; 
\node (cr) at (catr) {$\apsnodei$};
\node (cb) at (catbot) {$\apsnodei$};
\draw (llp) -- (tx);
\draw (tx) -- (cr);
\draw (tx) -- (cb);
\path (catbot) ++(2.4em,0em) coordinate (bmid);
\path (bmid) ++(0em,-4.8em) coordinate (bot);
\path (bmid) ++(2.4em,0em) coordinate (br);
\path (bot) ++(0em,+2.668em) coordinate (bc);
\node [tns] (tb) at (bc) {};
\node at (bc) {$@$};
\node (nph) at (br) {$np$};
\node (sc) at (bot) {$s$};
\draw (sc) -- (tb);
\draw (nph) -- (tb);
\draw (tb) -- (cb);
\path (catr) coordinate (bparl);
\path (bparl) ++(2.4em,0em) coordinate (bparm);
\path (bparm) ++(2.4em,0em) coordinate (bparr);
\path (bparm) ++(0em,4.8em) coordinate (bpart);
\path (bpart) ++(0em,-2.668em) coordinate (bparc);
\node [par] (bp) at (bparc) {};
\node at (bparc) {\textcolor{white}{\textbf{$\lambda$}}};
\node (nn) at (bparr) {$n\backslash n$};
\node (cpr) at (bpart) {$\apsnodei$};
\draw (bp) -- (nn);
\draw (bp) -- (cpr);
\path[>=latex,->]  (bp) edge (cr);
\path (bpart) coordinate (parl);
\path (parl) ++(2.4em,0em) coordinate (parm);
\path (parm) ++(2.4em,0em) coordinate (parr);
\path (parm) ++(0em,4.8em) coordinate (part);
\path (part) ++(0em,-2.668em) coordinate (parc);
\node [par] (bp) at (parc) {};
\node at (parc) {\textcolor{white}{\textbf{$\lambda$}}};
\node (npc) at (parr) {$np$};
\node (sh) at (part) {$s$};
\draw (bp) -- (npc);
\draw (bp) -- (sh);
\path[>=latex,->]  (bp) edge (cpr);
\end{tikzpicture}
\begin{tikzpicture}[scale=0.75]
\path (34em,26.8em) coordinate (sametop);
\path (sametop) ++(-2.4em,1em) coordinate (pcoord);
\path (sametop) ++(2.4em,0em) coordinate (samecoord);
\path (sametop) ++(0em,-4.8em) coordinate (samebot);
\path (samebot) ++(0em, +2.668em) coordinate (sameat);
\node (p) at (pcoord) {$\apsnodei$};
\node (sc) at (samecoord) {\textit{same}};
\node [tns] (tat) at (sameat) {};
\node at (sameat) {$@$};
\node (bot) at (samebot) {$\apsnodei$};
\draw (p) -- (tat);
\draw (bot) -- (tat);
\draw (sc) -- (tat);
\path (samebot) ++(2.4em,0em) coordinate (xattop);
\path (xattop) ++(2.4em,0em) coordinate (xatr);
\path (xattop) ++(0em,-4.8em) coordinate (xatbot);
\path (xatbot) ++(0em,+2.668em) coordinate (xat);
\node [tns] (tx) at (xat) {};
\node at (xat) {$@$}; 
\node (xr) at (xatr) {$np$};
\node (xb) at (xatbot) {$s$};
\draw (bot) -- (tx);
\draw (tx) -- (xr);
\draw (tx) -- (xb);
\path (pcoord) coordinate (bparl);
\path (bparl) ++(2.4em,1em) coordinate (bparm);
\path (bparm) ++(2.4em,0em) coordinate (bparr);
\path (bparm) ++(0em,4.8em) coordinate (bpart);
\path (bpart) ++(0em,-2.668em) coordinate (bparc);
\node [par] (bp) at (bparc) {};
\node at (bparc) {\textcolor{white}{\textbf{$\lambda$}}};
\node (nn) at (bparr) {$n\backslash n$};
\node (cpr) at (bpart) {$\apsnodei$};
\draw (bp) -- (nn);
\draw (bp) -- (cpr);
\path[>=latex,->]  (bp) edge (p);
\path (bpart) coordinate (parl);
\path (parl) ++(2.4em,0em) coordinate (parm);
\path (parm) ++(2.4em,0em) coordinate (parr);
\path (parm) ++(0em,4.8em) coordinate (part);
\path (part) ++(0em,-2.668em) coordinate (parc);
\node [par] (bp) at (parc) {};
\node at (parc) {\textcolor{white}{\textbf{$\lambda$}}};
\node (npc) at (parr) {$np$};
\node (sh) at (part) {$s$};
\draw (bp) -- (npc);
\draw (bp) -- (sh);
\path[>=latex,->]  (bp) edge (cpr);
\path (pcoord) ++(-3em,0em) coordinate (arrowc);
\path (arrowc) ++(0em,1em) coordinate (alc);
\node at (arrowc) {$\rightarrow$};
\node at (alc) {$\beta \times 2$};
\end{tikzpicture}
\end{center}
\caption{The analysis of ``same'' from \citet{bs14cont} translated into HTLG.}
\label{fig:samehtlg}
\end{figure}

The abstract proof structure on the right of Figure~\ref{fig:samehtlg} is again the left-right symmetric version of the abstract proof structure for ``same'' in $\nllam$ shown in Figure~\ref{fig:same}. 

\subsection{Dutch verb clusters}
\label{sec:dutch}

One well-studied topic in linguistics and  formal language theory are the crossed dependencies which occur for verb clusters and their objects in Dutch relative clauses. 
The complexity of the phenomenon is illustrated by the famous `hippopotamus' sentences such as the following.

\exg.(dat) Jan Marie de nijlpaarden zag voeren\\
  (that) Jan Marie the hippopotami saw  feed\\
  `(That) John saw Marie feed the hippopotami' 

\exg.\label{ex:hippo}(dat) Jan Henk Marie de nijlpaarden zag helpen voeren\\
  (that) Jan Henk Marie the hippopotami saw help feed\\
  `(That) John saw Henk help Marie feed the hippopotami' 

The key point of Sentence~\ref{ex:hippo} is the ``de nijlpaarden'' (the hippopotami) is the object of ``voeren'' (feed), ``Marie'' the object of ``helpen'' (help), and ``Henk'' the object of ``zag'' (saw), leading to `crossed' dependencies between the verbs and their objects, which is essential to produce the right meaning under the standard (minimal) type-logical assumptions of the syntax-semantic interface. 

Unlike the previous cases, we cannot directly apply Lemma~\ref{lem:nllamhtlg} here: the HTLG lexical items shown below crucially use lexical lambda terms which cannot be reduced directly in the abstract proof structures corresponding to the lexical unfolding.

However, there is a work-around for this problem.
Given the following \nllam{} lexicon, we generate exactly the correct readings.  
\newcommand{\inspoint}{j}
\newcommand{\inspointd}{J}
\newcommand{\infv}{\textit{inf}}
\newcommand{\subv}{s_\textit{sub}}
\newcommand{\ddl}[2]{#1 \mathbin{\backslash}_w #2}
\begin{align*}
\textit{dat} &\quad s_{\textit{that}}\ldr \subv \\
\textit{Jan} &\quad np	\\
\textit{Henk} &\quad np	\\
\textit{Marie} &\quad np	\\
\textit{de} &\quad np \ldr n \\
\textit{nijlpaarden} &\quad n \\
\textit{zag} &\quad (np\ldl(np\ldl \subv))\bdr (\inspoint\bdl\infv) \\
\textit{helpen} &\quad \inspoint\ldl((np\ldl\infv)\bdr (\inspoint\bdl\infv))   \\
\textit{voeren} &\quad \inspoint\ldl (np\ldl\infv)
\end{align*}
\newcommand{\denijlp}{\textit{dn}}
The key property is that we use a special atomic formula $\inspoint$ to mark a point for future extraction. In the lexical entry for ``voeren'', when combined with all its arguments will produce the structure  $(\textit{de} \lstr \textit{nijlpaarden}) \lstr (\inspoint \lstr \textit{voeren})$ with $\inspoint$ the left sister of $\textit{voeren}$. The following proof shows how we combine this phrase with ``Marie'' and ``helpen'' (to save space, we have abbreviated $(\textit{de} \lstr \textit{nijlpaarden})$ by $\denijlp$).
\[
\infer[\betared]{\textit{Marie} \lstr (\denijlp \lstr ((\inspoint \lstr \textit{helpen}) \lstr \textit{voeren}))\vdash \infv}{
\infer[\ldl L]{\textit{Marie} \lstr ((\inspoint \lstr \textit{helpen}) \bstr \lambda x. \denijlp \lstr (x \lstr \textit{voeren}))\vdash \infv}{
\infer[\textit{Ax}]{j\vdash j}{} &
\infer[\bdr L]{\textit{Marie} \lstr (((np\ldl\infv)\bdr (\inspoint\bdl\infv)) \bstr \lambda x. \denijlp \lstr (x \lstr \textit{voeren}))\vdash \infv}{
    \infer[\bdl R]{\lambda x. \denijlp \lstr (x \lstr \textit{voeren})\vdash j\bdl \infv}{
   \infer[\betaexp]{\inspoint \bstr \lambda x. \denijlp \lstr (x \lstr \textit{voeren})}{
      \infer*{\denijlp \lstr (\inspoint \lstr \textit{voeren})\vdash \infv}{}}}
    & \infer*{\textit{Marie} \lstr np\ldl\infv\vdash \infv}{}
      }}}
\]
We can see that ``Marie'' is concatenated before ``de nijlpaarden'' whereas the insertion point `$\inspoint$' is replaced by ``$\inspoint \lstr$ helpen'', effectively putting ``helpen'' before ``voeren'' and creating a new insertion point before ``helpen''. 

Reading the proof as backward chaining proof search, we start by moving $(\inspoint \lstr \textit{helpen})$ between ``Marie'' and ``de nijlpaarden''. We then use the $\ldl L$ rule to combine ``helpen'' with its $\inspoint$ argument. This produces the subproof $j\vdash j$ and replaces $(\inspoint \lstr \textit{helpen})$ by $ ((np\ldl\infv)\bdr (\inspoint\bdl\infv)$ in the other proof branch. In that branch, we can immediately apply the $\bdr L$ rule. The right branch is a trivial \nlambek{} derivation.  The left branch uses the standard combination of $\bdl R$ and $\betared$ to move the $\inspoint$ formula in the place from where we moved out $(\inspoint \lstr \textit{helpen})$ at the start of the proof. We can then complete the proof using the \nlambek{} rules.

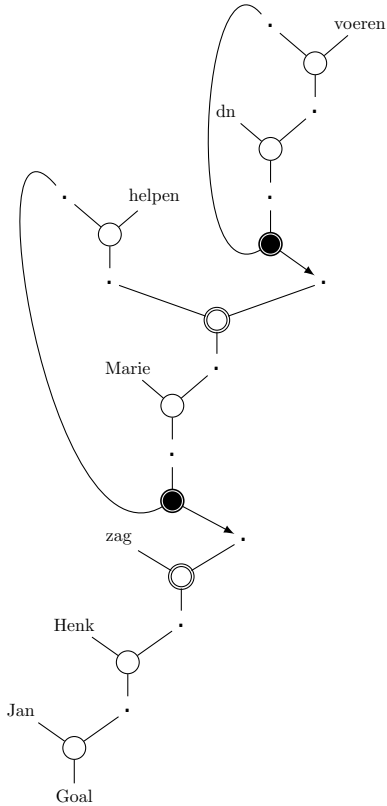
\begin{figure}
\begin{center}
\begin{tikzpicture}[scale=0.75]
\path (13em,0em) coordinate (sametop);
\path (sametop) ++(0em,-2.132em) coordinate (sparb);
\path (sametop) ++(-6em,0em) coordinate (tnstl);
\path (sametop) ++(6em,0em) coordinate (tnstr);
\path (sametop) ++(0em,-4.8em) coordinate (parbr);
\node [ttns] (tth) at (sparb) {};
\node (helpnpj) at (tnstl) {$\apsnodei$};
\node (helpjinf) at (tnstr) {$\apsnodei$};
\node (vpb) at (parbr) {$\apsnodei$};
\draw (tth) -- (vpb);
\draw (tth) -- (helpnpj);
\draw (tth) -- (helpjinf);
\path (parbr) ++(0em,0em) coordinate (trgt);
\path (trgt) ++(-2.5em,-4.8em) coordinate (tbot);
\path (tbot) ++(-2.5em,4.8em) coordinate (tlft);
\path (tbot) ++(0em,2.668em) coordinate (tcn);
\node [tns] (tth2) at (tcn) {};
\node (marie) at (tlft) {\text{Marie}};
\node (jh) at (tbot) {$\apsnodei$};
\draw (marie) -- (tth2);
\draw (tth2) -- (jh);
\draw (tth2) -- (vpb);
\path (tnstl) ++(0em,0em) coordinate (ttbot);
\path (ttbot) ++(0em,2.668em) coordinate (tcn);
\path (ttbot) ++(-2.5em,4.8em) coordinate (tlft);
\path (ttbot) ++(2.5em,4.8em) coordinate (trgt);
\node [tns] (tth3) at (tcn) {};
\draw (tth3) -- (helpnpj);
\node (j1) at (tlft) {$\apsnodei$};
\node (helpen) at (trgt) {\text{helpen}};
\draw (tth3) -- (j1);
\draw (tth3) -- (helpen);
\path (tbot) ++(0em,0em) coordinate (parcb);
\path (parcb) ++(0em,-2.668em) coordinate (parcc);
\path (parcb) ++(-4em,-5em) coordinate (parcl);
\path (parcb) ++(4em,-4.8em) coordinate (parcr);
\node (nnnps) at (parcr) {$\apsnodei$};
\node [ppar] (prc) at (parcc) {};
\draw (jh) -- (prc);
\path[>=latex,->]  (prc) edge (nnnps);
\draw (j1) to [out=130,in=210] (prc);
\path (parcr) ++(0em,0em) coordinate (tnr);
\path (tnr) ++(-7em,0em) coordinate (tnl);
\path (parcr) ++(-3.5em,-4.8em) coordinate (tnb);
\path (tnb) ++(0em,2.686em) coordinate (tnc);
\node (ts) [ttns] at (tnc) {};
\node (npbs) at (tnb) {$\apsnodei$};
\node (zag) at (tnl) {\text{zag}};
\draw (ts) -- (npbs);
\draw (ts) -- (zag);
\draw (ts) -- (nnnps);
\path (tnb) ++(0em,0em) coordinate (tr);
\path (tr) ++(-3em,-4.8em) coordinate (tb);
\path (tr) ++(-6em,0em) coordinate (tl);
\path (tb) ++(0em,2.686em) coordinate (tc);
\node (hnp) at (tl) {\text{Henk}};
\node (t) [tns] at (tc) {};
\node (ss) at (tb) {$\apsnodei$};
\draw (t) -- (ss);
\draw (t) -- (hnp);
\draw (t) -- (npbs);
\path (tb) ++(0em,0em) coordinate (tr);
\path (tr) ++(-3em,-4.8em) coordinate (tb);
\path (tr) ++(-6em,0em) coordinate (tl);
\path (tb) ++(0em,2.686em) coordinate (tc);
\node (jnp) at (tl) {\text{Jan}};
\node (t) [tns] at (tc) {};
\node (g) at (tb) {\text{Goal}};
\draw (t) -- (ss);
\draw (t) -- (jnp);
\draw (t) -- (g);
\path (helpjinf) ++(0em,0em) coordinate (ppr);
\path (ppr) ++(-3em,2.131em) coordinate (ppc);
\path (ppr) ++(-3em,4.8em) coordinate (ppt);
\node [ppar] (parh) at (ppc) {};
\path[>=latex,->]  (parh) edge (helpjinf);
\node (vinf) at (ppt) {$\apsnodei$};
\draw (parh) -- (vinf);
\path (ppt) ++(0em,0em) coordinate (tbot);
\path (tbot) ++(0em,2.668em) coordinate (tcn);
\path (tbot) ++(-2.5em,4.8em) coordinate (tlft);
\path (tbot) ++(2.5em,4.8em) coordinate (trgt);
\node [tns] (tth2) at (tcn) {};
\node (marie) at (tlft) {\text{dn}};
\node (jh) at (trgt) {$\apsnodei$};
\draw (marie) -- (tth2);
\draw (tth2) -- (jh);
\draw (tth2) -- (vinf);
\path (trgt) ++(0em,0em) coordinate (tbot);
\path (tbot) ++(0em,2.668em) coordinate (tcn);
\path (tbot) ++(-2.5em,4.8em) coordinate (tlft);
\path (tbot) ++(2.5em,4.8em) coordinate (trgt);
\node [tns] (tth3) at (tcn) {};
\draw (tth3) -- (jh);
\node (j2) at (tlft) {$\apsnodei$};
\node (helpen) at (trgt) {\text{voeren}};
\draw (tth3) -- (j2);
\draw (tth3) -- (helpen);
\draw (j2) to [out=130,in=210] (parh);
\end{tikzpicture}
\end{center}
\caption{Abstract proof structure for the \nllam{} analysis of ``(dat) Jan Henk Marie de nijlpaarden zag helpen voeren''.}
\label{fig:hippo}
\end{figure}

Figure~\ref{fig:hippo} shows the only abstract proof structure for Example~\ref{ex:hippo} (at least the only one which produces the correct noun phrase order). We leave the reader to verify that two applications of the $\betaexp\bdl$ conversion and two applications of the $\betared$ conversion produce a tree with the right yield. 


This analysis is extremely close to the Displacement calculus analysis of \citet[Section~3.2.8]{mvf11displacement} when we translate $A\bdr (\inspoint\bdl B)$ by $\ddl{B}{A}$ and $\inspoint$ by $\inspointd$.
\begin{align*}
\textit{zag} &\quad \ddl{\infv}{(np\ldl(np\ldl \subv)} \\
\textit{helpen} &\quad \inspointd\ldl (\ddl{\infv}{(np\ldl\infv)})   \\
\textit{voeren} &\quad \inspointd\ldl (np\ldl\infv)
\end{align*}
The key logical rule which makes this analysis work is the following\footnote{The rule presented here is simplified from the rule of \citet{mvf11displacement}. However, this simplification does not affect the analysis.}.
\[
\infer[\ddl L]{\Gamma_1,\Delta_1,\ddl{A}{C},\Delta_2,\Gamma_2\vdash C}{\Delta_1,\inspointd,\Delta_2\vdash A & \Gamma_1,C,\Gamma_2\vdash C}
\]
The $\ddl L$ rule allows a formula $\ddl{A}{C}$ to select the $\Delta_1$ and $\Delta_2$ structures which surround it, while marking the separation between the two with $\inspointd$. In the Displacement calculus, the \nllam{} proof above looks as follows.
\[
\infer[\ldl L]{\textit{Marie},\denijlp,\inspointd,\textit{helpen},\textit{voeren}\vdash \infv}{
   \infer[\textit{Ax}]{\inspointd\vdash\inspointd}{}
&  \infer[\ddl L]{\textit{Marie},\denijlp,\ddl{\infv}{(np\ldl\infv)},\textit{voeren}\vdash \infv}{
      \infer*{\denijlp,\inspointd,\textit{voeren}\vdash \infv}{}
     &\infer*{\textit{Marie},np\ldl\infv\vdash \infv}{}
   }
}
\]

The comparison with hybrid type-logical grammars is also instructive. The following lexicon allows HTLG to analyse the Dutch verbs (the other lexical entries stay the same). The atomic type $\infv$ is assigned the complex prosodic type $\str\rightarrow\str$. The abstracted variable $v$ plays the same role as the $\inspoint$ atomic formula in \nllam{} and as the $\inspointd$ formula in the Displacement calculus.

\medskip
\begin{tabular}{llll}
\emph{Word} & \emph{Syntactic type} & 
\emph{Prosodic term} \\[0.6ex]
zag & 
$\infv\himpl np \himpl np \himpl s$ &  
 $\lambda P^{\str\rightarrow\str}\lambda x^{\str}\lambda y^{\str}. x+y+(P\, \textit{zag})$ \\
helpen & 
$\infv \himpl (np \himpl \infv)$ &
$\lambda P^{\str\rightarrow\str} \lambda x^{\str}. \lambda v^{\str}\  x+P(v+\textit{helpen})$ \\
voeren & 
$np \himpl \infv$ & 
$\lambda x \lambda v. x+v+\textit{voeren}$ \\
\end{tabular}
\medskip

With these lexical entries, the phrase ``de nijlpaarden voeren'' is assigned the term $\lambda v. \textit{de}+\textit{nijlpaarden}+v+\textit{voeren}$. Then giving ``helpen'' this infinitive and ``Marie'' as arguments produces a term which normalises to the following. 
\[
\lambda v. \textit{Marie}+\textit{de}+\textit{nijlpaarden}+v+\textit{helpen}+\textit{voeren}
\]
Finally, applying this term and ``Jan'' and ``Henk'' to ``zag'' produces the required string.

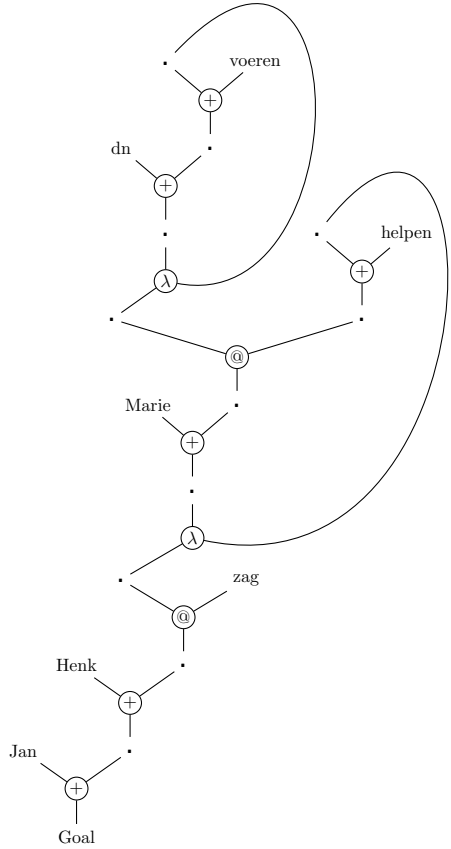
\begin{figure}
\begin{center}
\begin{tikzpicture}[scale=0.75]
\path (13em,0em) coordinate (sametop);
\path (sametop) ++(0em,-2.132em) coordinate (sparb);
\path (sametop) ++(7em,0em) coordinate (tnstl);
\path (sametop) ++(-7em,0em) coordinate (tnstr);
\path (sametop) ++(0em,-4.8em) coordinate (parbr);
\node [tns] (tth) at (sparb) {};
\node at (sparb) {$@$};
\node (helpnpj) at (tnstl) {$\apsnodei$};
\node (helpjinf) at (tnstr) {$\apsnodei$};
\node (vpb) at (parbr) {$\apsnodei$};
\draw (tth) -- (vpb);
\draw (tth) -- (helpnpj);
\draw (tth) -- (helpjinf);
\path (parbr) ++(0em,0em) coordinate (trgt);
\path (trgt) ++(-2.5em,-4.8em) coordinate (tbot);
\path (tbot) ++(-2.5em,4.8em) coordinate (tlft);
\path (tbot) ++(0em,2.668em) coordinate (tcn);
\node [tns] (tth2) at (tcn) {};
\node at (tcn) {$+$};
\node (marie) at (tlft) {\text{Marie}};
\node (jh) at (tbot) {$\apsnodei$};
\draw (marie) -- (tth2);
\draw (tth2) -- (jh);
\draw (tth2) -- (vpb);
\path (tnstl) ++(0em,0em) coordinate (ttbot);
\path (ttbot) ++(0em,2.668em) coordinate (tcn);
\path (ttbot) ++(-2.5em,4.8em) coordinate (tlft);
\path (ttbot) ++(2.5em,4.8em) coordinate (trgt);
\node [tns] (tth3) at (tcn) {};
\node at (tcn) {$+$};
\draw (tth3) -- (helpnpj);
\node (j1) at (tlft) {$\apsnodei$};
\node (helpen) at (trgt) {\text{helpen}};
\draw (tth3) -- (j1);
\draw (tth3) -- (helpen);
\path (tbot) ++(0em,0em) coordinate (parcb);
\path (parcb) ++(0em,-2.668em) coordinate (parcc);
\path (parcb) ++(-4em,-5em) coordinate (parcl);
\path (parcb) ++(4em,-4.8em) coordinate (parcr);
\node (nnnps) at (parcl) {$\apsnodei$};
\node [tns] (prc) at (parcc) {};
\node at (parcc) {$\lambda$};
\draw (jh) -- (prc);
\draw (prc) -- (nnnps);
\draw (prc)..controls(28em,-16em) and(28em,18em)..(j1);
\path (parcl) ++(0em,0em) coordinate (tnl);
\path (tnl) ++(+7em,0em) coordinate (tnr);
\path (parcl) ++(+3.5em,-4.8em) coordinate (tnb);
\path (tnb) ++(0em,2.686em) coordinate (tnc);
\node (ts) [tns] at (tnc) {};
\node at (tnc) {$@$};
\node (npbs) at (tnb) {$\apsnodei$};
\node (zag) at (tnr) {\text{zag}};
\draw (ts) -- (npbs);
\draw (ts) -- (zag);
\draw (ts) -- (nnnps);
\path (tnb) ++(0em,0em) coordinate (tr);
\path (tr) ++(-3em,-4.8em) coordinate (tb);
\path (tr) ++(-6em,0em) coordinate (tl);
\path (tb) ++(0em,2.686em) coordinate (tc);
\node (hnp) at (tl) {\text{Henk}};
\node (t) [tns] at (tc) {};
\node at (tc) {$+$};
\node (ss) at (tb) {$\apsnodei$};
\draw (t) -- (ss);
\draw (t) -- (hnp);
\draw (t) -- (npbs);
\path (tb) ++(0em,0em) coordinate (tr);
\path (tr) ++(-3em,-4.8em) coordinate (tb);
\path (tr) ++(-6em,0em) coordinate (tl);
\path (tb) ++(0em,2.686em) coordinate (tc);
\node (jnp) at (tl) {\text{Jan}};
\node (t) [tns] at (tc) {};
\node at (tc) {$+$};
\node (g) at (tb) {\text{Goal}};
\draw (t) -- (ss);
\draw (t) -- (jnp);
\draw (t) -- (g);
\path (helpjinf) ++(0em,0em) coordinate (ppr);
\path (ppr) ++(3em,2.131em) coordinate (ppc);
\path (ppr) ++(3em,4.8em) coordinate (ppt);
\node [tns] (parh) at (ppc) {};
\node at (ppc) {$\lambda$};
\draw (parh) -- (helpjinf);
\node (vinf) at (ppt) {$\apsnodei$};
\draw (parh) -- (vinf);
\path (ppt) ++(0em,0em) coordinate (tbot);
\path (tbot) ++(0em,2.668em) coordinate (tcn);
\path (tbot) ++(-2.5em,4.8em) coordinate (tlft);
\path (tbot) ++(2.5em,4.8em) coordinate (trgt);
\node [tns] (tth2) at (tcn) {};
\node at (tcn) {$+$};
\node (marie) at (tlft) {\text{dn}};
\node (jh) at (trgt) {$\apsnodei$};
\draw (marie) -- (tth2);
\draw (tth2) -- (jh);
\draw (tth2) -- (vinf);
\path (trgt) ++(0em,0em) coordinate (tbot);
\path (tbot) ++(0em,2.668em) coordinate (tcn);
\path (tbot) ++(-2.5em,4.8em) coordinate (tlft);
\path (tbot) ++(2.5em,4.8em) coordinate (trgt);
\node [tns] (tth3) at (tcn) {};
\node at (tcn) {$+$};
\draw (tth3) -- (jh);
\node (j2) at (tlft) {$\apsnodei$};
\node (helpen) at (trgt) {\text{voeren}};
\draw (tth3) -- (j2);
\draw (tth3) -- (helpen);
\draw (parh)..controls(20em,0em) and(20em,26em)..(j2);
\end{tikzpicture}
\end{center}
\caption{Abstract proof structure for the HTLG analysis of ``(dat) Jan Henk Marie de nijlpaarden zag helpen voeren''.}
\label{fig:hippohtlg}
\end{figure}

Figure~\ref{fig:hippohtlg} shows the HTLG abstract proof structure for ``(dat) Jan Henk Marie de nijlpaarden zag helpen voeren'', after some $\beta$ reductions have simplified the structure. We can again see the similarity with the \nllam{} abstract proof structure of Figure~\ref{fig:hippo}: applying the $\betaexp\bdl$ conversion twice to the structure of Figure~\ref{fig:hippo} produces a structure which is equivalent under the now standard left-right symmetry of the `$\lambda$' and `$@$' links with respect to the corresponding \nllam{} links. 

However, there is an important difference in the two analyses: 
the $\betaexp\bdl$ rewrite, which produces the $\lambda$ link in \nllam{}, requires the $\bdl R$ par link (it is produced by the positive $\inspoint\bdl\infv$ subformula of ``zag'' and ``helpen''; the topmost par link is part of the lexical entry for ``helpen'' and the bottom par link is par of the lexical entry for ``zag'') whereas in the HTLG analysis $\lambda$ links come from the lexicon (the topmost lambda link is part of the lexical entry for ``voeren'', the bottom lambda link of the lexical entry for ``helpen''). So although the abstract proof structures end up as equivalent, the lexical entries divide the links in different ways.

\subsection{Differences and untranslatables}


Having the residuals of the continuation mode (and the formula corresponding to the empty string)  allows $\nllam${} to have a simple treatment of the across-the-board cases of extraction such as the following.

\ex.\label{ex:atb} Which book does Peter like $\_$ but Mary hate $\_$?

Sentences like~\ref{ex:atb} are grammatical only when the two coordinated phrases ``Peter like'' and ``Mary hate'' both miss the same material (here, both sentences are incomplete for a noun phrase).
In \nllam{} we 
 can simply use the standard formula for ``and'' $(X\ldl X)\ldr X$ with $X= t \bpr (np \bdl s)$. The left rule for the product of the continuation mode (corresponding to its par link) and the left rule for $t$ (again a par link) ensure this works correctly. This analysis can therefore not be directly adapted to HTLG, but unlike the Dutch verb cluster case of Section~\ref{sec:dutch}, there doesn't appear to be a work-around either. 
 
 Conversely, lexical entries in HTLG using `irreducible' lambda terms (that is, lambda terms which cannot be $\beta$-reduced at the level of the formula unfolding) need not allow a reanalysis like the Dutch verb cluster example, and this would point to an advantage of HTLG over \nllam{}. 

One conclusion to draw from all this is that there appears to be a `common core' to modern type-logical grammars (including HTLG, $\nllam$ and the Displacement calculus) which, even though it is a sort of least common denominator, already handles quite a few sophisticated phenomena.

Where a logic allows for operations outside of this common core, and where a direct translation is therefore impossible, provides a useful starting point for discovering potential advantages of one system over another.

Another way to resolve the tension between HTLG and \nllam{} is to \emph{combine} the two logics into a single, convergent framework (a doubly-hybrid type-logical grammar). This would move HTLG into a more standard multimodal setting, where all connectives are part of a residuated triple, and give it an analysis of across-the-board phenomena. From the \nllam{} perspective, this would give \nllam{} access to lambda terms at the level of the lexicon as well, and allow it to use the HTLG analyses from \cite{kl20tls} directly. 

\section{Formal language}
\label{sec:fl}

%
%
%

We have seen in Section~\ref{sec:dutch} how \nllam{} can handle Dutch verb clusters. A natural follow-up question is: what class of formal languages can be generated by \nllam{}. In this section we provide some preliminary answers to this question.  
 We assume the reader has some basic familiarity  with mildly context-sensitive formalisms \citep{conv,mcfg,k10parsing}.

It is well-know that a wrapping operator as defined for the analysis of crossing dependencies in Section~\ref{sec:dutch} suffices for generating (at least) the simplest class of the mildly context-sensitive languages \citep{headtag,conv} --- the tree adjoining languages or the languages generated by well-nested $\text{2-MCFL}$ \citep{conv,mcfg}. This means that in addition to giving an analysis of Dutch verb clusters, we have also shown that \nllam{} generates at least the weakest class of the mildly context-sensitive languages.

\begin{lemma}
\nllam{} generates mildly context-sensitive languages	
\end{lemma}
With enough separation symbols $\inspoint_1, \ldots, \inspoint_n$ the construction of Section~\ref{sec:dutch} can be extended to generate the well-nested mildly context-sensitive languages\footnote{When desired, multiple occurrences of the same symbol $\inspoint_i$ can be used to emulate the non-deterministic wrapping operation of \citet[Section~3.4]{mvf11displacement}}. I am unsure whether this construction can be extended to generate non-wellnested languages; this would require us the extend the methodology used for Dutch verb clusters to much more complicated lexical lambda terms. 

There is also a trivial upper bound on the language class generated.

\begin{lemma} The languages generated by \nllam{} are included in the context-sensitive languages.
\end{lemma}

\paragraph{Proof} Given that the context-sensitive languages correspond to linear bounded Turing machines \citep{hu79}, and  Lemma~\ref{lem:dec} establishes a linear space bound on parsing, this is trivial.  \qed

However, this leaves a rather large distance between the lower- and the upper bounds and it seems neither bound is tight.

Like most other type-logical grammars, \nllam{} can also generate some languages which may not be mildly context-sensitive, namely those allowing some permutation closure, such as scrambling \citep{scramble} and the generalised MIX languages. This last class is the language of all permutations of $(a_1 a_2 \ldots a_k)^+$ for a given $k$ (with $k=3$ for the standard MIX language). 

\citet{emmsextr} shows that given a type-logical grammar handling extraction, we extend any grammar written in this formalism to generate the permutation closure of its original language. This holds for \nllam{} as well. 

\begin{lemma}\label{lem:perm} Let $g$ be an \nllam{} grammar generating a language $L$. There is an \nllam{} grammar $g'$ which generates the permutation closure of $L$.
\end{lemma}

\paragraph{Proof} The proof of \citet{emmsextr} adapts without problem to \nllam{}.
Assume a unique goal formula $s$. For each lexical entry assigning formula $A$ to word $w$, add an additional lexical entry $s\ldr (t\bpr (A\bdl s))$ (this is a standard `topicalisation' lexical entry). This lexical entry allows us to derive the sentence as before, using the $A$ subformula, and then move $w$ to the first position as follows (we can remove `$\lidstr$' from the end-sequent if desired, using either the $\lidstr\lstr$ or the $\lstr\lidstr$ structural rule). 
\[
\infer[\bdr L]{s\ldr (t\bpr (A\bdl s)) \lstr \Gamma[\lidstr]\vdash s}{
\infer[\textit{Ax}]{s \proofspace\vdash s}{}
& \infer[\betared]{\Gamma[\lidstr]\vdash t\bpr (A\bdl s)}{
\infer[\bpr R]{\lidstr \bstr\lambda x.\Gamma[x]\vdash t\bpr (A\bdl s)}{
\infer[\lidc R]{\lidstr\vdash \lidc}{} &
\infer[\bdl R]{\lambda x.\Gamma[x]\vdash A\bdl s}{
   \infer[\betaexp]{A \bstr\lambda x.\Gamma[x]\vdash s}{\infer*{\Gamma[A]\vdash s}{}}
}}}}
\]
By moving the words to the front, from last to first with respect to the desired order, we can generate any permutation of the original sentence. \qed

\begin{lemma}\nllam{} generates the $\text{MIX}_k$ languages for all $k$
\end{lemma}

\paragraph{Proof} We can generate  $(a_1 a_2 \ldots a_k)^+$ using the following lexicon.
\begin{align*}
a_1 &\quad ((s\ldr t_k)\ldots \ldr t_2) \\
a_1 &\quad ((s\ldr t_k)\ldots \ldr t_2)\ldr s \\
a_2 &\quad t_2 \\
\ldots &\quad \ldots \\
a_n &\quad t_k
\end{align*}
Lemma~\ref{lem:perm} gives us the permutation closure of this language as required. \qed

$\text{MIX}_3$ is known to be a mildly context-sensitive language --- it is a $\text{2-MCFL}$ \citep{s15mix} but not well-nested \citep{ks12mix}. However, the precise place of the generalised MIX languages in the standard inclusion hierarchy of formal languages used for computational linguistics \citep{k10parsing} is not known. There are simple ways to derive them in quite expressive formalisms outside of the mildly context-sensitive classes such as Range Concatenation Grammars \citep{boullier99mix,k10parsing}. 
It therefore appears that the classes of languages generated by \nllam{} (and many other type-logical grammars) fall outside the standard language-theoretic classes. For example, the formalisms which can handle the generalised MIX languages also handle the language which, for all $n$, generates exactly  $2^n$ occurrences of $a$ \citep{k10parsing}, and this `exponential' language appears outside of what most type-logical grammars can handle: generating $2^n$ occurrences of a symbol strongly suggests a formalism-internal mechanism for copying, which contradicts the multiplicative nature of type-logical grammars.

With respect to the formalisms whose language classes do not yet have a precise place in the standard hierarchy, some of the formalisms which provide non-local permutation mechanisms, such as multi-set valued linear indexed grammars and  some extensions of tree adjoining grammars \citep{rambo,beckerphd} are candidates for formalisms deriving the same language classes as type-logical grammars\footnote{There are a number of technical difficulties adapting the proof of \citet{pentus97} --- which shows the Lambek calculus generates exactly the context-free languages --- to other type-logical grammars, as discussed by \citet[Section~3.2]{Bus97} and \citet[Section~3]{mr19nls}.}. This would point to a refinement of the standard picture, possibly adding the presence and absence of permutation closure as a parameter comparable to the presence and absence of well-nestedness. Alternatively, it may turn out that these language classes correspond to one of the classes in the standard hierarchy\footnote{Extending the proof of \citet{s15mix} to the generalised MIX languages would similarly require a solution to many technical difficulties \citep{nederhof16short}.}.

Concluding, although we have show that \nllam{} can generate (well-nested) mildly context-sensitive languages and the permutation closure of any context-free language, these are only lower bounds on the language class generated by the formalism. As with many other type-logical grammars, there is no precise upper bound (other than the context-sensitive languages). 
A precise formal characterisation of the language classes generated by type-logical grammars remains an important open question.

\section{Conclusion}

We have investigated the proof theoretic aspects of \nllam{} by introducing  a proof net calculus for the logic, and proving proof nets are  sound and complete with respect to the standard sequent calculus presentation. Thanks to this proof net calculus, we have also shown that excluding the empty antecedent means \nllam{} proof search is in NP, an improvement over the previous upper bound. Finally, we have shown there to be a surprising convergence between the linguistic analyses in \nllam{} and those in hybrid type-logical grammars. This extends the empirical coverage of \nllam{} and gives a first analysis of the logic in terms of the class of formal languages generated.

\appendix
\section{Treating the unit as an atom}
\label{sec:unitatom}

In proof theory, there are generally two ways to treat the logical constants like `$\top$' and `$\bot$' in classical logic and `$\textbf{1}$' and `$\bot$' in multiplicative linear logic: we can treat them as 0-ary connectives, or we can treat them as atomic formulas with special rules.

In the above, we have treated the unit `$\lidc$' as a 0-ary connective. In this section we briefly present what it would entail to treat `$\lidc$' as an atomic formula. If $t$ is treated as an atomic formula, it neither has its own logical links\footnote{In the abstract proof structure, we will still have the link for its structural connvective `$\lidstr$'.} nor its own contractions. Formula unfolding would simply stop like for any other atomic formula. Instead of the contractions, we have the two structural rules shown below.

\begin{center}
\begin{tikzpicture}[scale=0.75]
\node at (19em,22em) {$\apsnode{t}{c}$};
\node at (25.0em,22.0em) {$\longleftarrow$}; 
\node at (25.0em,21.0em) {$t^{-1}$};
\node at (25.0em,23em) {$\longrightarrow$}; 
\node at (25.0em,24em) {$t$};
\node at (32em,22em) {$\apsnode{}{c}$};
\node (cc) at (32em,22em) {\emptynode};
\node[tns] (tid) at (32em,24.4em) {};
\node at (32em,24.4em) {\lidstr};
\draw (cc) -- (tid);
\end{tikzpicture}
\end{center}

The structural rules simply allow us to rewrite a `$\lidc$' atomic formula (occurring as a hypothesis) for its structural connective `$\lidstr$' and vice versa. While this works as it should, the fact that the $\lidc^{-1}$ rule introduces a new atomic formula means that we lose the clear separation of stages: first unfolding the formulas, then linking the axioms, and finally rewriting. The $\lidc^{-1}$ can require us to have an axiom connection as a rewrite step on the abstract proof structure, for example for the following proof of $A \vdash A\ldr\lidc$. We start from the unique abstract proof structure obtained by unfolding the formulas, identifying the $A$ formulas and translating the resulting proof structure to its abstract proof structure.
The result is shown below as the leftmost abstract proof structure.

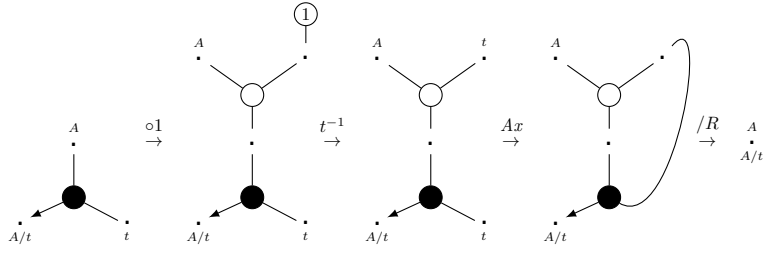
\begin{center}
\begin{tikzpicture}[scale=0.75]
\node (pa) at (10em,0) {$\apsnode{}{A\ldr \lidc}$};
\node (pat) at (10.2em,0.44em) {};
\node[par] (pc) at (13em,1.732em) {};
\node (pb) at (16em,0.15em) {$\apsnode{}{\lidc}$};
\node (pd) at (13em,4.8em) {$\apsnode{A}{}$};
\node (pdd) at (13em,4.5em) {};
\draw (pc) -- (pb);
\draw (pc) -- (pdd);
\path[>=latex,->]  (pc) edge (pat);
\node (pa) at (20em,0) {$\apsnode{}{A\ldr \lidc}$};
\node (pat) at (20.2em,0.44em) {};
\node[par] (pc) at (23em,1.732em) {};
\node (pb) at (26em,0.15em) {$\apsnode{}{\lidc}$};
\node (pd) at (23em,4.8em) {$\apsnodei$};
\node (pdd) at (23em,4.5em) {};
\draw (pc) -- (pb);
\draw (pc) -- (pdd);
\path[>=latex,->]  (pc) edge (pat);
\node[tns] (tns) at (23em,7.468em) {};
\node (newa) at (20em,9.6em) {$\apsnode{A}{}$};
\node (epsbot) at (26em,9.6em) {$\apsnodei$};
\draw (pd) -- (tns);
\draw (tns) -- (epsbot);
\draw (tns) -- (newa);
\node [tns] (1) at (26em,12em) {};
\node (tl) at (26em,12em) {$\lidstr$};
\draw (1) -- (epsbot);
\node (pa) at (30em,0) {$\apsnode{}{A\ldr \lidc}$};
\node (pat) at (30.2em,0.44em) {};
\node[par] (pc) at (33em,1.732em) {};
\node (pb) at (36em,0.15em) {$\apsnode{}{\lidc}$};
\node (pd) at (33em,4.8em) {$\apsnodei$};
\node (pdd) at (33em,4.5em) {};
\draw (pc) -- (pb);
\draw (pc) -- (pdd);
\path[>=latex,->]  (pc) edge (pat);
\node[tns] (tns) at (33em,7.468em) {};
\node (newa) at (30em,9.6em) {$\apsnode{A}{}$};
\node (epsbot) at (36em,9.6em) {$\apsnode{t}{}$};
\draw (pd) -- (tns);
\draw (tns) -- (epsbot);
\draw (tns) -- (newa);
\node (pa) at (40em,0) {$\apsnode{}{A\ldr \lidc}$};
\node (pat) at (40.2em,0.44em) {};
\node[par] (pc) at (43em,1.732em) {};
\node (pd) at (43em,4.8em) {$\apsnodei$};
\node (pdd) at (43em,4.5em) {};
\draw (pc) -- (pdd);
\path[>=latex,->]  (pc) edge (pat);
\node[tns] (tns) at (43em,7.468em) {};
\node (newa) at (40em,9.6em) {$\apsnode{A}{}$};
\node (epsbot) at (46em,9.6em) {$\apsnodei$};
\draw (pd) -- (tns);
\draw (tns) -- (epsbot);
\draw (tns) -- (newa);
\draw (epsbot) to [out=50,in=330] (pc);
\node (end) at (51em,4.8em) {$\apsnode{A}{A\ldr \lidc}$};
\node at (17.5em,4.8em) {$\rightarrow$};
\node at (17.5em,5.8em) {$\lstr\lidstr$};
\node at (27.5em,4.8em) {$\rightarrow$};
\node at (27.5em,5.8em) {$\lidc^{-1}$};
\node at (37.5em,4.8em) {$\rightarrow$};
\node at (37.5em,5.8em) {$\textit{Ax}$};
\node at (48.5em,4.8em) {$\rightarrow$};
\node at (48.5em,5.8em) {$\ldr R$};
\end{tikzpicture}
\end{center}

We expand the $A$ formula to $A\lstr\lidstr$ using the $\lstr\lidstr$ structural rule, then replace the structural $\lidstr$ by the $\lidc$. This adds a $\lidc$ hypothesis to the abstract proof structure. We then identify the vertices with the $\lidc$ hypothesis and with the $\lidc$ conclusion to produce a redex for the $\ldr R$ contraction, which we perform to complete the proof. 

We can modify our proof net calculus to do the connections of atomic formula at the abstract proof structure level instead of at the proof structure level. 

For comparison, we show the same proof but with the unit treated as a 0-ary connective. The conversion sequences are quite similar, and the contraction for $\lidc L$ works rather like the axiom connection in the reduction sequence above.
 
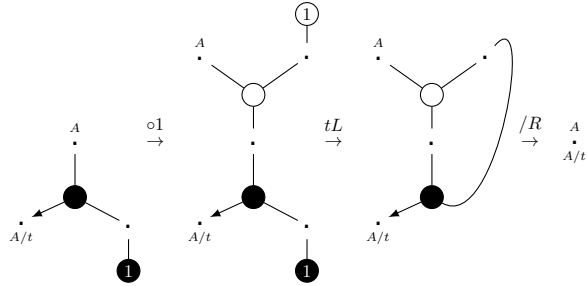
\begin{center}
\begin{tikzpicture}[scale=0.75]
\node (pa) at (10em,0em) {$\apsnode{}{A\ldr \lidc}$};
\node (pat) at (10.2em,0.44em) {};
\node[par] (pc) at (13em,1.732em) {};
\node (pb) at (16em,0.15em) {$\apsnodei$};
\node (pd) at (13em,4.8em) {$\apsnode{A}{}$};
\node (pdd) at (13em,4.5em) {};
\draw (pc) -- (pb);
\draw (pc) -- (pdd);
\path[>=latex,->]  (pc) edge (pat);
\node[par] (zpar) at (16em,-2.4em) {};
\node (zparl) at (16em,-2.4em) {\textcolor{white}{$\lidstr$}};
\draw (zpar) -- (pb);
\node (pa) at (20em,0) {$\apsnode{}{A\ldr \lidc}$};
\node (pat) at (20.2em,0.44em) {};
\node[par] (pc) at (23em,1.732em) {};
\node (pb) at (26em,0.15em) {$\apsnodei$};
\node (pd) at (23em,4.8em) {$\apsnodei$};
\node (pdd) at (23em,4.5em) {};
\draw (pc) -- (pb);
\draw (pc) -- (pdd);
\path[>=latex,->]  (pc) edge (pat);
\node[tns] (tns) at (23em,7.468em) {};
\node (newa) at (20em,9.6em) {$\apsnode{A}{}$};
\node (epsbot) at (26em,9.6em) {$\apsnodei$};
\draw (pd) -- (tns);
\draw (tns) -- (epsbot);
\draw (tns) -- (newa);
\node [tns] (1) at (26em,12em) {};
\node (tl) at (26em,12em) {$\lidstr$};
\draw (1) -- (epsbot);
\node[par] (zpar) at (26em,-2.4em) {};
\node (zparl) at (26em,-2.4em) {\textcolor{white}{$\lidstr$}};
\draw (zpar) -- (pb);
\node (pa) at (30em,0) {$\apsnode{}{A\ldr \lidc}$};
\node (pat) at (30.2em,0.44em) {};
\node[par] (pc) at (33em,1.732em) {};
\node (pd) at (33em,4.8em) {$\apsnodei$};
\node (pdd) at (33em,4.5em) {};
\draw (pc) -- (pdd);
\path[>=latex,->]  (pc) edge (pat);
\node[tns] (tns) at (33em,7.468em) {};
\node (newa) at (30em,9.6em) {$\apsnode{A}{}$};
\node (epsbot) at (36em,9.6em) {$\apsnodei$};
\draw (pd) -- (tns);
\draw (tns) -- (epsbot);
\draw (tns) -- (newa);
\draw (epsbot) to [out=50,in=330] (pc);
\node (end) at (41em,4.8em) {$\apsnode{A}{A\ldr \lidc}$};
\node at (17.5em,4.8em) {$\rightarrow$};
\node at (17.5em,5.8em) {$\lstr\lidstr$};
\node at (27.5em,4.8em) {$\rightarrow$};
\node at (27.5em,5.8em) {$\lidc L$};
\node at (38.5em,4.8em) {$\rightarrow$};
\node at (38.5em,5.8em) {$\ldr R$};
\end{tikzpicture}
\end{center}

\section{Meaning assignment for the unit}
\label{sec:unitsem}

\newcommand{\idconstrterm}{E^{\lidc}}
\newcommand{\idconstr}[2]{\idconstrterm(#1,#2)}
\newcommand{\idtermunit}{*}
\newcommand{\idtypeunit}{\textbf{1}}

Under the standard term assignments for intuitionistic (linear) logic \citep{Troelstra}, we obtain the following term assignment rules for the unit $\lidc$ in \nllam{}. 

\begin{center}
\begin{tabular}{ccc}
	\infer[\lidc I]{\lidstr\vdash \idtermunit^{\idtypeunit}:\lidc}{} &&
	\infer[\lidc E]{\Gamma[\Delta]\vdash \idconstr{M}{N}^{\alpha}:C}{\Delta\vdash N^{\idtypeunit}:t & \Gamma[1]\vdash M^{\alpha}:C}
\end{tabular}
\end{center}

As usual, the term assignments are for the natural deduction rules, and we note that the $\lidc E$ rule uses a context formula $C$, requiring permutation conversions for the term assignment rules  (confirming it behaves like  a 0-ary version of the binary `$\bullet$').

The term corresponding to the $\lidc I$ rule is the term constant `$*$' of type $\idtypeunit$ (the semantic type corresponding to the unit $\lidc$). The term constructor $\idconstrterm$, corresponding to the $\lidc E$ rule, takes a term $M$ of type $\alpha$ (corresponding to the semantic type of the formula $C$) and a term $N$ of type $\idtypeunit$ (the semantic type corresponding to the unit $t$), to produce a term of type $\alpha$.

The $\eta$ and $\beta$ conversion steps correspond to the proof normalisations shown, respectively as Equation~\ref{id:eta} and~\ref{id:beta} below.
\begin{align}\label{id:eta}
\infer[\lidc E]{\Gamma[\lidstr]\vdash \idconstr{M}{\idtermunit}^{\alpha}:C}{\infer[\lidc I]{\lidstr\vdash \idtermunit:t^{\idtypeunit}}{} & \infer*[\delta]{\Gamma[\lidstr]\vdash M^{\alpha}:C}{}} &
\quad\leadsto\quad
\infer*[\delta]{\Gamma[\lidstr]\vdash M^{\alpha}:C}{}
\end{align}

\begin{align}\label{id:beta}
\infer[\lidc E]{\Delta\vdash \idconstr{\idtermunit}{N}^{\idtypeunit}:t}{\infer*[\delta]{\Delta\vdash N^{\idtypeunit}:t}{} & \infer[\lidc I]{\lidstr\vdash \idtermunit^{\idtypeunit}:\lidc}{}} &
\quad\leadsto\quad
\infer*[\delta]{\Delta\vdash N^{\idtypeunit}:t}{}
\end{align}
The corresponding term rewrites are shown as Equations~\ref{eq:r1} and~\ref{eq:r2}.
\begin{align}
\label{eq:r1}\idconstr{M}{\idtermunit}^{\alpha}	&\leadsto M^{\alpha} \\
\label{eq:r2}\idconstr{\idtermunit}{N}^{\idtypeunit}	&\leadsto N^{\idtypeunit}
\end{align}

We will not list the full set of commutative conversions which are necessary here: the $\lidc E$ rule can permute with all logical rules\footnote{It permutes with the structural rules as well, but these rule permutations do not correspond the term equations.} and although the term conversions are fairly standard, there are rather many of them.
\section{A more complicated example}
\label{sec:example}

As a more complicated example showing how to use proof nets for \nllam{} theorem proving, we show how to derive ``everyone read the same book'', an example from Section~\ref{sec:parasitic}. We use the following lexical entries. 

\begin{align*}
\textit{everyone} &\quad s \bdr (np \bdl s) \\
\textit{read} &\quad (np \ldl s) \ldr np \\ 	
\textit{the} &\quad np\ldr n \\
\textit{same} &\quad (np \bdl s) \bdr ((n\ldl n)\bdl (np \bdl s))\\
\textit{book} &\quad n
\end{align*}

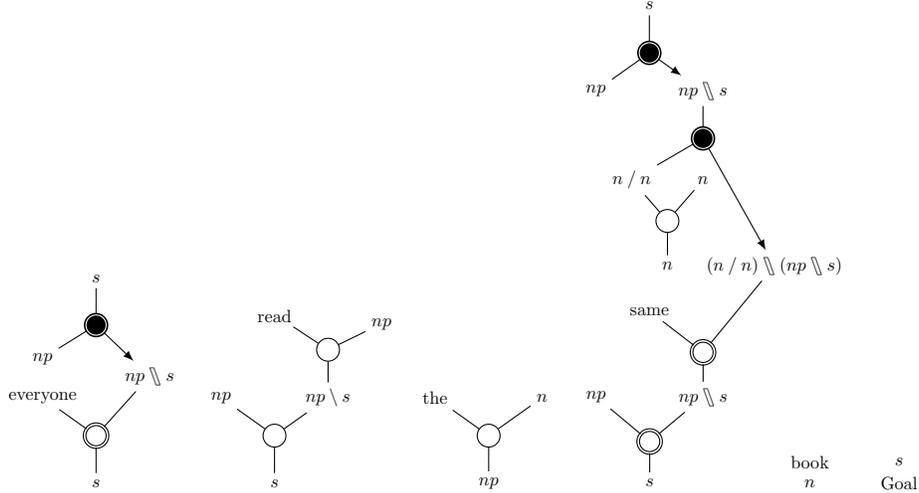
\begin{figure}
\begin{center}
\begin{tikzpicture}[scale=0.75]
\node (spar) at (3em,0em) {$s$};
\node (a) at (6em,6em) {$np\bdl s$};
\node (b) at (0em,4.8em) {everyone};
\node[ttns] (c) at (3em,2.668em) {};
\draw (c) -- (spar);
\draw (c) -- (a);
\draw (c) -- (b);
\node (d) at (0em,7em) {$np$};
\node [ppar] (pc) at (3em,8.868em) {};
\node (e) at (3em,11.5em) {$s$};
\path[>=latex,->]  (pc) edge (a);
\draw (pc) -- (d);
\draw (pc) -- (e);
\node (ab) at (16em,4.8em) {$np\ldl s$};
\node (a) at (13em,9.4em) {read};
\node (aa) at (13.7em,9em) {};
\node (obj) at (19em,8.95em) {$np$};
\node[tns] (c) at (16em,7.486em) {};
\draw (c) -- (ab);
\draw (c) -- (aa);
\draw (c) -- (obj);
\node (subj) at (10em,4.8em) {$np$};
\node [tns] (cc) at (13em,2.668em) {};
\node (stv) at (13em,0em) {$s$};
\draw (cc) -- (stv);
\draw (cc) -- (subj);
\draw (cc) -- (ab);
\node (narg) at (28em,4.8em) {$n$};
\node (the) at (22em,4.8em) {the};
\node [tns] (cdet) at (25em,2.668em) {};
\node (npgoal) at (25em,0em) {$np$};
\draw (cdet) -- (narg);
\draw (cdet) -- (npgoal);
\draw (cdet) -- (the);
\node (book) at (43em,1.2em) {book};
\node (bookn) at (43em,0em) {$n$};
\node (goals) at (48em,1.2em) {$s$};
\node (goal) at (48em,0em) {Goal};
\path (34em,26.8em) coordinate (sametop);
\path (sametop) ++(0em,-2.668em) coordinate (sparb);
\path (sametop) ++(-3em,-4.8em) coordinate (parbl);
\path (sametop) ++(3em,-4.8em) coordinate (parbr);
\node (sbot) at (sametop) {$s$};
\node [ppar] (parb) at (sparb) {};
\draw (sbot) -- (parb);
\node (npa) at (parbl) {$np$};
\node (vpb) at (parbr) {$np\bdl s$};
\draw (parb) -- (npa);
\path[>=latex,->]  (parb) edge (vpb);
\path (parbr) ++(0em,0em) coordinate (parcb);
\path (parcb) ++(0em,-2.668em) coordinate (parcc);
\path (parcb) ++(-4em,-5em) coordinate (parcl);
\path (parcb) ++(4em,-9.8em) coordinate (parcr);
\node (nn) at (parcl) {$n\ldr n$};
\node (nnnps) at (parcr) {$(n\ldr n)\bdl(np\bdl s)$};
\node [ppar] (prc) at (parcc) {};
\draw (vpb) -- (prc);
\draw (prc) -- (nn); 
\path[>=latex,->]  (prc) edge (nnnps);
\path (nn) ++(0em,0em) coordinate (tnnl);
\path (tnnl) ++(4em,0em) coordinate (tnnr);
\path (tnnl) ++(2em,-2.3em) coordinate (tnnc);
\path (tnnl) ++(2em,-4.8em) coordinate (tnnb);
\node (nb) at (tnnb) {$n$};
\node [tns] (nt) at (tnnc) {};
\node (nr) at (tnnr) {$n$};
\draw (nt) -- (nr);
\draw (nt) -- (nn);
\draw (nt) -- (nb);
\path (nnnps) ++(0em,0em) coordinate (nnnpsa);
\path (parcc) ++(0em,-12em) coordinate (sametp);
\path (sametp) ++(0em,-2.5em) coordinate (samebp);
\path (sametp) ++(-3em,+2.3em) coordinate (samep);
\node (same) at (samep) {same};
\node [ttns] (st) at (sametp) {};
\node (vpr) at (samebp) {$np\bdl s$};  
\draw (st) -- (same);
\draw (st) -- (vpr);
\draw (st) -- (nnnps);
\path (samebp) ++(0em,0em) coordinate (vprp);
\path (vprp) ++(-6em,0em) coordinate (vplp);
\path (vprp) ++(-3em,-2.5em) coordinate (vpcp);
\path (vpcp) ++(0em,-2.3em) coordinate (vpbp);
\node (subj) at (vplp) {$np$};
\node [ttns] (tc) at (vpcp) {};
\node (bot) at (vpbp) {$s$};
\draw (tc) -- (bot);
\draw (tc) -- (subj);
\draw (tc) -- (vpr);
\end{tikzpicture}
\end{center}
\caption{Formula unfolding for ``everyone read the same book''.}
\label{fig:sameunfold}
\end{figure}

Given the above lexicon, Figure~\ref{fig:sameunfold} shows the formula unfolding for the sentence ``everyone read the same book''. To improve readability of the structure and to save space, the formula of each lexical entry has been replaced by the label of the corresponding word, so, for example, ``the'' denotes the formula $np\ldr n$.

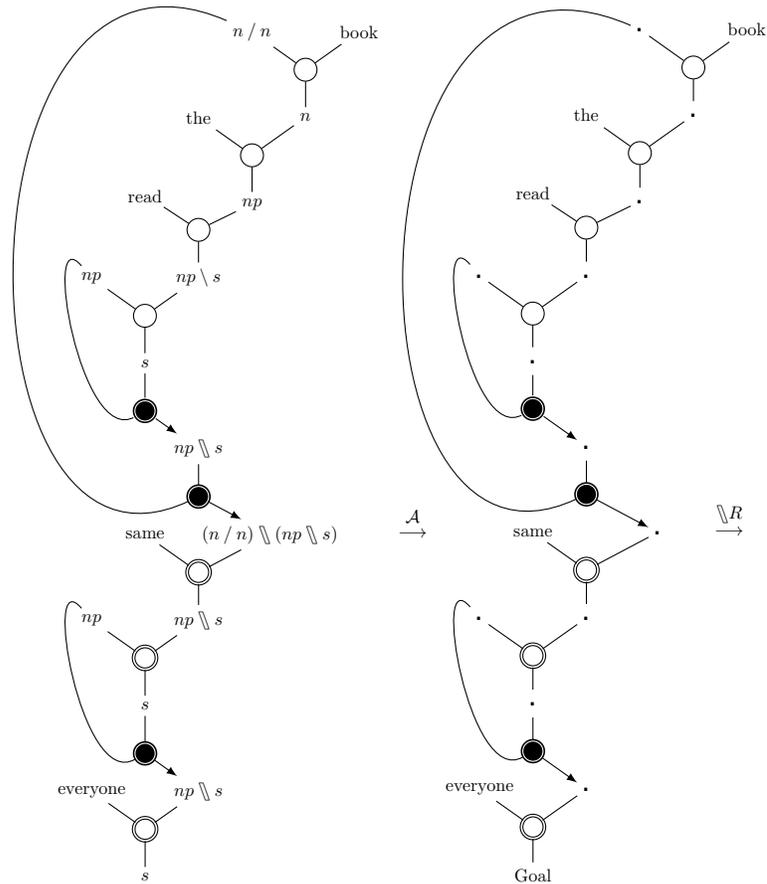
\begin{figure}
\begin{center}
\begin{tikzpicture}[scale=0.75]
\node (ab) at (16em,4.8em) {$np\ldl s$};
\node (a) at (13em,9.4em) {read};
\node (aa) at (13.7em,9em) {};
\node (obj) at (19em,8.95em) {$np$};
\node[tns] (c) at (16em,7.486em) {};
\draw (c) -- (ab);
\draw (c) -- (aa);
\draw (c) -- (obj);
\node (subj) at (10em,4.8em) {$np$};
\node [tns] (cc) at (13em,2.668em) {};
\node (stv) at (13em,0em) {$s$};
\draw (cc) -- (stv);
\draw (cc) -- (subj);
\draw (cc) -- (ab);
\node (narg) at (22em,13.8em) {$n$};
\node (the) at (16em,13.8em) {the};
\node [tns] (cdet) at (19em,11.668em) {};
\draw (cdet) -- (narg);
\draw (cdet) -- (obj);
\draw (cdet) -- (the);
\node (nbook) at (25em,18.6em) {book};
\node (adj) at (19em,18.6em) {$n\ldr n$};
\node [tns] (cadj) at (22em,16.468em) {};
\draw (cadj) -- (nbook);
\draw (cadj) -- (narg);
\draw (cadj) -- (adj);
\path (13em,0em) coordinate (sametop);
\path (sametop) ++(0em,-2.668em) coordinate (sparb);
\path (sametop) ++(-3em,-4.8em) coordinate (parbl);
\path (sametop) ++(3em,-4.8em) coordinate (parbr);
%
\node [ppar] (parb) at (sparb) {};
\draw (stv) -- (parb);
\node (vpb) at (parbr) {$np\bdl s$};
\path[>=latex,->]  (parb) edge (vpb);
\path (parbr) ++(0em,0em) coordinate (parcb);
\path (parcb) ++(0em,-2.668em) coordinate (parcc);
\path (parcb) ++(-4em,-5em) coordinate (parcl);
\path (parcb) ++(4em,-4.8em) coordinate (parcr);
\path (parcr) ++(8em,0em) coordinate (arc);
\path (arc) ++(0em,1em) coordinate (arcl);
\node (arrow) at (arc) {$\longrightarrow$};
\node at (arcl) {$\mathcal{A}$};
%
\node (nnnps) at (parcr) {$(n\ldr n)\bdl(np\bdl s)$};
\node [ppar] (prc) at (parcc) {};
\draw (vpb) -- (prc);
\path[>=latex,->]  (prc) edge (nnnps);
\draw (subj) to [out=130,in=210] (parb);
\draw (adj)..controls(2em,26em)and(2em,-14em)..(prc);
\path (parcr) ++(0em,0em) coordinate (tnr);
\path (tnr) ++(-7em,0em) coordinate (tnl);
\path (parbr) ++(0em,-9.6em) coordinate (tnb);
\path (tnb) ++(0em,2.686em) coordinate (tnc);
\node (ts) [ttns] at (tnc) {};
\node (npbs) at (tnb) {$np\bdl s$};
\node (same) at (tnl) {same};
\draw (ts) -- (npbs);
\draw (ts) -- (same);
\draw (ts) -- (nnnps);
\path (tnb) ++(0em,0em) coordinate (tr);
\path (tr) ++(-3em,-4.8em) coordinate (tb);
\path (tr) ++(-6em,0em) coordinate (tl);
\path (tb) ++(0em,2.686em) coordinate (tc);
\node (snp) at (tl) {$np$};
\node (t) [ttns] at (tc) {};
\node (ss) at (tb) {$s$};
\draw (t) -- (ss);
\draw (t) -- (snp);
\draw (t) -- (npbs);
\path (tb) ++(0em,0em) coordinate (evtop);
\path (evtop) ++(0em,-2.668em) coordinate (eparc);
\path (evtop) ++(-3em,-4.8em) coordinate (eparl);
\path (evtop) ++(3em,-4.8em) coordinate (eparr);
\node [ppar] (parb) at (sparb) {};
\node (vpe) at (eparr) {$np\bdl s$};
\node (ep) [ppar] at (eparc) {};
\draw (ss) -- (ep);
\path[>=latex,->]  (ep) edge (vpe);
\draw (snp) to [out=130,in=210] (ep);
\path (eparr) ++(-6em,0em) coordinate (etl);
\path (etl) ++(3em,-4.8em) coordinate (etb);
\path (etb) ++(0em,2.686em) coordinate (etc);
\node (et) [ttns] at (etc) {};
\node (ev) at (etl) {everyone};
\node (bot) at (etb) {$s$};
\draw (et) -- (bot);
\draw (et) -- (ev);
\draw (et) -- (vpe);
\end{tikzpicture}
\kern-4em
\begin{tikzpicture}[scale=0.75]
\node (ab) at (16em,4.8em) {$\apsnodei$};
\node (a) at (13em,9.4em) {read};
\node (aa) at (13.7em,9em) {};
\node (obj) at (19em,8.95em) {$\apsnodei$};
\node[tns] (c) at (16em,7.486em) {};
\draw (c) -- (ab);
\draw (c) -- (aa);
\draw (c) -- (obj);
\node (subj) at (10em,4.8em) {$\apsnodei$};
\node [tns] (cc) at (13em,2.668em) {};
\node (stv) at (13em,0em) {$\apsnodei$};
\draw (cc) -- (stv);
\draw (cc) -- (subj);
\draw (cc) -- (ab);
\node (narg) at (22em,13.8em) {$\apsnodei$};
\node (the) at (16em,13.8em) {the};
\node [tns] (cdet) at (19em,11.668em) {};
\draw (cdet) -- (narg);
\draw (cdet) -- (obj);
\draw (cdet) -- (the);
\node (nbook) at (25em,18.6em) {book};
\node (adj) at (19em,18.6em) {$\apsnodei$};
\node [tns] (cadj) at (22em,16.468em) {};
\draw (cadj) -- (nbook);
\draw (cadj) -- (narg);
\draw (cadj) -- (adj);
\path (13em,0em) coordinate (sametop);
\path (sametop) ++(0em,-2.668em) coordinate (sparb);
\path (sametop) ++(-3em,-4.8em) coordinate (parbl);
\path (sametop) ++(3em,-4.8em) coordinate (parbr);
%
\node [ppar] (parb) at (sparb) {};
\draw (stv) -- (parb);
\node (vpb) at (parbr) {$\apsnodei$};
\path[>=latex,->]  (parb) edge (vpb);
\path (parbr) ++(0em,0em) coordinate (parcb);
\path (parcb) ++(0em,-2.668em) coordinate (parcc);
\path (parcb) ++(-4em,-5em) coordinate (parcl);
\path (parcb) ++(4em,-4.8em) coordinate (parcr);
%
\path (parcr) ++(4em,0em) coordinate (arc);
\path (arc) ++(0em,1em) coordinate (arcl);
\node (arrow) at (arc) {$\longrightarrow$};
\node at (arcl) {$\bdl R$};
%
\node (nnnps) at (parcr) {$\apsnodei$};
\node [ppar] (prc) at (parcc) {};
\draw (vpb) -- (prc);
\path[>=latex,->]  (prc) edge (nnnps);
\draw (subj) to [out=130,in=210] (parb);
\draw (adj)..controls(2em,26em)and(2em,-14em)..(prc);
\path (parcr) ++(0em,0em) coordinate (tnr);
\path (tnr) ++(-7em,0em) coordinate (tnl);
\path (parbr) ++(0em,-9.6em) coordinate (tnb);
\path (tnb) ++(0em,2.686em) coordinate (tnc);
\node (ts) [ttns] at (tnc) {};
\node (npbs) at (tnb) {$\apsnodei$};
\node (same) at (tnl) {same};
\draw (ts) -- (npbs);
\draw (ts) -- (same);
\draw (ts) -- (nnnps);
\path (tnb) ++(0em,0em) coordinate (tr);
\path (tr) ++(-3em,-4.8em) coordinate (tb);
\path (tr) ++(-6em,0em) coordinate (tl);
\path (tb) ++(0em,2.686em) coordinate (tc);
\node (snp) at (tl) {$\apsnodei$};
\node (t) [ttns] at (tc) {};
\node (ss) at (tb) {$\apsnodei$};
\draw (t) -- (ss);
\draw (t) -- (snp);
\draw (t) -- (npbs);
\path (tb) ++(0em,0em) coordinate (evtop);
\path (evtop) ++(0em,-2.668em) coordinate (eparc);
\path (evtop) ++(-3em,-4.8em) coordinate (eparl);
\path (evtop) ++(3em,-4.8em) coordinate (eparr);
\node [ppar] (parb) at (sparb) {};
\node (vpe) at (eparr) {$\apsnodei$};
\node (ep) [ppar] at (eparc) {};
\draw (ss) -- (ep);
\path[>=latex,->]  (ep) edge (vpe);
\draw (snp) to [out=130,in=210] (ep);
\path (eparr) ++(-6em,0em) coordinate (etl);
\path (etl) ++(3em,-4.8em) coordinate (etb);
\path (etb) ++(0em,2.686em) coordinate (etc);
\node (et) [ttns] at (etc) {};
\node (ev) at (etl) {everyone};
\node (bot) at (etb) {Goal};
\draw (et) -- (bot);
\draw (et) -- (ev);
\draw (et) -- (vpe);
\end{tikzpicture}
\end{center}
\caption{Proof structure (left) and the corresponding abstract proof structure (right) for ``everyone read the same book''.}
\label{fig:sameps}
\end{figure}

We identify the atomic formulas of the structure to obtain the proof structure shown on the left of Figure~\ref{fig:sameps}. The par link of ``everyone'' forms a redex with the bottom tensor link of ``same'', the topmost par link of ``same'' will ensure that ``everyone'' appears as the subject, whereas the other par link of ``same'' has its position determined by the place of the adjective subformula $n\ldr n$, which appears between ``the'' and ``book'', since this is where we want the word ``same'' to end up.

The corresponding abstract proof structure is shown on the right of Figure~\ref{fig:sameps}. 
This should by now be familiar: all internal formulas have been removed and only the hypotheses and conclusions of the structure are still assigned formulas.

To show this is a valid proof net, we have to convert the abstract proof structure of Figure~\ref{fig:sameps} to a tree of Lambek (singly circled) tensor links, using the contraction rules of Table~\ref{tab:contr}, the structural rules of Table~\ref{tab:srlam}. There easiest way to do this, it to use the rewrite rules of Table~\ref{tab:srlam} from left to right only and to use the derived  $\betaexp\bdl$ rule of Table~\ref{tab:deriv}, following the discussion in Section~\ref{sec:deci}.

\begin{figure}
\begin{center}
\begin{tikzpicture}[scale=0.75]
\node (ab) at (16em,4.8em) {$\apsnodei$};
\node (a) at (13em,9.4em) {read};
\node (aa) at (13.7em,9em) {};
\node (obj) at (19em,8.95em) {$\apsnodei$};
\node[tns] (c) at (16em,7.486em) {};
\draw (c) -- (ab);
\draw (c) -- (aa);
\draw (c) -- (obj);
\node (subj) at (10em,4.8em) {$\apsnodei$};
\node [tns] (cc) at (13em,2.668em) {};
\node (stv) at (13em,0em) {$\apsnodei$};
\draw (cc) -- (stv);
\draw (cc) -- (subj);
\draw (cc) -- (ab);
\node (narg) at (22em,13.8em) {$\apsnodei$};
\node (the) at (16em,13.8em) {the};
\node [tns] (cdet) at (19em,11.668em) {};
\draw (cdet) -- (narg);
\draw (cdet) -- (obj);
\draw (cdet) -- (the);
\node (nbook) at (25em,18.6em) {book};
\node (adj) at (19em,18.6em) {$\apsnodei$};
\node [tns] (cadj) at (22em,16.468em) {};
\draw (cadj) -- (nbook);
\draw (cadj) -- (narg);
\draw (cadj) -- (adj);
\path (13em,0em) coordinate (sametop);
\path (sametop) ++(0em,-2.668em) coordinate (sparb);
\path (sametop) ++(-3em,-4.8em) coordinate (parbl);
\path (sametop) ++(3em,-4.8em) coordinate (parbr);
\path (sametop) ++(12em,0em) coordinate (arc);
\path (arc) ++(0em,1em) coordinate (arcl);
\node (arrow) at (arc) {$\longrightarrow$};
\node at (arcl) {$\betaexp\bdl$};
\node [ppar] (parb) at (sparb) {};
\draw (stv) -- (parb);
\node (vpb) at (parbr) {$\apsnodei$};
\path[>=latex,->]  (parb) edge (vpb);
\path (parbr) ++(0em,0em) coordinate (parcb);
\path (parcb) ++(0em,-2.668em) coordinate (parcc);
\path (parcb) ++(-4em,-5em) coordinate (parcl);
\path (parcb) ++(4em,-4.8em) coordinate (parcr);
\node (nnnps) at (parcr) {$\apsnodei$};
\node [ppar] (prc) at (parcc) {};
\draw (vpb) -- (prc);
\path[>=latex,->]  (prc) edge (nnnps);
\draw (subj) to [out=130,in=210] (parb);
\draw (adj)..controls(2em,26em)and(2em,-14em)..(prc);
\path (parcr) ++(0em,0em) coordinate (tnr);
\path (tnr) ++(-7em,0em) coordinate (tnl);
\path (parbr) ++(0em,-9.6em) coordinate (tnb);
\path (tnb) ++(0em,2.686em) coordinate (tnc);
\node (ts) [ttns] at (tnc) {};
\node (npbs) at (tnb) {$\apsnodei$};
\node (same) at (tnl) {same};
\draw (ts) -- (npbs);
\draw (ts) -- (same);
\draw (ts) -- (nnnps);
\path (tnb) ++(0em,0em) coordinate (tr);
\path (tr) ++(-3em,-4.8em) coordinate (tb);
\path (tr) ++(-6em,0em) coordinate (tl);
\path (tb) ++(0em,2.686em) coordinate (tc);
\node (snp) at (tl) {everyone};
\node (t) [ttns] at (tc) {};
\node (ss) at (tb) {Goal};
\draw (t) -- (ss);
\draw (t) -- (snp);
\draw (t) -- (npbs);
\end{tikzpicture}
\begin{tikzpicture}[scale=0.75]
\node (ab) at (16em,4.8em) {$\apsnodei$};
\node (a) at (13em,9.4em) {read};
\node (aa) at (13.7em,9em) {};
\node (obj) at (19em,8.95em) {$\apsnodei$};
\node[tns] (c) at (16em,7.486em) {};
\draw (c) -- (ab);
\draw (c) -- (aa);
\draw (c) -- (obj);
\node (subj) at (10em,4.8em) {$\apsnodei$};
\node [tns] (cc) at (13em,2.668em) {};
\node (stv) at (13em,0em) {$\apsnodei$};
\draw (cc) -- (stv);
\draw (cc) -- (subj);
\draw (cc) -- (ab);
\node (narg) at (22em,13.8em) {$\apsnodei$};
\node (the) at (16em,13.8em) {the};
\node [tns] (cdet) at (19em,11.668em) {};
\draw (cdet) -- (narg);
\draw (cdet) -- (obj);
\draw (cdet) -- (the);
\node (nbook) at (25em,18.6em) {book};
\node (adj) at (19em,18.6em) {$\apsnodei$};
\node [tns] (cadj) at (22em,16.468em) {};
\draw (cadj) -- (nbook);
\draw (cadj) -- (narg);
\draw (cadj) -- (adj);
\path (13em,0em) coordinate (sametop);
\path (sametop) ++(0em,-2.668em) coordinate (sparb);
\path (sametop) ++(-3em,-4.8em) coordinate (parbl);
\path (sametop) ++(3em,-4.8em) coordinate (parbr);
%
\path (sametop) ++(10em,0em) coordinate (arc);
\path (arc) ++(0em,1em) coordinate (arcl);
\node (arrow) at (arc) {$\longrightarrow$};
\node at (arcl) {$\betaexp\bdl$};
\node [tns] (parb) at (sparb) {};
\node at (sparb) {$\lambda$};
\draw (stv) -- (parb);
\node (vpb) at (parbr) {$\apsnodei$};
\draw (parb) -- (vpb);
\path (parbr) ++(0em,0em) coordinate (parcb);
\path (parcb) ++(0em,-2.668em) coordinate (parcc);
\path (parcb) ++(-4em,-5em) coordinate (parcl);
\path (parcb) ++(4em,-4.8em) coordinate (parcr);
\node (nnnps) at (parcr) {$\apsnodei$};
\node [ppar] (prc) at (parcc) {};
\draw (vpb) -- (prc);
\path[>=latex,->]  (prc) edge (nnnps);
\draw (subj) to [out=130,in=210] (parb);
\draw (adj)..controls(2em,26em)and(2em,-14em)..(prc);
\path (parcr) ++(0em,0em) coordinate (tnr);
\path (tnr) ++(-7em,0em) coordinate (tnl);
\path (parbr) ++(0em,-9.6em) coordinate (tnb);
\path (tnb) ++(0em,2.686em) coordinate (tnc);
\node (ts) [ttns] at (tnc) {};
\node (npbs) at (tnb) {$\apsnodei$};
\node (same) at (tnl) {same};
\draw (ts) -- (npbs);
\draw (ts) -- (same);
\draw (ts) -- (nnnps);
\path (tnb) ++(0em,0em) coordinate (tr);
\path (tr) ++(-3em,-4.8em) coordinate (tb);
\path (tr) ++(-6em,0em) coordinate (tl);
\path (tb) ++(0em,2.686em) coordinate (tc);
\node (snp) at (tl) {everyone};
\node (t) [ttns] at (tc) {};
\node (ss) at (tb) {Goal};
\draw (t) -- (ss);
\draw (t) -- (snp);
\draw (t) -- (npbs);
\end{tikzpicture}
\end{center}
\caption{The abstract proof structure on the right of Figure~\ref{fig:sameps} after the $\bdl R$ contraction (left) and after the $\betaexp\bdl$ conversion (right).}
\label{fig:samec}
\end{figure}
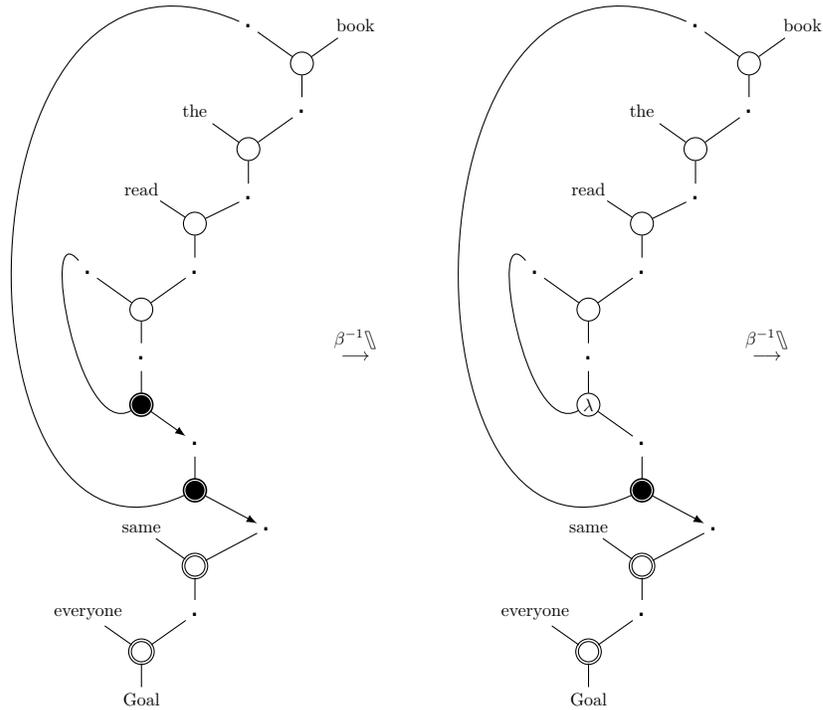

\begin{figure}
\begin{center}
\begin{tikzpicture}[scale=0.75]
\node (ab) at (16em,4.8em) {$\apsnodei$};
\node (a) at (13em,9.4em) {read};
\node (aa) at (13.7em,9em) {};
\node (obj) at (19em,8.95em) {$\apsnodei$};
\node[tns] (c) at (16em,7.486em) {};
\draw (c) -- (ab);
\draw (c) -- (aa);
\draw (c) -- (obj);
\node (subj) at (10em,4.8em) {$\apsnodei$};
\node [tns] (cc) at (13em,2.668em) {};
\node (stv) at (13em,0em) {$\apsnodei$};
\draw (cc) -- (stv);
\draw (cc) -- (subj);
\draw (cc) -- (ab);
\node (narg) at (22em,13.8em) {$\apsnodei$};
\node (the) at (16em,13.8em) {the};
\node [tns] (cdet) at (19em,11.668em) {};
\draw (cdet) -- (narg);
\draw (cdet) -- (obj);
\draw (cdet) -- (the);
\node (nbook) at (25em,18.6em) {book};
\node (adj) at (19em,18.6em) {$\apsnodei$};
\node [tns] (cadj) at (22em,16.468em) {};
\draw (cadj) -- (nbook);
\draw (cadj) -- (narg);
\draw (cadj) -- (adj);
\path (13em,0em) coordinate (sametop);
\path (sametop) ++(0em,-2.668em) coordinate (sparb);
\path (sametop) ++(-3em,-4.8em) coordinate (parbl);
\path (sametop) ++(3em,-4.8em) coordinate (parbr);
\node [tns] (parb) at (sparb) {};
\node at (sparb) {$\lambda$};
\draw (stv) -- (parb);
\node (vpb) at (parbr) {$\apsnodei$};
\draw (parb) -- (vpb);
\path (parbr) ++(0em,0em) coordinate (parcb);
\path (parcb) ++(0em,-2.668em) coordinate (parcc);
\path (parcb) ++(-4em,-5em) coordinate (parcl);
\path (parcb) ++(4em,-4.8em) coordinate (parcr);
\node (nnnps) at (parcr) {$\apsnodei$};
\node [tns] (prc) at (parcc) {};
\node at (parcc) {$\lambda$};
\draw (vpb) -- (prc);
\draw (prc) -- (nnnps);
\draw (subj) to [out=130,in=210] (parb);
\draw (adj)..controls(2em,26em)and(2em,-14em)..(prc);
\path (parcr) ++(0em,0em) coordinate (tnr);
\path (tnr) ++(-7em,0em) coordinate (tnl);
\path (parbr) ++(0em,-9.6em) coordinate (tnb);
\path (tnb) ++(0em,2.686em) coordinate (tnc);
\node (ts) [ttns] at (tnc) {};
\node (npbs) at (tnb) {$\apsnodei$};
\node (same) at (tnl) {same};
\draw (ts) -- (npbs);
\draw (ts) -- (same);
\draw (ts) -- (nnnps);
\path (tnb) ++(0em,0em) coordinate (tr);
\path (tr) ++(-3em,-4.8em) coordinate (tb);
\path (tr) ++(-6em,0em) coordinate (tl);
\path (tb) ++(0em,2.686em) coordinate (tc);
\node (snp) at (tl) {everyone};
\node (t) [ttns] at (tc) {};
\node (ss) at (tb) {Goal};
\draw (t) -- (ss);
\draw (t) -- (snp);
\draw (t) -- (npbs);
\path (parbr) ++(7em,0em) coordinate (arc);
\path (arc) ++(0em,1em) coordinate (arcl);
\node (arrow) at (arc) {$\longrightarrow$};
\node at (arcl) {$\betared$};
\end{tikzpicture}
\begin{tikzpicture}[scale=0.75]
\node (ab) at (16em,4.8em) {$\apsnodei$};
\node (a) at (13em,9.4em) {read};
\node (aa) at (13.7em,9em) {};
\node (obj) at (19em,8.95em) {$\apsnodei$};
\node[tns] (c) at (16em,7.486em) {};
\draw (c) -- (ab);
\draw (c) -- (aa);
\draw (c) -- (obj);
\node (subj) at (10em,4.8em) {$\apsnodei$};
\node [tns] (cc) at (13em,2.668em) {};
\node (stv) at (13em,0em) {$\apsnodei$};
\draw (cc) -- (stv);
\draw (cc) -- (subj);
\draw (cc) -- (ab);
\node (narg) at (22em,13.8em) {$\apsnodei$};
\node (the) at (16em,13.8em) {the};
\node [tns] (cdet) at (19em,11.668em) {};
\draw (cdet) -- (narg);
\draw (cdet) -- (obj);
\draw (cdet) -- (the);
\node (nbook) at (25em,18.6em) {book};
\node (adj) at (19em,18.6em) {same};
\node [tns] (cadj) at (22em,16.468em) {};
\draw (cadj) -- (nbook);
\draw (cadj) -- (narg);
\draw (cadj) -- (adj);
\path (13em,0em) coordinate (sametop);
\path (sametop) ++(0em,-2.668em) coordinate (sparb);
\path (sametop) ++(-3em,-4.8em) coordinate (parbl);
\path (sametop) ++(3em,-4.8em) coordinate (parbr);
\node [tns] (parb) at (sparb) {};
\node at (sparb) {$\lambda$};
\draw (stv) -- (parb);
\node (vpb) at (parbr) {$\apsnodei$};
\draw (parb) -- (vpb);
\draw (subj) to [out=130,in=210] (parb);
\path (parbr) ++(0em,0em) coordinate (tr);
\path (tr) ++(-3em,-4.8em) coordinate (tb);
\path (tr) ++(-6em,0em) coordinate (tl);
\path (tb) ++(0em,2.686em) coordinate (tc);
\node (snp) at (tl) {everyone};
\node (t) [ttns] at (tc) {};
\node (ss) at (tb) {Goal};
\draw (t) -- (ss);
\draw (t) -- (snp);
\draw (t) -- (vpb);
\path (ab) ++(7em,0em) coordinate (arc);
\path (arc) ++(0em,1em) coordinate (arcl);
\node (arrow) at (arc) {$\longrightarrow$};
\node at (arcl) {$\betared$};
\end{tikzpicture}
\end{center}
\caption{The abstract proof structure of Figure~\ref{fig:samec} after the $\betaexp\bdl$ contraction (left) and the $\beta$ conversion (right).}
\label{fig:samed}
\end{figure}
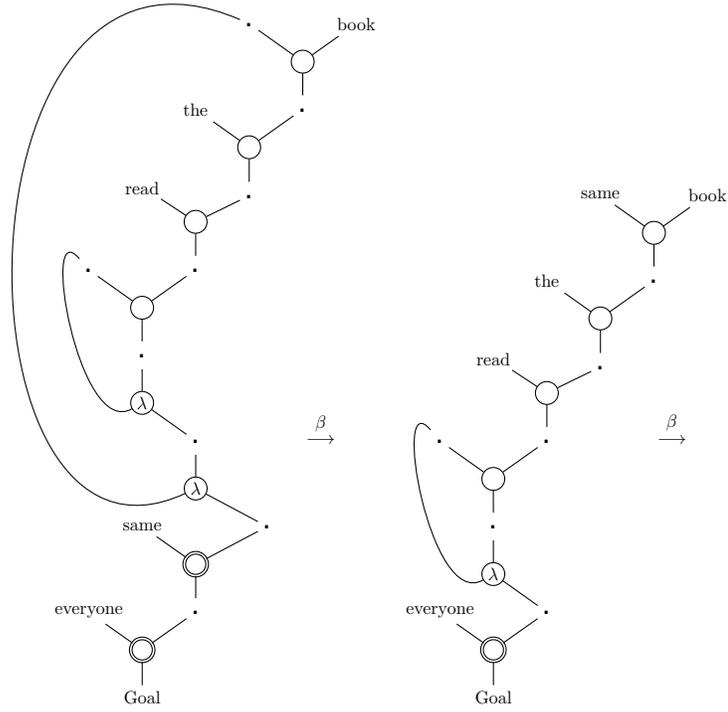

We start with the $\bdl R$ contraction, which produces the abstract proof structure shown on the left of Figure~\ref{fig:samec}. There are no further contractions possible, so we apply the $\betaexp\bdl$ conversion to the two remaining $\bdl R$ par links to obtain first the structure shown in Figure~\ref{fig:samec} on the right, then the structure shown in Figure~\ref{fig:samed} on the left.

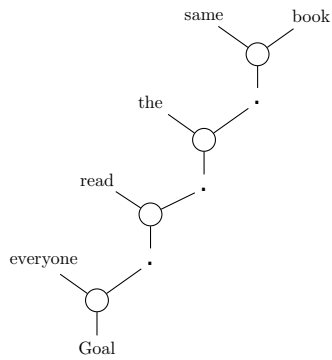
\begin{figure}
\begin{center}
\begin{tikzpicture}[scale=0.75]
\node (ab) at (16em,4.8em) {$\apsnodei$};
\node (a) at (13em,9.4em) {read};
\node (aa) at (13.7em,9em) {};
\node (obj) at (19em,8.95em) {$\apsnodei$};
\node[tns] (c) at (16em,7.486em) {};
\draw (c) -- (ab);
\draw (c) -- (aa);
\draw (c) -- (obj);
\node (subj) at (10em,4.8em) {everyone};
\node [tns] (cc) at (13em,2.668em) {};
\node (stv) at (13em,0em) {Goal};
\draw (cc) -- (stv);
\draw (cc) -- (subj);
\draw (cc) -- (ab);
\node (narg) at (22em,13.8em) {$\apsnodei$};
\node (the) at (16em,13.8em) {the};
\node [tns] (cdet) at (19em,11.668em) {};
\draw (cdet) -- (narg);
\draw (cdet) -- (obj);
\draw (cdet) -- (the);
\node (nbook) at (25em,18.6em) {book};
\node (adj) at (19em,18.6em) {same};
\node [tns] (cadj) at (22em,16.468em) {};
\draw (cadj) -- (nbook);
\draw (cadj) -- (narg);
\draw (cadj) -- (adj);
\end{tikzpicture}
\end{center}
\caption{Final tree computed after $\beta$ reduction from the abstract proof structure on the right of Figure~\ref{fig:samed}.}
\label{fig:sameend}
\end{figure}

We are now in the situation where we can apply the $\betared$ conversion twice in succession, first putting ``same'' in the right place, , as shown in Figure~\ref{fig:samed} on the right, then ``everyone'' as shown in Figure~\ref{fig:sameend}. This is a Lambek tree with the required yield ``everyone read the same book''. We have therefore shown that the proof structure back in Figure~\ref{fig:sameps} is a proof net.

\bibliographystyle{agsm}
\bibliography{moot}

\end{document}